\pdfoutput=1

\documentclass[preprint,authoryear,10pt]{elsarticle}

\usepackage[english]{babel}
\usepackage{lineno} 
\usepackage{geometry}
\usepackage[utf8]{inputenc} 
\usepackage[T1]{fontenc}    
\usepackage{microtype}      
\usepackage{lipsum}		

\usepackage{amsmath}
\usepackage{amssymb}
\usepackage{amsfonts}
\usepackage{fancyvrb}
\usepackage{fancybox}
\usepackage{bm}
\usepackage{graphicx}
\usepackage{subcaption}
\usepackage[colorlinks=true, allcolors=blue]{hyperref}
\usepackage{algorithm}
\usepackage{algpseudocode}
\usepackage{tabularray}
\usepackage{threeparttable}
\usepackage[at]{easylist}
\usepackage{appendix} 
\usepackage[figuresright]{rotating}
\usepackage{bbding}
\usepackage{natbib}
\setcitestyle{authoryear, open={(},close={)}}

\usepackage{listings}
\usepackage{latexsym}

\usepackage[T1]{fontenc}
\usepackage[utf8]{inputenc}
\usepackage{lmodern}

\usepackage{tcolorbox}
\usepackage[frozencache=true,cachedir=minted-cache]{minted} 
\usepackage{mdframed}
\tcbuselibrary{minted,skins, breakable}
\usepackage{fvextra}[breaklines=true]

\usepackage{xcolor}
\definecolor{mylightblue}{RGB}{100,149,237} 

\tcbset{
  enhanced, 
  colback=white!200!black, 
  title filled, 
  coltitle=white, 
  fonttitle=\bfseries, 
  arc=3mm, 
  outer arc=3mm, 
  boxrule=0.5mm, 
  toprule=0.5mm, 
  bottomrule=0.5mm, 
  titlerule=0.5mm, 
  drop fuzzy shadow, 
  minted options={ 
    breaklines, 
    fontsize=\footnotesize, 
    linenos, 
    numbersep=2mm, 
    framesep=1.5mm, 
  },
}

\usepackage{color}

\usepackage{colortbl}

\begin{document}
\begin{frontmatter}

\title{Large Language Model-Enhanced Reinforcement Learning for Generic Bus Holding Control Strategies}

\author{Jiajie Yu}
\ead{jiajie.yu@polyu.edu.hk}

\author{Yuhong Wang}
\ead{yuhong.wang@polyu.edu.hk}

\author{Wei Ma\corref{mycorrespondingauthor}}
\cortext[mycorrespondingauthor]{Corresponding author}
\ead{wei.w.ma@polyu.edu.hk}

\address{Civil and Environmental Engineering, The Hong Kong Polytechnic University\\Hung Hom, Kowloon, Hong Kong Special Administrative Region, China}

\begin{abstract}
Bus holding control is a widely-adopted strategy for maintaining stability and improving the operational efficiency of bus systems. Traditional model-based methods often face challenges with the low accuracy of bus state prediction and passenger demand estimation. In contrast, Reinforcement Learning (RL), as a data-driven approach, has demonstrated great potential in formulating bus holding strategies. 
RL determines the optimal control strategies in order to maximize the cumulative reward, which reflects the overall control goals.
However, translating sparse and delayed control goals in real-world tasks into dense and real-time rewards for RL is challenging, normally requiring extensive manual trial-and-error. 
In view of this, this study introduces an automatic reward generation paradigm by leveraging the in-context learning and reasoning capabilities of Large Language Models (LLMs). This new paradigm, termed the LLM-enhanced RL, comprises several LLM-based modules: reward initializer, reward modifier, performance analyzer, and reward refiner. These modules cooperate to initialize and iteratively improve the reward function according to the feedback from training and test results for the specified RL-based task. Ineffective reward functions generated by the LLM are filtered out to ensure the stable evolution of the RL agents' performance over iterations. To evaluate the feasibility of the proposed LLM-enhanced RL paradigm, it is applied to extensive bus holding control scenarios that vary in the number of bus lines, stops, and passenger demand. The results demonstrate the superiority, generalization capability, and robustness of the proposed paradigm compared to vanilla RL strategies, the LLM-based controller, physics-based feedback controllers, and optimization-based controllers. This study sheds light on the great potential of utilizing LLMs in various smart mobility applications. 
\end{abstract}


\end{frontmatter}


\section{Introduction}

Bus holding control is widely recognized as an efficient strategy for maintaining the stability of bus systems and minimizing passenger waiting time by preventing bus bunching \citep{jinwenlong2017partb, gkiotsalitis2021atstopControlReview}. It can be achieved by holding an early bus at the stop until it adheres to the predefined schedule or target headway, classified as schedule-based or headway-based holding control, respectively \citep{van2019trr}. The latter has been proven to be more suitable for high-frequency bus systems \citep{dai2019partb}. Numerous optimization-based and model-based feedback holding control strategies have been proposed to determine the appropriate holding duration for buses \citep{gkiotsalitis2019tranmetricab,berrebi2018partc,wang2024modellingBusBunching}. These studies demonstrate the capability of bus holding control and provide analytical solutions for its implementation. However, the model-based methods require accurate knowledge of bus propagation and passenger arrival patterns, which are difficult to obtain in real-world applications characterized by fluctuating traffic dynamics \citep{wang2020partc,rodriguez2023partc,lee2022scheduling}. Consequently, model-free methods are gaining traction in developing bus holding control strategies.

Drawing on their broad applicability in traffic control tasks, Reinforcement Learning (RL) methods have been explored for bus holding control, demonstrating promising performance outcomes while enabling the model-free property \citep{geng2023RL_TITS,chen2022partc,yu2023perimeter}. For example, \citet{wang2020partc} utilized a Multi-Agent Reinforcement Learning (MARL) method to derive the real-time bus holding control strategy in a single bus line system. The robustness of the MARL-based bus holding control strategy was further enhanced by \citet{wang2023tits} through the use of implicit quantile networks and meta-learning. \citet{wang2023partc} applied the MARL method to design the holding strategy in a bus system with multiple lines sharing a common corridor. The cooperation between two bus lines was defined and achieved through weight evolution in the reward function. These studies demonstrate that RL-based methods offer significant advantages for formulating bus holding control strategies, particularly for real-time applications in dynamic environments with fluctuating passenger demands. 

Although existing RL-based bus holding control strategies have shown promise, they also exhibit several limitations. First, these strategies are tailored for specific operational scenarios such as single-line or multi-line services, thereby limiting their transferability and adaptiveness\textcolor{black}{/generalization capability (the ability to apply to different scenarios with minimal or no adjustments)} across varying real-world scenarios. In practical control tasks, numerous scenarios can arise due to the varied configurations of real-world bus systems. Therefore, the transferability and adaptiveness of bus holding control strategies are crucial, posing challenges in developing generic strategies capable of handling multiple scenarios \citep{ke2020tits,han2023rl_review}. Second, the evaluation metrics and objectives used in these strategies, such as the overall efficiency of bus systems or the variance in bus headways over an operation period, tend to be delayed and sparse \citep{karnchanachari2020ldc,eschmann2021reward}. RL agents typically receive accurate feedback from the environment only at the end of a task or after a delay following the action execution \citep{yu2023preprint,chan2024dense}. Consequently, decomposing these sparse objectives (i.e., sparse rewards), which do not provide clear signals to evaluate the effectiveness of individual actions, into dense learning signals (i.e., dense rewards) at each step within the RL framework typically necessitates substantial input from human experts and iterative manual adjustments. \citep{cao2024preprint,xie2023preprint}. This challenge is not unique to bus holding control but is also common in many other control scenarios, such as robotic control, potentially compromising the efficacy of RL methods in these real-world applications. Developing more generic and automated designs for RL-based control applications requires further exploration.

The recent advent of large language models (LLMs) has significantly enhanced the AI domain with their substantial capabilities \citep{cao2024preprint}. Leveraging pre-trained world knowledge, LLMs exhibit notable in-context learning and reasoning abilities, surpassing previous AI technologies \citep{yu2023kola,huang2022towards,wei2023larger}. The LLMs can understand and process multi-modal information and incorporate domain-specific knowledge through tools such as plugins, APIs, and expandable models (referred to as LLM-as-agent) or through Supervised Fine-Tuning (SFT) using domain-specific data. The expansive capabilities of LLMs pave the way for potential advancements in Artificial General Intelligence (AGI) \citep{ge2024openagi}. As LLMs surpass human performance in large-scale data analysis and processing speed, they hold the potential to enhance the efficacy of RL in complex control tasks by substituting human expert input with contributions from LLMs. This substitution may also provide an opportunity to increase the automation in the formulation phase of RL-based control method \citep{yang2024harnessing}.

\begin{figure}
\centering
\includegraphics[width=0.6\linewidth]{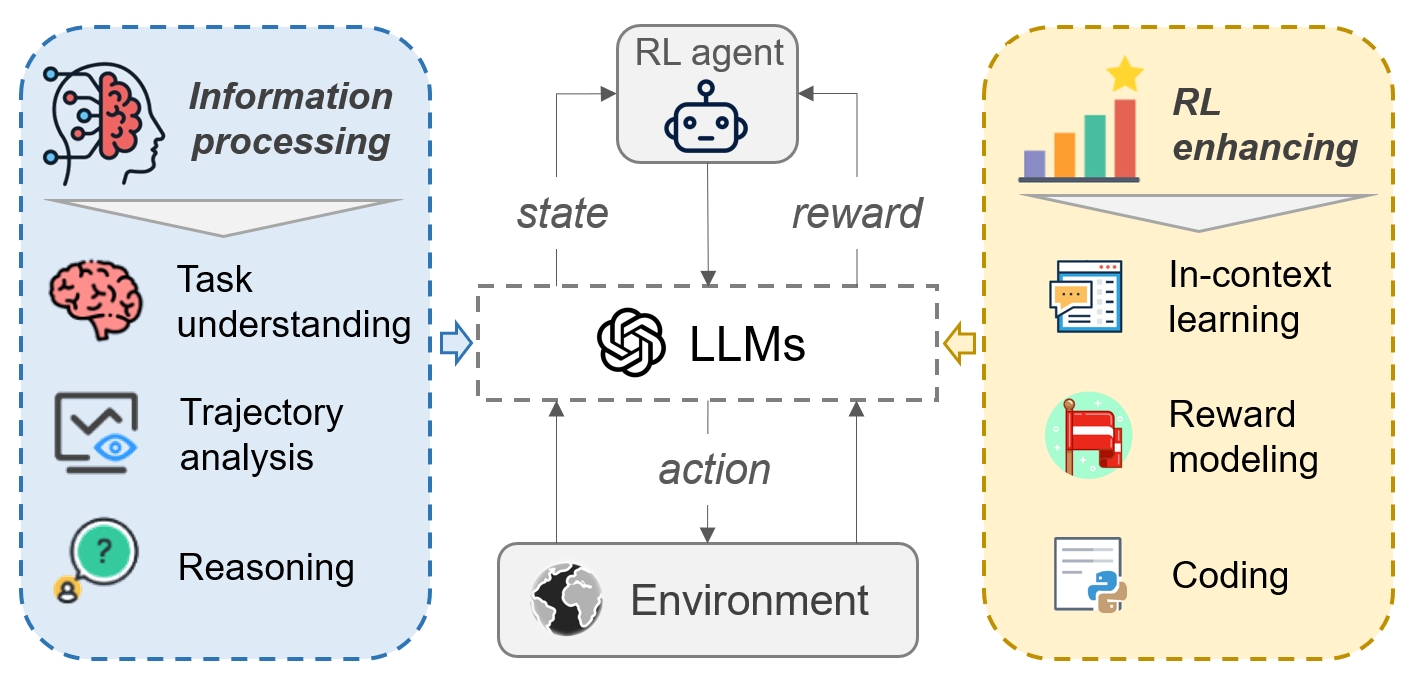}
\caption{\label{fig:LLM_role}LLMs' role in enhancing RL-based control methods}
\end{figure}

To improve the transferability and effectiveness of RL-based control methods in real-world applications, such as bus holding control, LLMs offer significant advantages in information processing and RL performance enhancement \citep{cao2024preprint}. Figure \ref{fig:LLM_role} illustrates the integration and enhancement of RL-based control methods through LLMs. From the perspective of information processing, LLMs are capable of managing complex environmental data and task instructions, extracting essential features from multi-modal observations, and analyzing the reasons for failure experiences in task execution \citep{pang2023natural,adeniji2023language}. Regarding RL performance enhancement, LLMs can improve RL by assisting in the design of the agent, interpreting improvement suggestions from feedback, and generating executable code for reward functions \citep{cao2024preprint}. \citet{yu2023preprint} integrated LLMs into RL-based robotic tasks by defining reward parameters. They prompted the LLM to generate several reward function candidates, selecting the best-performing one for the RL agents, which improved the pass rates of the robotic tasks. \citet{li2024autoMC} designed an iterative framework to automatically initialize and improve the reward functions for an RL-based game player. They prompted the LLM to formulate the Chain of Thought (CoT) to provide interpretable reward functions. With feedback from failure trajectories and reward modifications, the RL agent achieved significant improvements in success rate and learning efficiency in the game. These studies suggest that integrating LLMs in the formulation of RL-based control methods can eliminate the need for expert input and extensive manual trial-and-error, and make the reward function more interpretable \citep{cao2024preprint,li2024autoMC}. However, the interaction patterns between LLMs and RL in these studies overlook the inefficient outputs from LLMs due to the misunderstanding of input prompts, which can disrupt the improvement loops of the reward function and lead to unstable performance for RL agents \citep{zhang2024vpgtrans}. The reliability of LLM-enhanced RL methods requires careful consideration and further refinements in real-world engineering tasks.

To summarize, the limitations and emerging challenges of existing studies on RL-based bus holding control strategies and LLM applications are primarily reflected in the following aspects: 1) designing an appropriate dense reward function for the RL method in control scenarios with sparse and delayed goals is challenging; 2) existing RL-based bus holding control strategies lack adaptiveness and generalization across different bus systems; 3) existing LLM-enhanced RL methods do not ensure stable and reliable performance for the RL agents due to the occasional inefficient output of LLMs.

Therefore, this study develops a novel LLM-enhanced RL paradigm for general control tasks, and we apply the paradigm to various bus holding control scenarios. In this paradigm, the LLM is utilized to generate interpretable reward functions for RL agents. Several LLM-based modules are designed for reward initialization, reward modification, agent performance analysis, and reward refinement. Through the cooperation of these modules, the reward function can be iteratively and automatically improved. The proposed iteration rules ensure stable improvements and reliable performance for RL agents. The capability and adaptiveness of the proposed paradigm across multiple bus holding control scenarios are verified through numerical tests. 

\textcolor{black}{Although well-performing reward functions can be generated based on expert rules or domain knowledge, manually fine-tuning each parameter through trial and error is highly time-consuming. The proposed paradigm is specifically designed to overcome the limitations of manual trial-and-error tuning. Moreover, although some reward settings may appear reasonable to humans, they do not yield satisfactory performance due to the limited interpretability and randomness of RL. Identifying why a seemingly well-designed reward function fails to perform well can be challenging. In contrast, the proposed paradigm eliminates the need for manually diagnosing reward function failures or meticulously accounting for numerous variables. The integrated LLM assists in refining the reward function, ensuring a more reliable and effective design through the proposed workflow.}

The main contributions of this study are summarized below:

\begin{itemize}
\item An LLM-enhanced RL paradigm is designed for general control tasks. This paradigm can automatically initialize and improve RL reward function without requiring human expertise input and manual trial-and-error. The designed iteration rule ensures the reliability and interpretability of the final reward function.
\item The LLM-enhanced RL paradigm is applied to derive bus holding control strategies, demonstrating applicability to various scenarios, including single-line and multi-line systems. The adaptiveness of the proposed paradigm to generic bus holding control scenarios is verified.
\item The proposed bus holding control strategy is tested with both synthetic and real-world bus systems, showing promising performances and robustness compared to vanilla RL, LLM-based, and model-based feedback controllers.
\end{itemize}

The remainder of the paper is organized as follows. Section \ref{section:literature_review} provides the literature review on related studies. Section \ref{section:RL-based_bus_holding_control} formulates the RL-based bus holding control strategy. Section \ref{section:LLM-enhanced_RL_with_reward_generation} exhibits the overall framework of the proposed LLM-enhanced RL paradigm and the function of each module. Section \ref{section:Numerical_tests} describes the settings of three numerical tests and evaluates the performance of the proposed paradigm. The conclusions and perspectives are summarized in Section \ref{section:conclusions}.

\section{Literature review}
\label{section:literature_review}
\subsection{Bus operation control}

To enhance the fidelity of bus propagation simulations, model-based bus holding control strategies incorporate a diverse set of factors, such as bus queueing, overtaking between buses, passenger boarding behaviors, dynamic target headways, real-time bus route state forecasting, multiple bus services, and charging scheduling for electric buses \citep{jinwenlong2017partb,gu2015bus_queue_TS,gkiotsalitis2019tranmetricab,he2020cie,berrebi2018partc,laskaris2021partc,hernandez2015partb,seman2019tits,wang2024modellingBusBunching,lacombe2024integrated}. These dynamic models provide valuable insights for the realistic simulation of bus systems. In existing bus holding control strategies, the holding formats consist of various modes, including single and multiple fix control points, dynamic control points, and speed control \citep{dai2019partb,bian2020partc,berrebi2018partc,ma2021partc,gu2021signal_control_bus_partc}. In addition to these studies in which the control points are bus stops, \citet{chow2021partc} and \citet{yu2023partc} implemented bus holding at traffic signals when the buses encounter stop signals at intersections, using centralized and decentralized RL formulations, respectively. These different control modes help bus holding control to be flexible and adapt to different scenarios.

Recognizing the asynchronous nature of bus holding control, where buses in the same system are not controlled simultaneously and the intervals between consecutive control actions are inconsistent. \citet{wang2021async} and \citet{shen2024preprint_bus} addressed the formulation of asynchronous bus control steps via the MARL with credit assignment schemes achieved through graph neural networks. \citet{rodriguez2023partc} trained the asynchronous agents with discrete-event MARL, utilizing a variable discount factor to scale with the heterogeneous duration between actions.

In addition to bus holding control, bus stop skipping and transit signal priority are also frequently employed to manage bus operations. \citet{rodriguez2023partc} integrated stop-skipping with bus holding control using a MARL approach, enhancing the flexibility of the control scheme in single bus line systems. \citet{huang2021multi} introduced a hybrid operational scheme that merges bus lane reservation with stop-skipping to achieve a consistently high operational speed for buses. Nonetheless, stop-skipping may receive negative feedback from passengers, particularly those who are not served timely at skipped stops. \citet{xu2022MP+TSP} and \citet{long2022TSP+DRL} implemented traffic signal prioritization strategies using the maximum pressure algorithm and deep RL, respectively, aimed at reducing transit delays and managing conflicting bus priority demands. However, when buses are bunched, transit signal priority is less effective as it cannot differentiate priority between two closely spaced buses. Consequently, bus holding control is identified as an effective strategy for managing the issue of bus bunching. Existing methods in the literature, though effective, are specific to particular scenarios and lack transferability and adaptability. This study proposes the development of generic bus holding control strategies that are adaptable to both single and multi-line bus systems, aiming to enhance overall system responsiveness and efficiency.

\subsection{Reward design of RL}

Designing effective reward signals for RL, particularly in the context of sparse rewards, presents significant challenges that can impede the learning process through reliance on undirected exploration \citep{eschmann2021reward}. Addressing this, the design of appropriate reward signals to expedite learning and ensure stable convergence is a fundamental issue \citep{mao2020reward}. \citet{devidze2021explicable} introduced an optimization framework to craft explicable reward functions, which strategically balances the informativeness and sparseness of rewards within a discrete optimization perspective. \citet{eschmann2021reward} highlighted the importance of a feedback loop involving the engineer, where the reward function and the learning algorithm are iteratively refined based on task requirements by expert input. \citet{zhou2022programmatic} developed a probabilistic framework capable of deducing the most suitable programmatic reward function from expert demonstrations, which was experimentally proven to surpass existing baselines in complex settings. \citet{knox2023reward} critically assessed the reward functions reported in the literature, applying eight sanity checks to each and uncovering pervasive flaws in RL reward designs for autonomous driving applications. \citet{mao2020reward} integrated system-wide global rewards with local rewards for individual agents in a MARL-based packet routing scenario, employing a dynamic weighting factor to adjust the focus between local and global rewards during training, thereby enhancing learning efficiency and policy effectiveness.

These methods for reward design generally necessitate explicit definitions of desired behavior or extensive expert demonstrations. However, the advent of LLMs offers novel prospects for more automated reward design. For instance, \citet{kwon2023reward} generated reward signals aligned with user-defined objectives by prompting the LLM with descriptions of desired behaviors, testing the efficacy of these rewards in various scenarios such as the Ultimatum Game, matrix games, and negotiation tasks. \citet{xie2023preprint} utilized LLMs to design dense reward functions for RL-based robotic manipulation and refined these reward functions with human feedback, surpassing the performance of RL agents trained with expert-designed rewards in the majority of tasks and addressing the challenge of language ambiguity. 
This approach enables more effective training of RL agents, closely aligning with specific human objectives without the necessity for extensive labeled data or manually crafted reward functions. 
Consequently, this study leverages LLMs to generate reward functions and evaluate the performance of RL agents. This performance evaluation serves as feedback for LLMs to refine the reward function in subsequent iterations. We design this feedback loop to make the reward design process more autonomous and efficient.

\subsection{LLM-based applications in transportation}

In the transportation domain, LLMs have been deployed as controllers in various applications, including traffic signal control \citep{lai2023tsc,wang2024tsc,da2023tsc}, urban itinerary planning \citep{tang2024llm}, personal mobility generation \citep{wang2024llm}, and urban planning \citep{zhou2024llm}. These applications highlight LLMs' adeptness in scenario understanding and natural language processing. 
Despite their capabilities, LLMs may not be well-perceived for real-time control tasks due to their limited generation speed and the non-interpretability of direct output actions. Specifically, the slow generation speed of LLMs cannot meet the latency requirements of real-time, dense control tasks \citep{nam2024using}. Moreover, the outputs of the LLM-based controller are the control action, instruction, or signal only in most cases, lacking interpretability which is crucial for addressing the black-box problem of AI techniques in practical applications \citep{lin2023can,castelvecchi2016nature}. These limitations hinder the implementation of LLMs on real-time control tasks.

Compared to LLM-based controllers, the integration of LLMs with RL can mitigate the impact of slow responses of LLMs through the intermediate interface of offline reward function generation, facilitating the practical application of LLMs in real-time control tasks \citep{carta2023grounding}. For instance, \citet{han2024LLM_RL} leveraged LLMs to generate and evolve RL reward functions through iterative feedback for autonomous driving in highway scenarios. They applied a similar design of reward function candidates to \citet{yu2023preprint} in each iteration, potentially boosting reward improvement efficiency. \citet{pang2024illm} combined LLMs with RL in traffic signal control, where LLMs assess and adjust the RL agent's decisions to ensure their reasonableness. 

These studies highlight the benefits of integrating LLMs with RL to enhance RL-based control systems through various methods of integration. However, the reliability of the responses from LLMs is not taken into account in these studies. Therefore, this study proposes a rule for the feedback loop within the LLM-enhanced RL paradigm. This rule is designed to filter out inefficient outputs from LLMs, thereby ensuring a reliable and stable method for formulating the bus holding control strategies.

\section{RL-based bus holding control for generic scenarios}
\label{section:RL-based_bus_holding_control}

\subsection{Problem statement}

In bus holding control, early buses require holding control at control points (e.g., bus stops and the departure area) to homogenize the time headways in the system and reduce the average waiting time for passengers \citep{xuan2011dynamic_bus_holding}. However, unreasonable holding control can lead to excessive delays for onboard passengers, resulting in conflicting objectives of balancing bus headways and minimizing passengers' travel time. In multi-line services, the interaction between different bus lines must also be considered. In RL-based control methods, integrating these diverse objectives requires careful design of the dense rewards at each control step. 

\textcolor{black}{Furthermore, the ultimate objective of bus holding control is to minimize the total/average waiting time and travel time of all passengers. However, the waiting time and travel time are only observed when passengers board the bus and alight at their destinations, respectively. The holding action influences the waiting time of passengers at subsequent stops and the travel time of onboard passengers, but it is executed before these passengers board or alight. Consequently, the reward associated with a holding action is only received after the relevant boarding and alighting events occur, leading to a delayed reward. Moreover, if no passengers board or alight at a particular stop, no waiting time or travel time reward is generated for that stop, resulting in a sparse reward signal. 
Therefore, providing denser and more immediate rewards is essential. The dense rewards provide a more informative and accessible learning signal for RL agents, helping them better balance short-term actions with long-term objectives.}

The control action in bus holding control is taken every time a bus arrives at the control point, i.e., the bus stop in this study, and the travel time of buses between two adjacent stops is inconsistent. Thus, bus holding control is asynchronous control in which RL agents do not act simultaneously, and the default MARL method cannot tackle the asynchronous control when considering the cooperation among agents \citep{wang2021async,shen2024preprint_bus}. 
In this study, as we focus on the LLM-enhanced reward design in the RL method, the asynchronous issue in bus holding control is addressed in a simplified way: through control action discretion (to homogenize the duration of decision steps) and knowledge sharing among agents (to stabilize learning in the decentralized control system \citep{li2021sharing_knowledge}). 

In the RL method of this study, each bus stop is represented by an RL agent. The RL agents are decentralized and observe only partial information around their respective controlled stops from the environment. Thus, the agents within the same bus system are considered homogeneous \textcolor{black}{(sharing identical configurations of observation, state, and action)}, allowing them to share experiential knowledge from their memory buffers during training to enhance exploration efficiency \citep{wang2020partc}. Sharing experiential knowledge also improves cooperation among agents and stabilizes training in the decentralized system \citep{li2021sharing_knowledge}. During testing, the learned policy parameters can be shared among agents in the same system due to their homogeneity. Given the strong correlation and consistent performance between bus time headway and space headway \citep{yu2023partc,ampountolas2021tits}, space headway is utilized as the direct observation for the RL agents. The long holding duration is discretized into the combination of a series of homogeneous smaller actions (e.g., 5 seconds). Buses traveling between two bus stops do not participate in the control process, thus their states are not considered by agents or recorded in the memory buffer. Real-time space headway information is continuously available, and the discretization of action steps standardizes the duration of decision steps. Then, the design of the bus holding control strategy is transformed into a discrete-event task with a uniform decision step duration, making it amenable to formulation with the RL method.

\subsection{Design of the RL agent}
\label{section:RL_design}

To implement bus holding control with the RL method, the state, action, and reward need to be designed to follow the Markov decision process \citep{sutton2018reinforcement}. Every time a bus dwells at a stop and completes the passenger boarding and alighting process, the corresponding agent needs to decide whether to hold the bus for the next control step (i.e., 5 seconds in this study). If the action is to hold the bus, the agent will continue making decisions at the end of each holding period until either a no-holding signal is received or the predefined maximum holding time is reached. Thus, the \textbf{action} of the agents is a binary variable representing \textit{to hold} or \textit{not to hold} the bus at the next action step (i.e., the next control step). If the agent chooses not to hold the bus, the bus will depart from the stop and proceed to the next stop. During the bus’s travel between stops, no action is taken regarding the bus.

The \textbf{state} of the agent is an array comprising a set of observations from the environment, including the space headways, the number of onboard passengers, and holding duration information of the current bus dwelling at a stop. We formulate the agents' state in a generic way that is not specific to any particular scenario. Given the set of all bus lines in a bus system as $\bm{M}$, the state of the current dwelling bus $b$ at stop $i$ at time step $t$, $S_{i,b,t}$, is defined as in Eqs. (\ref{eq:state_all}) and (\ref{eq:state_headway}). 
\begin{align}
    & S_{i,b,t} = S^{\text{headway}}_{i,b,t} \cup \left[ n^{\text{onboard}}_{i,b,t},t^{\text{holding}}_{i,b,t}\right]  \label{eq:state_all},\\
    & S^{\text{headway}}_{i,b,t} = \left[h^{m+}_{i,b,t},h^{m-}_{i,b,t}\right], \forall m \in \bm{M} \label{eq:state_headway} ,
\end{align}
where $S^{\text{headway}}_{i,b,t}$ is the set of space headways of the current dwelling bus $b$ at stop $i$ at time step $t$; $h^{m+}_{i,b,t}$ and $h^{m-}_{i,b,t}$ are the forward and backward space headways between the current dwelling bus $b$ and its nearest forward and backward buses from bus line $m$ at time step $t$, respectively; $n^{\text{onboard}}_{i,b,t}$ is the number of on-board passengers of the current dwelling bus $b$ at stop $i$ at time step $t$; $t^{\text{holding}}_{i,b,t}$ is the holding time the current dwelling bus $b$ has already spent at stop $i$ at time step $t$.
When multiple buses are dwelling at the same stop simultaneously, the state of each bus is recorded separately, and the agent corresponding to this stop is duplicated into several instant versions to make decisions for each bus individually. The exploration and exploitation transitions of these instant agents are all updated to a common memory buffer for knowledge sharing.

Regarding single bus line systems, $S^{\text{headway}}_{i,b,t}$ only contains one pair of headways between buses within the same line. For multi-line systems, the headways between buses from all different bus lines also need to be included in the state. Thus, there are $|\bm{M}|$ pairs of headways in $S^{\text{headway}}_{i,b,t}$ for multi-line services, where $|\bm{M}|$ is the number of bus lines in the entire system. It is worth noting that not all bus stops in a multi-line system serve all bus lines. As depicted in the right part (white-background box) of Fig. \ref{fig:framework}, bus stops located on branches outside the shared corridor only serve one bus line. In these cases, the forward and backward space headways between the current dwelling bus and buses from different bus lines are set to 0. This setting for headways is described by Eqs. (\ref{eq:headways+}) and (\ref{eq:headways-}).
\begin{align}
    h^{m+}_{i,b,t} &= \begin{cases} 
    x^{m+}_{i,b,t} - x_{i,b,t} & m \in \bm{M}_i \\
    0 & m \notin \bm{M}_i \\
    \end{cases},
    \label{eq:headways+} \\
    h^{m-}_{i,b,t} &= \begin{cases} 
    x_{i,b,t} - x^{m-}_{i,b,t} & m \in \bm{M}_i \\
    0 & m \notin \bm{M}_i \\
    \end{cases},
    \label{eq:headways-}
\end{align}
where the $x_{i,b,t}$ is the travel distance of the current dwelling bus of stop $i$ along its bus route; $x^{m+}_{i,b,t}$ and $x^{m-}_{i,b,t}$ are the travel distances of the nearest forward and backward buses of the current bus $b$ from bus line $m$ along the shared bus route with standardized origins; $\bm{M}_i$ is the set of bus lines that bus stop $i$ serves. This formulation allows for a general state representation of heterogeneous bus stops in multi-line systems.

As the major contribution of this paper, the \textbf{reward} is not defined manually, instead, it is generated by LLMs. All agents within the same bus system share the same reward function. Although agents controlling bus stops located in the branch and the shared corridor may have different objectives in the multi-line service, we use homogeneous agents for control to evaluate whether LLMs can provide a generic reward function for such heterogeneous scenarios. The potential success in designing a generic reward function may enhance the adaptiveness and transferability of RL-based bus holding control strategies. 

\subsection{Modeling dynamics of bus propagation}

A dynamic simulation environment of bus systems that allows overtaking is built considering the passenger modeling and bus capacities \citep{jinwenlong2017partb}. Assuming that passengers' arrivals follow Poisson processes \citep{berrebi2018partc,wang2020partc}, the \textcolor{black}{travel information of} passenger $p$ traveling with bus line $m$ and originating from stop $i$, \textcolor{black}{denoted as $P_{i,m,p}$,} is defined as in Eq. (\ref{eq:passenger}).
\begin{equation}
    P_{i,m,p} = \left[t^{\text{arrive}}_{i,m,p},I^{\text{alight}}_{i,m,p},t^{\text{board}}_{i,m,p},t^{\text{alight}}_{i,m,p}\right],
    \label{eq:passenger}
\end{equation}
where $t^{\text{arrive}}_{i,m,p}$ is the arrival time of passenger $p$ at stop $i$ traveling with bus line $m$; $I^{\text{alight}}_{i,m,p}$ is the alighting stop ID of passenger $p$; $t^{\text{board}}_{i,m,p}$ and $t^{\text{alight}}_{i,m,p}$ are the boarding and alighting times of passenger $p$, respectively, which are updated during the simulation. Then, the total waiting time, $\bar{T}^{\text{wait}}$, and total travel time, $\bar{T}^{\text{travel}}$, for all passengers are derived from Eqs. (\ref{eq:passenger_wait_time}) and (\ref{eq:passenger_travel_time}) to evaluate the control strategies.
\begin{align}
    \bar{T}^{\text{wait}} &= \sum_{m \in \bm{M}} \sum_{i \in \bm{I}_m} \sum_{p \in \bm{P}_{m,i}} T^{\text{wait}}_{i,m,p} = \sum_{m \in \bm{M}} \sum_{i \in \bm{I}_m} \sum_{p \in \bm{P}_{m,i}} \left(t^{\text{board}}_{i,m,p} - t^{\text{arrive}}_{i,m,p} \right) \label{eq:passenger_wait_time},\\
    \bar{T}^{\text{travel}} &= \sum_{m \in \bm{M}} \sum_{i \in \bm{I}_m} \sum_{p \in \bm{P}_{m,i}} T^{\text{travel}}_{i,m,p} = \sum_{m \in \bm{M}} \sum_{i \in \bm{I}_m} \sum_{p \in \bm{P}_{m,i}} \left(t^{\text{alight}}_{i,m,p} - t^{\text{arrive}}_{i,m,p} \right) \label{eq:passenger_travel_time},
\end{align}
where $\bm{I}_m$ is the set of all bus stops within bus line $m$; $\bm{P}_{m,i}$ is the set of all passengers traveling with bus line $m$ and originating from stop $i$; $T^{\text{wait}}_{i,m,p}$ and $T^{\text{travel}}_{i,m,p}$ are the waiting and travel time of passenger $p$ traveling with bus line $m$ and originating from stop $i$. In multi-line systems, passengers whose origin and destination stops are both shared among several lines are categorized as shared passengers. Shared passengers can board buses from any line that reaches their destinations. The specific bus line that a shared passenger belongs to is determined based on the bus line they actually board during the simulation. 

The dwell time for bus $b$ at stop $i$ with arrival time step $t$, denoted as $D_{i,b,t}$, is the maximum boarding time and alighting time of passengers, computed as in Eq. (\ref{eq:dwell_time}).
\begin{equation}
    D_{i,b,t} = \max \left[\alpha_b B_{i,b,t}, \alpha_a A_{i,b,t} \right],
    \label{eq:dwell_time}
\end{equation}
where $B_{i,b,t}$ and $A_{i,b,t}$ are the number of boarding and alighting passengers at stop $i$ for bus $b$ with the bus arrival time step $t$, respectively; $\alpha_b$ and $\alpha_a$ are the time taken for each passenger to board and alight from the bus, respectively.

The number of boarding passengers, $B_{i,b,t}$, is determined by the bus's capacity limit and the number of waiting passengers at the stop, as given in Eq. (\ref{eq:num_board_passenger}). For bus $b$ belongs to bus line $m$, the number of waiting passengers for bus line $m$ at stop $i$, denoted $B^\prime_{i,m,t}$, is derived from the set of passengers whose origins are at stop $i$ and also includes hold-up passengers who could not board the previous bus due to capacity constraints, as shown in Eq. (\ref{eq:num_wait_passenger}). The number of alighting passengers, $A_{i,b,t}$, is obtained from the set of onboard passengers of bus $b$, as depicted in Eq. (\ref{eq:num_alight_passenger}).
\begin{align}
    B_{i,b \in \bm{b}_m,t} &= \min \left[B^\prime_{i,m,t}, c - n^{\text{onboard}}_{b,t} + A_{i,b,t} \right] \label{eq:num_board_passenger},\\
    B^\prime_{i,m,t} &= \left|\left[P_{i,m,p}\right]_{\forall p \text{ if } t^- < t^{\text{arrive}}_{i,m,p} \leq t} \right| + n^{\text{holdup}}_{i,m,t^-} \label{eq:num_wait_passenger},\\
    n^{\text{holdup}}_{i,m,t^-} &= \max \left[0, B^\prime_{i,m,t^-} - \left(c - n^{\text{onboard}}_{b,t^-} + A_{i,b,t^-} \right) \right] \label{eq:num_holdup_passenger},\\
    A_{i,b,t} &= \left|\left[P^{\text{onboard}}_{b,p}\right]_{\forall p \text{ if } I^{\text{alight}}_{b,p} = i} \right| \label{eq:num_alight_passenger},
\end{align}
where $\bm{b}_m$ is the set of buses operating on bus line $m$; $c$ is the bus capacity; $n^{\text{onboard}}_{b,t}$ is the number of onboard passengers on bus $b$ at time step $t$, and $n^{\text{onboard}}_{i,t}$ refers to the same quantity when bus $b$ dwells at stop $i$ at time $t$; $|\cdot|$ denotes the cardinality of the inside set; $t^-$ is the arrival time of the previous bus from line $m$ at the stop $i$; $n^{\text{holdup}}_{i,m,t^-}$ is the number of passengers who could not board the previous bus and are held up at the stop $i$ due to capacity constraints; $P^{\text{onboard}}_{b,p} = \left[t^{\text{arrive}}_{b,p},I^{\text{alight}}_{b,p},t^{\text{board}}_{b,p},t^{\text{alight}}_{b,p} \right]$ refers to $P_{i,m,p}=\left[t^{\text{arrive}}_{i,m,p},I^{\text{alight}}_{i,m,p},t^{\text{board}}_{i,m,p},t^{\text{alight}}_{i,m,p}\right]$ for passenger $p$ represented by $P_{i,m,p}$ who have boarded bus $b$. This formulation of $P^{\text{onboard}}_{b,p}$ is intended to simulate the state of passengers onboard the buses. The set of onboard passengers for each bus is updated with the simulation by appending boarding passengers and removing those alighting. 

The bus propagation for bus $b$, updated every second, is calculated using Eq. (\ref{eq:bus_travel}).
\begin{align}
    x_{b \in \bm{b}_m,t} = \begin{cases} 
    x_{b,t-1} + v_{b,t} & x_{b,t-1} \notin \bm{x}^s_m \\
    x_{b,t-1} + 0 & x_{b,t-1} \in \bm{x}^s_m \text{ and } t \leq a_{b,i} + D_{i,b,a_{b,i}} + g_{i,b} \\
    x_{b,t-1} + v_{b,t} & x_{b,t-1} \in \bm{x}^s_m \text{ and } t > a_{b,i} + D_{i,b,a_{b,i}} + g_{i,b}
    \end{cases},
    \label{eq:bus_travel} 
\end{align}
where $x_{b,t}$ is the travel distance of bus $b$ along its route at time step $t$, with $x_{b,t} = x_{i,t}$ when bus $b$ dwells at stop $i$; $v_{b,t}$ is the average bus speed from time step $t-1$ to $t$; $\bm{x}^s_m$ is the set of distances corresponding to all stops on the bus line $m$; when bus dwells, assuming the current dwelling stop ID of bus $b$ is $i$, $a_{b,i}$ is the arrival time of bus $b$ at stop $i$; $g_{i,b}$ is the holding time of this dwelling bus $b$ at stop $i$, determined by the RL agents. The average value and standard deviation of time headways within each bus line can be calculated from the comprehensive observation of bus trajectories, which can be used for evaluating the performances of control strategies.

Complex traffic conditions (e.g., congestion and traffic signals) are captured through the travel time between consecutive bus stops, which directly affects the bus speed $v_{b,t}$. Between two bus stops, buses are assumed to travel at the average speed derived from the observed travel time, as defined in Eq. (\ref{eq:bus_speed}).

\begin{equation}
    v_{b,t} = \frac{x_m^s(i+1) - x_m^s(i)}{T_{b}^{\text{travel}}(i)} \quad \text{if} \quad x_m^s(i) \leq x_{b,t-1} < x_m^s(i+1),
    \label{eq:bus_speed}
\end{equation}
where $x_m^s(i)$ denotes the distance of bus stop $i$ from the origin of bus line $m$, and $T_{b}^{\text{travel}}(i)$  represents the travel time of bus $b$ between stop $i$ and $i+1$. Traffic congestion or delays at traffic signals between stops $i$ and $i+1$ increase the travel time, resulting in a reduced bus speed.

\subsection{Action-value function approximation}

In this study, the Deep Q-Network (DQN) is employed to approximate the action-value function for the bus holding control agents \citep{mnih2015DQN}. The agents aim to interact with the environment by selecting actions that maximize future rewards \citep{mnih2013DQN}. To differentiate between the time step $t$ (i.e., per second) in the environment simulation and the decision step (i.e., 5 seconds in this study) in the RL method, we denote the current decision step of RL agents as $\hat{t}$. The future discounted return, $R_{\hat{t}} = \sum_{\hat{t}^\prime = \hat{t}}^{\hat{T}} \gamma^{\hat{t}^\prime - \hat{t}}r_{\hat{t}^\prime}$, is defined as the discounted sum of future rewards, where $\hat{T}$ is the maximum number of decision steps within an episode, $\gamma$ is the discounted factor, and $r_{\hat{t}^\prime}$ is the reward of the single decision step $\hat{t}^\prime$. The optimal action-value function $Q^\ast (s,a)$ is the maximum expected return achieved by selecting an action $a$ with the state $s$, as described in Eq. (\ref{eq:max_Q}).
\begin{equation}
    Q^\ast (s,a) = \max_\pi \mathbb{E} \left[R_{\hat{t}}|s_{\hat{t}}=s,a_{\hat{t}}=a,\pi \right],
    \label{eq:max_Q} 
\end{equation}
where $s_{\hat{t}}$ and $a_{\hat{t}}$ are the state and selected action at decision step $\hat{t}$, respectively; $\pi = P(a|s)$ is the behavior policy of the agents.


The neural network, with weights $\theta$, is trained by minimizing the loss function $L_k(\theta_k)$ at each learning iteration $k$, as shown in Eq. (\ref{eq:NN_update}).
\begin{equation}
    L_k(\theta_k) = \mathbb{E}_{(s,a,r,s^\prime) \sim \text{U}(\mathcal{D})} \left[\left(r + \gamma \max_{a^\prime} Q(s^\prime,a^\prime;\theta_{k-1}) - Q(s,a;\theta_k) \right)^2 \right],
    \label{eq:NN_update} 
\end{equation}
where $\text{U}(\mathcal{D})$ denotes drawing samples of experience uniformly at random from the pool of stored samples $\mathcal{D}$ (i.e., the shared memory buffer in this study); $s^\prime$ represents the state of next decision step; $a^\prime$ denotes the possible actions for the next decision step.

The training process with knowledge sharing among agents in this study is summarized in Algorithm \ref{algorithm_DQN}. Since the agents in the same bus system are considered homogeneous, only one set of parameters of the Q-network is trained using the algorithm. All agents share this set of parameters for decision-making.

\begin{algorithm}
\caption{\label{algorithm_DQN}DQN with knowledge sharing \citep{mnih2013DQN,li2021sharing_knowledge}}
\begin{algorithmic}
\Require Total training episodes ($\mathcal{K}$), maximum simulation time in each episode ($T$), total training epochs with each episode ($\mathcal{P}$), number of bus stops in the environment ($\mathcal{I}$), the duration of each action step ($\hat{a}$), Q-network function $Q$, minibatch
\Ensure Final parameters of Q-network ($\theta$)
\State Initialize shared replay memory ($\mathcal{D}$), parameters of Q-network ($\theta$)
\For{$\text{episode} = 1 \text{ to } \mathcal{K}$}
    \For{$t = 1 \text{ to } T$}
        \For{$i = 1 \text{ to } \mathcal{I}$}         
            \If{bus $b$ arrives stop $i$ and rest dwell time = 0}
                \State Retrieve state $s_{i,b,t}$
                \State With probability $\epsilon$ select a random action $a_{i,b,t}$
                \State otherwise duplicate an instant version of Q-network $Q^{\text{instant}}$ 
                \State and select $a_{i,b,t} = {\arg \max}_a Q^{\text{instant}}(s_{i,b,t},\bm{a};\theta) $
                \State Store bus $b$ to set $\bm{b}_{i,t}$
            \EndIf
            \For{bus $b^-$ in $\bm{b}_{i,t-\hat{a}}$}
                \State Retrieve state $s_{i,b^-,t}$ and reward $r_{i,b^-,t-\hat{a}}$
                \State Match the state and reward with the previous state and action when $t = t-\hat{a}$
                \State Store transition $\left(s_{i,b^-,t-\hat{a}}, a_{i,b^-,t-\hat{a}}, r_{i,b^-,t-\hat{a}}, s_{i,b^-,t} \right)$ in the shared replay memory $\mathcal{D}$
            \EndFor
        \EndFor
        \State Update environment for all buses according to Eq. (\ref{eq:bus_travel})
    \EndFor
    \For{$\text{epoch} = 1 \text{ to } \mathcal{P}$}
        \State Sample random minibatch of transitions $\left(s_{\cdot,\cdot,j-\hat{a}},a_{\cdot,\cdot,j-\hat{a}},r_{\cdot,\cdot,j-\hat{a}},s_{\cdot,\cdot,j} \right)$ from $\mathcal{D}$
        \State Update $\theta$ by performing a gradient descent step according to Eq. (\ref{eq:NN_update})
    \EndFor
\EndFor
\end{algorithmic}
\end{algorithm}

\section{LLM-enhanced RL with reward generation}
\label{section:LLM-enhanced_RL_with_reward_generation}

\subsection{Overview of LLM-enhanced RL paradigm}

To circumvent the extensive manual trial-and-error process traditionally needed to develop effective dense reward functions for RL, we utilize LLMs to generate and iteratively improve reward functions for RL-based control tasks automatically. This LLM-enhanced RL paradigm is capable of improving the generalization ability of existing RL-based methods for generic bus holding control strategies, as it utilizes LLMs (regarded as AGI) to replace expert input, thereby enhancing the agent's adaptability and efficiency. When designing the prompts, the \textbf{principle} is to provide only the essential information regarding the task and the specific role of the LLM with precision. 

\begin{figure}
\centering
\includegraphics[width=1.00\linewidth]{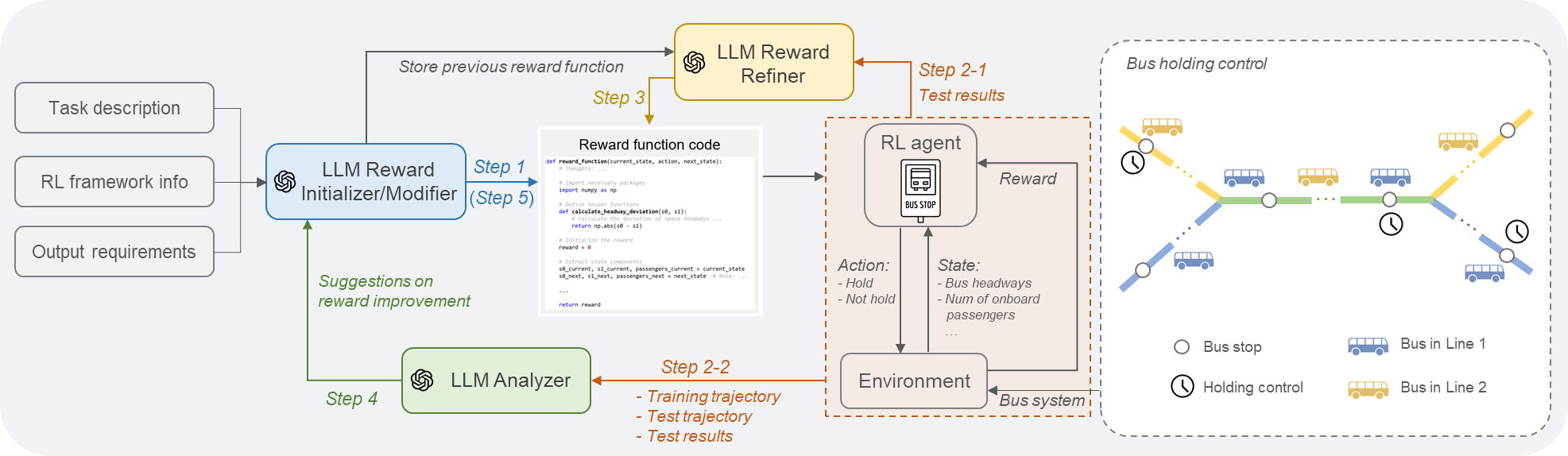}
\caption{\label{fig:framework}\textcolor{black}{LLM-enhanced RL paradigm}}
\end{figure}

The proposed LLM-enhanced RL reward generator consists of four LLM-based modules (reward initializer, reward modifier, agent performance analyzer, and reward refiner), the RL agent, and the bus holding control environment. The interactions among these components are depicted in Fig. \ref{fig:framework}. The \textit{reward initializer} and \textit{modifier} are designed for generating and improving the reward function. The \textit{agent performance analyzer} evaluates agents' performance and provides improvement suggestions for the reward function. Following \citet{li2024autoMC}, we separate the coding and data-analyzing modules. This division reduces the task complexity of each module and enhances the response efficiency of LLMs. However, random or incorrect outputs from LLMs are inevitable even with accurate prompts, resulting in uncertain performances for agents trained with the modified reward function \citep{zhang2024vpgtrans,liu2022meta}. Therefore, the \textit{reward refiner} is designed to refine the modified reward function when test performances degrade, ensuring that only the reward functions that provide improved or similar test performance compared to the previous iteration are used in subsequent iterations. This paradigm enables the initialization and stable improvement of the reward function through iterations.

The workflow of the proposed paradigm is summarized as follows:

\begin{itemize}
\item \textbf{Step 1} (\textbf{Reward initializer}): The LLM initializes an executable reward function based on the input prompt, which describes the information about the bus holding control strategy, the RL method, and output requirements. 
\item \textbf{Step 2} (\textbf{RL agent and environment}): Using the provided reward function, the RL agents are trained and tested in the environment. The training and test results are recorded.
\item \textbf{Step 3} (\textbf{Reward refiner}): \textit{This module is activated only when it is not the first iteration and the agents trained with the current modified reward function do not achieve better or similar test performance compared to the previous iteration.} The LLM refines the reward function repeatedly based on the previous version, using the knowledge of the failure experience from the current reward function, until a better test performance is obtained.
\item \textbf{Step 4} (\textbf{Agent performance analyzer}): The LLM-based agent performance analyzer provides suggestions for improving the reward function based on the training and test results.
\item \textbf{Step 5} (\textbf{Reward modifier}): The LLM modifies the current reward function considering the improvement suggestions. The process then returns to \textbf{Step 2}.
\end{itemize}

The workflow terminates when it reaches the maximum number of iterations, the test performance converges, or the test performance reaches the expected performance. The settings for each LLM-based module are introduced below.

\subsection{Reward initializer and modifier}

The reward initializer and modifier are designed to generate executable reward functions for RL agents based on the input prompts. To achieve the required reward output, the input prompts for these two modules include the role setting for the LLM system, task description, basic definition of bus holding control, agent definition, input parameters of the reward function, environment information, output requirements, and output format. Additionally, the prompt for the reward modifier includes the current reward function and improvement suggestions. Each component is briefly introduced below, and the complete prompt examples can be found in \ref{appendix:reward_initializer} and \ref{appendix:reward_modifier}.

\textbf{Role setting for the LLM system and task background:} This component defines the role of the LLM as a reward designer for RL agents and provides a general introduction to the bus holding control. 

\textbf{Task description and basic definitions:} This component clarifies the control objectives and basic terms related to bus holding control.

\textbf{Agent definitions:} The state space, action space, and attributes of the agents are listed within this component.

\textbf{Input parameters of the reward function:} This component specifies the parameters that the LLM can utilize when formulating the reward functions.

\textbf{Environment information:} Basic information about the simulation environment is introduced in this component.

\textbf{Output requirements and format:} This component defines the format of the reward function and requires the LLM to formulate the reward function using only the input parameters. It also asks the LLM to provide the thoughts and reasons for the reward design in the first part of the function and explain the objective of each line of code. This approach helps the LLM to formulate a CoT to enhance the accuracy and interpretability of the generated reward functions \citep{wei2022CoT}.

\textbf{Current reward function and improvement suggestions:} \textit{This component is only for the reward modifier.} It includes the current reward function and the improvement suggestions from the agent performance analyzer, aiding the reward modifier in improving the reward function.

With these components in the prompt, the LLM can understand the task and formulate its CoT for the output reward function. The initialization and modification of the reward function can be expressed as Eqs. (\ref{eq:reward_initializer}) and (\ref{eq:reward_modifier}).
\begin{align}
    R_0 &= \text{RewardInitializer}(Task, Env, \hat{R}) \label{eq:reward_initializer},\\
    R_{\textcolor{black}{\mathcal{I}}} &= \text{RewardModifier}(R_{\textcolor{black}{\mathcal{I}}-1}, Sugg_{\textcolor{black}{\mathcal{I}}-1}, Task, Env, \hat{R}) \label{eq:reward_modifier},
\end{align}
where $R_0$ represents the initialized reward function, and $R_{\textcolor{black}{\mathcal{I}}}$ is the reward function derived from the reward modifier at the ${\textcolor{black}{\mathcal{I}}}$-th iteration; The functions $\text{RewardInitializer}(\cdot)$ and $\text{RewardModifier}(\cdot)$ are \textcolor{black}{prompted LLMs utilized by} the \textit{reward initializer} and \textit{modifier}, respectively; $Task$ and $Env$ denote the control task and RL environment information; $\hat{R}$ refers to the reward requirements, including input parameters, output requirements, and the desired format for the reward function; $Sugg_{\textcolor{black}{\mathcal{I}}}$ represents the improvement suggestions to the reward function provided by the agent performance analyzer module at the ${\textcolor{black}{\mathcal{I}}}$-th iteration.

\subsection{Agent performance analyzer}

While the RL agent is trained with the provided reward function, the total reward evolution along episodes is recorded for training performance evaluation. Subsequently, the parameters of the trained agent are inherited by each agent in the bus system to operate a test episode. The agents' decision trajectories during the test and final result indicators are also recorded for evaluation. Using the training reward evolution, test trajectories, and test results, the LLM is tasked with analyzing the training and test performances of the RL agents. Notably, the agent's decision trajectories consist of the agent's observations and the actions taken at each step, which differs from the spatial trajectory of the bus. The decision trajectory captures the sequence of decisions made by the agent based on its perception of the environment.

The input prompt of the analyzer partially consists of the role setting for the LLM system, task description, basic definitions of bus holding control, agent definitions, input parameters of the reward function, and environment information, which are the same as those in the reward initializer and modifier modules. The remaining components of the input prompt include agents' training and test data, format descriptions of these data, and output requirements. The complete prompt examples can be found in \ref{appendix:analyzer}.

\textbf{Training and test data:} The total reward evolution during training, agents' decision trajectories (truncated to the last $L = 50$ steps if the action steps exceed 50 \citep{li2024autoMC}), and final test results are encoded into JSON format as an input component of the prompt.

\textbf{Format descriptions of input data:} This component explains the definitions of each item in the input data (i.e., training and test data) to help the LLM better understand and parse the data.

\textbf{Output requirements:} This module requires the LLM to analyze the input data from the training and test performances separately, and then provide suggestions on reward improvement. The LLM is also asked to formulate a CoT that can provide a more accurate and deeper evaluation of agents' performances \citep{wei2022CoT}.

These components enable the LLM to depict the control environment and the performance of RL agents effectively. The agent performance analyzer focuses on processing and analyzing the data without dealing with code understanding and generation. The division of data analysis and code generation simplifies the tasks for both the analyzer and reward modifier modules. The improvement suggestions for the reward function are generated as described in Eq. (\ref{eq:analyzer}).
\begin{align}
    Sugg_{\textcolor{black}{\mathcal{I}}} = \text{Analyzer}(Evol_{\textcolor{black}{\mathcal{I}}}^{\text{train}}, Traj_{\textcolor{black}{\mathcal{I}}}^{\text{test}}, Rslt_{\textcolor{black}{\mathcal{I}}}) \label{eq:analyzer},
\end{align}
where the function $\text{Analyzer}(\cdot)$ is the \textcolor{black}{prompted LLM utilized by} the \textit{agent performance analyzer}; $Evol_{\textcolor{black}{\mathcal{I}}}^{\text{train}}$, $Traj_{\textcolor{black}{\mathcal{I}}}^{\text{test}}$, $Rslt_{\textcolor{black}{\mathcal{I}}}$ correspond to the total reward evolution of the RL agent during training, the agent's test trajectories, and test result indicators, respectively, which are derived from Eqs. (\ref{eq:train_agent}) and (\ref{eq:test_agent}).
\begin{align}
    & Evol_{\textcolor{black}{\mathcal{I}}}^{\text{train}}, A_{\textcolor{black}{\mathcal{I}}} = \text{TrainAgent}(A_{\text{init}}, R_{\textcolor{black}{\mathcal{I}}}, Env) \label{eq:train_agent},\\
    & Traj_{\textcolor{black}{\mathcal{I}}}^{\text{test}}, Rslt_{\textcolor{black}{\mathcal{I}}} = \text{TestAgent}(A_{\textcolor{black}{\mathcal{I}}}, Env) \label{eq:test_agent},
\end{align}
where the functions $\text{TrainAgent}(\cdot)$ and $\text{TestAgent}(\cdot)$ represent the training and test algorithms of the RL agent; $A_{\text{init}}$ denotes the initialized structure of the neural network for the RL agent; $A_{\textcolor{black}{\mathcal{I}}}$ is the trained agent at the ${\textcolor{black}{\mathcal{I}}}$-th iteration.

\subsection{Reward refiner}

As illustrated in Fig. \ref{fig:framework}, the reward modifier and analyzer have formed the closed loop of iterations with the RL agents and environment. However, the modified reward function cannot guarantee a better performance for RL agents compared to their previous performance due to the inevitably inefficient generation from LLMs \citep{zhang2024vpgtrans}. The reward refiner is designed to further stabilize the improvement of reward functions over iterations by performing a performance check on every modified reward function. The complete prompt examples can be found in \ref{appendix:reward_refiner}.

The reward refiner evaluates the test results of RL agents learned with the modified reward function and is activated when the test results fail to meet the refiner's criterion $p_r[\cdot]$, where $\cdot$ represents the defined performance evaluation metric. The refiner's criterion ensures that the test performance does not significantly deteriorate compared to previous results. For example, if total travel time ($TTT$) is used as the performance metric, the refiner's criterion can be expressed as $p_r[TTT_\text{prev}]=s_1 \times TTT_\text{prev}$ or $p_r[TTT_\text{prev}]=TTT_\text{prev}+s_2$, where $TTT_\text{prev}$ denotes the previous $TTT$, and $s_1$ and $s_2$ are the slack parameters slightly greater than 1 and 0, respectively. If the current $TTT$ exceeds $p_r[TTT_\text{prev}]$, it indicates a significant performance decline compared to the previous $TTT$, which is unacceptable for reward function improvement. Consequently, the reward refiner is activated to refine the previous reward function using insights gained from the failure of the current reward function. The refined reward function is then retested and re-evaluated against the refiner's criterion. This refinement continues repeatedly until the refiner's criterion is satisfied by a reward function. Once a satisfactory reward function is identified, it is incorporated into the workflow for the next iteration. The slack parameter in the refiner's criterion can be adjusted dynamically throughout the iterations to impose increasingly stringent requirements, thereby facilitating the convergence of test performance. 

In addition to the components shared with the reward modifier, the \textbf{failure modification}, which includes the current reward function and test results, is a specific component of the reward refiner. This failure modification will be labeled as a failure exploration that must be avoided. The refinement of the reward function is described by Eq. (\ref{eq:reward_refiner}).   
\begin{align}
    R_{\textcolor{black}{\mathcal{M}}} = \text{RewardRefiner}(R_{{\textcolor{black}{\mathcal{I}}}-1}, Sugg_{{\textcolor{black}{\mathcal{I}}}-1}, R_{{\textcolor{black}{\mathcal{M}}}-1}, Rslt_{{\textcolor{black}{\mathcal{M}}}-1}, Task, Env, \hat{R}) \label{eq:reward_refiner},
\end{align}
where $R_{\textcolor{black}{\mathcal{M}}}$ is the reward function obtained from the reward refiner, and $\text{RewardRefiner}(\cdot)$ represents the \textcolor{black}{prompted LLM utilized by} the \textit{reward refiner}. The relation between the indices ${\textcolor{black}{\mathcal{M}}}$ and ${\textcolor{black}{\mathcal{I}}}$ is outlined in Algorithm \ref{algorithm_LLM}.

\subsection{Implementation details}

\begin{figure}[h]
\centering
\includegraphics[width=1.00\linewidth]{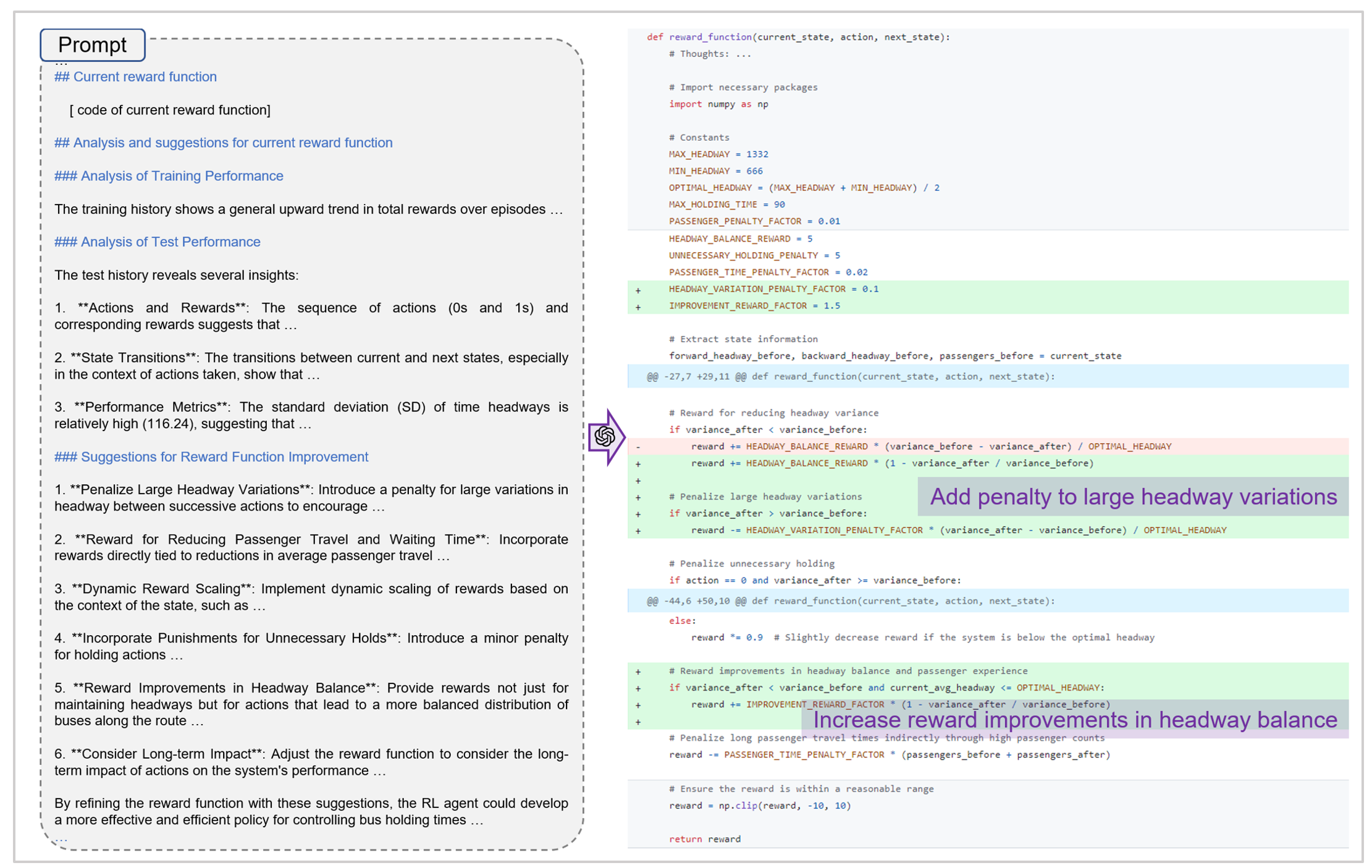}
\caption{\label{fig:reward_modification}Example of reward function modification}
\end{figure}

With the integration of the above-mentioned LLM-based modules, the RL agents are trained at each iteration. When the maximum number of iterations serves as the termination criterion for reward generation, the pseudo-code of the reward generation process is outlined in Algorithm \ref{algorithm_LLM}. To illustrate the improvement of the reward function over iterations, Fig. \ref{fig:reward_modification} presents an example of one round of improving the reward function processed by the reward modifier. In the input prompt of the reward modifier, suggestions for improving the reward function include penalizing large headway variations and improving reward in headway balance. Consequently, the modified reward function incorporates these suggestions by adding two new reward components to the previous reward function. One component penalizes the reward when headway variation increases compared to the previous step. The other component increases the reward when headway variance decreases. Through continuous updates of the reward function, the test performance of the RL agents can be progressively enhanced over iterations.

\begin{algorithm}[h]
\caption{\label{algorithm_LLM}LLM-enhanced RL reward generation}
\begin{algorithmic}
\Require Task ($Task$), Environment ($Env$), Reward Requirements ($\hat{R}$), Refiner's criterion ($p_r[\cdot]$), Maximum iterations ($N$)
\Ensure Final Agent ($A_N$), Final Reward ($R_N$)
\State Initialize agent \(A_{\text{init}}\)
\State Initialize reward function $R_0$ based on Eq. (\ref{eq:reward_initializer})
\State Obtain $Evol_0^{\text{train}}$ and $A_0$ by training the RL agent with $R_0$ based on Eq. (\ref{eq:train_agent})
\State Obtain $Traj_0^{\text{test}}$ and $Rslt_0$ by testing the RL agent with $A_0$ based on Eq. (\ref{eq:test_agent})
\State Obtain $Sugg_0$ with agent performance analyzer based on Eq. (\ref{eq:analyzer})
\For{${\textcolor{black}{\mathcal{I}}} = 1, \dots, N$}
    \State Update $R_{\textcolor{black}{\mathcal{I}}}$ with reward modifier based on Eq. (\ref{eq:reward_modifier})
    \State Update $Evol_{\textcolor{black}{\mathcal{I}}}^{\text{train}}$ and $A_{\textcolor{black}{\mathcal{I}}}$ by training the RL agent with $R_{\textcolor{black}{\mathcal{I}}}$ based on Eq. (\ref{eq:train_agent}) 
    \State Update $Traj_{\textcolor{black}{\mathcal{I}}}^{\text{test}}$ and $Rslt_{\textcolor{black}{\mathcal{I}}}$ by testing the RL agent with $A_{\textcolor{black}{\mathcal{I}}}$ based on Eq. (\ref{eq:test_agent}) 
    \State \({\textcolor{black}{\mathcal{M}}} = {\textcolor{black}{\mathcal{I}}}\)
    \While{\(Rslt_{\textcolor{black}{\mathcal{M}}} > p_r[Rslt_{{\textcolor{black}{\mathcal{I}}}-1}]\)}
        \State ${\textcolor{black}{\mathcal{M}}} \mathrel{+}= 1$
        \State Update $R_{\textcolor{black}{\mathcal{M}}}$ with reward refiner based on Eq. (\ref{eq:reward_refiner})
        \State Update $Evol_{\textcolor{black}{\mathcal{M}}}^{\text{train}}$ and $A_{\textcolor{black}{\mathcal{M}}}$ by training the RL agent with $R_{\textcolor{black}{\mathcal{M}}}$ based on Eq. (\ref{eq:train_agent}) 
        \State Update $Traj_{\textcolor{black}{\mathcal{M}}}^{\text{test}}$ and $Rslt_{\textcolor{black}{\mathcal{M}}}$ by testing the RL agent with $A_{\textcolor{black}{\mathcal{M}}}$ based on Eq. (\ref{eq:test_agent}) 
    \EndWhile
    \State \(R_{\textcolor{black}{\mathcal{I}}} = R_{\textcolor{black}{\mathcal{M}}}\)
    \State \(A_{\textcolor{black}{\mathcal{I}}} = A_{\textcolor{black}{\mathcal{M}}}\)
    \State \(Evol_{\textcolor{black}{\mathcal{I}}}^{\text{train}} = Evol_{\textcolor{black}{\mathcal{M}}}^{\text{train}}\)
    \State \(Traj_{\textcolor{black}{\mathcal{I}}}^{\text{test}} = Traj_{\textcolor{black}{\mathcal{M}}}^{\text{test}}\)
    \State \(Rslt_{\textcolor{black}{\mathcal{I}}} = Rslt_{\textcolor{black}{\mathcal{M}}}\)
    \State Update $Sugg_{\textcolor{black}{\mathcal{I}}}$ with agent performance analyzer based on Eq. (\ref{eq:analyzer})
\EndFor
\end{algorithmic}
\end{algorithm}

\section{Numerical tests}
\label{section:Numerical_tests}

To verify the effectiveness and adaptiveness of the proposed LLM-enhanced RL paradigm, \textcolor{black}{three} different bus holding control scenarios are tested in this section. The first scenario is a synthetic single-line bus system (case study 1), and the second is a real-world two-bus line system with a partially shared corridor (case study 2). \textcolor{black}{The third scenario is a real-world large-scale system consisting of 6 bus lines (case study 3).} In all scenarios, RL agents are considered homogeneous and share a generic reward function. \textcolor{black}{The code and data for this study are available at the~\href{https://github.com/JiajieTSC/LLM-enhanced-RL-for-generic-bus-holding-control}{\textit{link}}.}

\subsection{RL setup and baseline strategies}

\begin{figure}[h]
\centering
\begin{subfigure}{0.33\textwidth}
    \includegraphics[width=\textwidth]{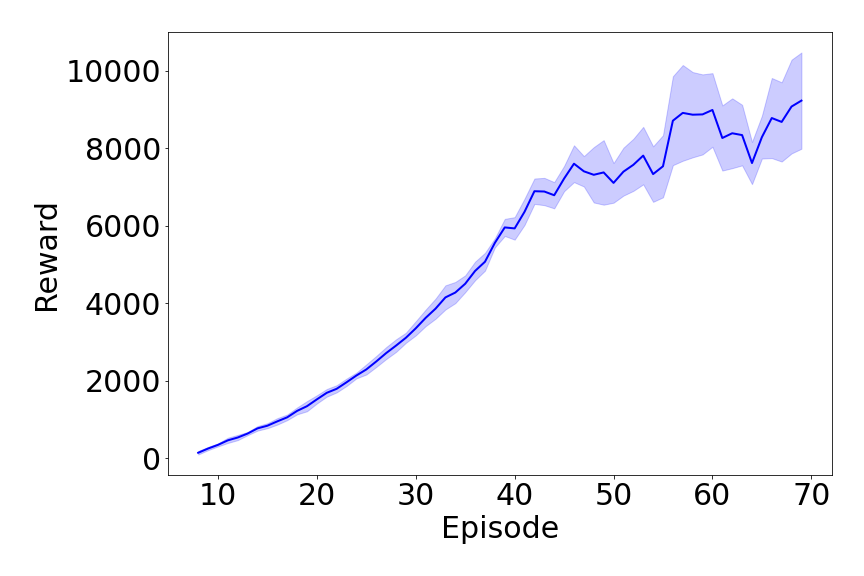}
    \caption{\textcolor{black}{Case 1 (cold start).}}
    \label{fig:reward_curves_case1_cold_start}
\end{subfigure}\hspace{-6mm}
\hfill
\begin{subfigure}{0.33\textwidth}
    \includegraphics[width=\textwidth]{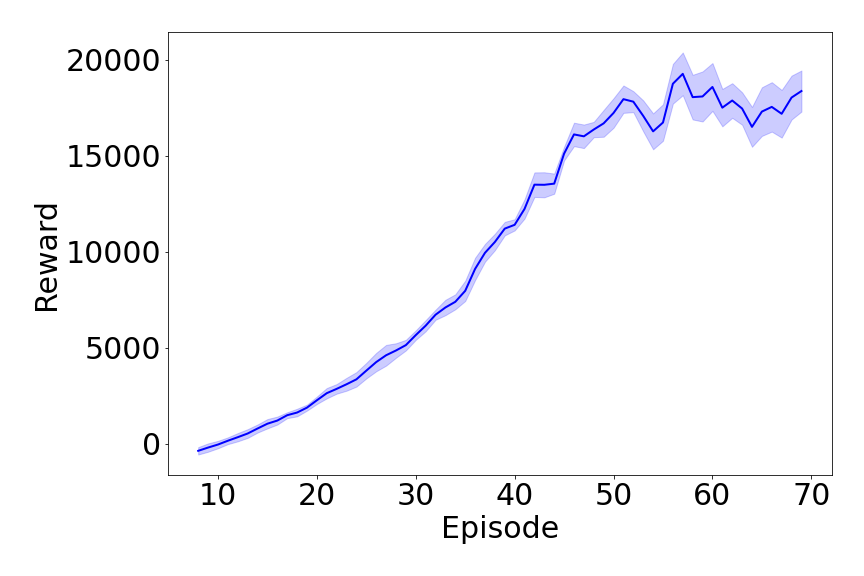}
    \caption{\textcolor{black}{Case 1 (warm start).}}
    \label{fig:reward_curves_case1_warm_start}
\end{subfigure}\hspace{-6mm}
\hfill
\begin{subfigure}{0.33\textwidth}
    \includegraphics[width=\textwidth]{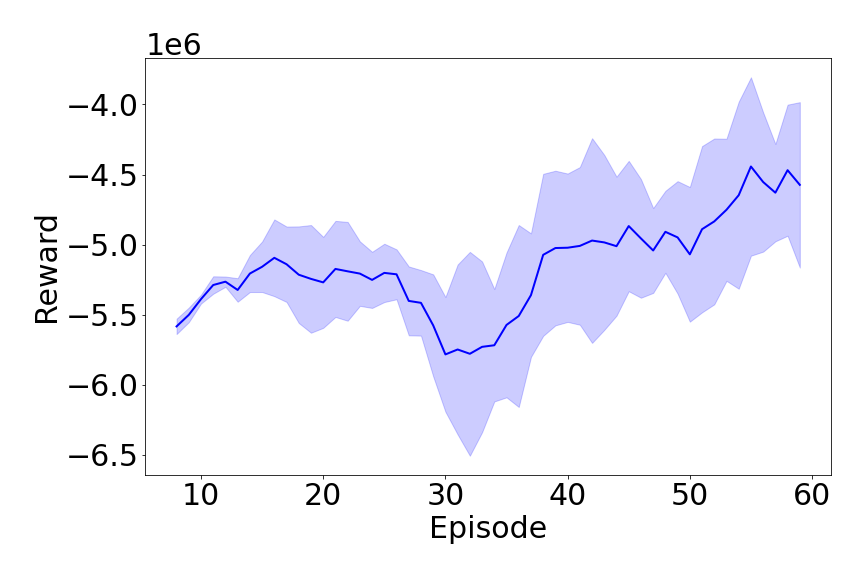}
    \caption{\textcolor{black}{Case 2.}}
    \label{fig:reward_curves_case2}
\end{subfigure}
\caption{\textcolor{black}{Reward curves of trained RL agents.}}
\label{fig:reward_curves}
\end{figure}

The RL-relevant settings are consistent in all case studies. In each iteration, the RL agent is trained for 70 episodes before testing, with each training episode comprising 200 epochs. The random seed for demand generation and environment setup differs in each training and test episode. The simulation runs for 4 hours per training and test episode in case study 1, while in case study 2, it runs for 2 hours per training episode and 4 hours per test episode. \textcolor{black}{The reward curves of the final trained RL agents in case studies 1 and 2 are displayed in Fig. \ref{fig:reward_curves}. The trained RL agent in case study 2 is transferred to case study 3 for tests. The simulation in each test episode of case study 3 runs for 4 hours.} The duration of each action step is 5 seconds. The maximum holding duration is 90 seconds. The structure of the neural network and RL relevant parameters follow the settings in \citet{yu2023partc} due to the similar state and action dimensions of the control task. The neural network is fully connected, with an input layer, four hidden layers of 400 neurons each, and an output layer. The learning rate, $\alpha$, is 0.001. The discounted factor, $\gamma$, for future rewards is 0.95. The exploration rate, $\epsilon$, decreases linearly from 1 to 0.02 for the first 50 episodes and then keeps 0.02 for the remaining episodes.


The baseline bus holding control strategies include vanilla RL-based controllers, the LLM-based controller, \textcolor{black}{physics-based feedback controllers, and optimization-based controllers}. The definitions of these baseline strategies are as follows. 

\begin{easylist}[itemize]
@ \textbf{Vanilla RL-based controllers}: The RL agents are trained with expert-designed reward functions inspired by \citet{wang2020partc} and \citet{yu2023partc}. \textcolor{black}{These existing RL-based methods are included to compare LLM-generated reward functions with manually designed ones.}
@@ RL - local: The reward encourages the variance between forward and backward space headways of the current bus being smaller and penalizes it when the variance becomes greater. In multi-line systems, the reward is calculated for each line and summed up as the final reward. \textcolor{black}{The objective is closely related to the SD of headways and the average waiting time.}
@@ RL - global: The reward aims to minimize the total travel time of all passengers. \textcolor{black}{The objective is directly related to the average travel time.}
@@ RL - local + global: The reward consists of two parts. The first part is the same as the reward design in ``RL - local''. The second part is a penalty for the holding action, as it increases the $TTT$ of in-vehicle passengers. \textcolor{black}{The objective is related to the average travel time and the SD of headways.}
@ \textbf{LLM-based controller:} The LLM is utilized to directly generate holding decisions, including to hold or not to hold (and the holding duration if to hold is chosen), every time a bus dwells at the stop. The input prompt of the LLM-based controller can be found in \ref{appendix:LLM-based_controller}. \textcolor{black}{This method is included for comparison to evaluate the necessity of integrating RL in the proposed paradigm.}
@ \textcolor{black}{\textbf{Physics-Based Feedback Controllers:} These include the model-based feedback controller (denoted as ``model-based controller'') and the space headway-based feedback controller (denoted as ``feedback controller''), both of which represent common and advanced bus holding control methods.}
@@ \textcolor{black}{Model-based controller: This method calculates the holding duration according to the forward and backward bus headways and the number of onboard passengers. The detailed formulation can be found in \citep{laskaris2021partc}.}
@@ Feedback controller: When the backward space headway is larger than the forward space headway, the bus will be held until the backward space headway is no longer larger than the forward headway or the holding time reaches the maximum holding duration. This method has been listed in the benchmarks of \citet{laskaris2021partc}.
@ \textcolor{black}{\textbf{Optimization-based controllers:} These controllers aim to minimize the total waiting time for all passengers. The simulation of bus propagation is incorporated into the optimization framework to accurately estimate passenger travel time and waiting time. Particle Swarm Optimization (PSO) is employed to solve the resulting optimization problems \citep{marini2015PSO}. Robust optimization and stochastic optimization methods are included to determine which approach (RL- or optimization-based methods) is more suitable for real-time control tasks.} 

\textcolor{black}{In these two methods, the control action is optimized every 2,000 seconds. We decompose the optimization process because the entire simulation spans 14,400 seconds, resulting in a large number of decision variables, which makes it challenging to search for a promising solution. Additionally, we set the objective to minimize the total waiting time, rather than the total travel time. Minimizing travel time would encourage the control action to avoid holding altogether to reduce holding delays, which is not suitable for such decomposed settings.}

@@ \textcolor{black}{Robust optimization: Minimize the worst-case total waiting time of all passengers.}
@@ \textcolor{black}{Stochastic optimization: Minimize the expected total waiting time of all passengers.}
\end{easylist}

The evaluation indicators of the test results include the following metrics. These indicators provide a comprehensive assessment of the effectiveness of different bus holding control strategies.

\begin{itemize}
    \item \textbf{Standard deviation (SD) of time headways:} Measures the consistency of bus arrival gaps.
    \item \textbf{Average travel time of passengers:} Indicates the overall efficiency of the bus service in terms of passenger travel cost.
    \item \textbf{Average waiting time of passengers:} Reflects the average time passengers spend at the stops waiting for buses.
    \item \textbf{Average holding time of buses:} Tracks the average duration buses are held at stops.
\end{itemize}

\subsection{Case study 1: single line}

\subsubsection{Scenario description}

\begin{figure}
\centering
\includegraphics[width=0.35\linewidth]{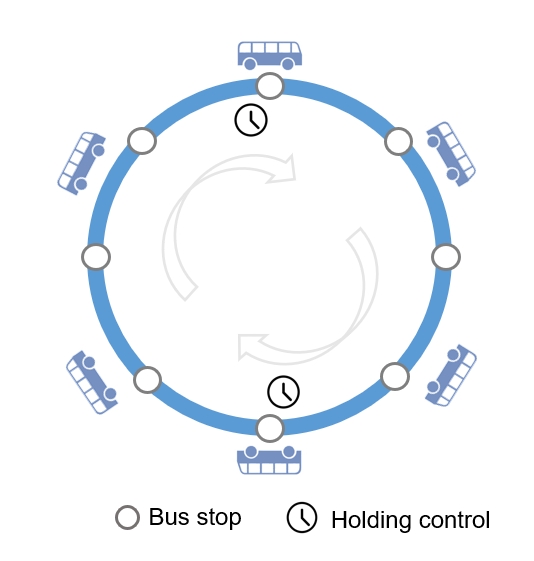}
\caption{\label{fig:busline_case1}Bus lines in case 1}
\end{figure}

In case study 1, the environment is a synthetic single-line bus system referring to \citet{wang2020partc}. 6 buses travel circularly along the bus line. 8 bus stops are distributed evenly along the bus line, as shown in Fig. \ref{fig:busline_case1}. The length of the loop corridor is 10.66 km. The buses travel at a constant speed of 5.55 m/s (20 km/h), with a uniform travel time of 4 minutes between every two adjacent stops. The departure interval between buses is 320 seconds, and overtaking among buses is allowed. Bus dwell time is equal to the boarding time of passengers, with each passenger taking 3 seconds to board. This case study assumes uniform bus speed and travel time, does not account for randomness, and considers only the boarding time for dwell time, excluding the influence of passengers' alighting time and bus capacity. These factors will be addressed in case study 2, which is based on a real-world scenario.


Passengers' arrival follows Poisson processes \citep{berrebi2018partc,wang2020partc}. The average arrival rates of passengers at each bus stop are randomly generated between 40 and 180 pax/h. During the 4-hour simulation, the hourly arrival rate varies as follows: 0.6, 0.8, 1.2, and 0.5 times the average rate, respectively, simulating a morning rush period and incorporating demand variance \citep{luethi2007passenger}.  Each passenger’s alighting stop is randomly chosen from the remaining stops.

\subsubsection{Comparisons with baselines}

The proposed LLM-enhanced RL paradigm is evaluated with both a cold start and a warm start. In the cold start trial, the LLM generates the reward function from scratch via the reward initializer. In the warm start trial, the paradigm starts with an expert-designed reward function, which is the same as the reward in ``RL - local''. The maximum number of iterations is set to 10. The test performance evaluation for the reward refiner's criterion is the Average Travel Time ($ATT$) of all passengers. The criterion for refinement is defined as $p_r[ATT_\text{prev}]=1.10\times ATT_\text{prev}$, meaning the current $ATT$ must not exceed 10\% more than the previous $ATT$ for the modified reward function to be considered acceptable. The LLM used for these experiments (except the sensitivity test on different LLMs specified in subsection \ref{subsection:sensitive_LLMs}) is GPT-4 \citep{openai2023gpt-4}, with a temperature setting of 0.3.  

Table \ref{table:baseline_case1} summarizes the test results of the proposed paradigm and baseline strategies. Each control method was evaluated using 30 different random seeds to mitigate the effects of stochasticity. ``GPT-4 + RL'' and ``GPT-4'' refer to the proposed LLM-enhanced RL paradigm and the LLM-based controller, respectively. The proposed paradigm with a warm start achieves the lowest SD of time headways, average travel time, and average waiting time among all strategies. Specifically, it reduces the average travel time and average waiting time by 3.73\% and 12.29\%, respectively, compared to the best performance of the vanilla RL strategies. In contrast, the LLM-based controller fails to control the buses effectively, resulting in a longer average travel time than the no-holding scenario. Although the average holding time for the LLM-based controller is lower than that for ``GPT-4 + RL'', its significantly higher average waiting time and average travel time indicate that the control actions selected directly by the LLM are unreasonable. ``RL - global'' despite having the same training settings as other RL methods, shows poor exploration performance and test results. This highlights the inefficiency of using sparse rewards in tasks with dense actions. \textcolor{black}{``RL - local + global'' underperforms ``RL - local'' because the holding delay penalty in RL-local+global may cause agents to avoid holding in too many steps to minimize holding delay, leading to a local optimum that is not an effective overall strategy.} Feedback control also yields promising results, closely approaching the performance of the proposed approach. In this simplified synthetic single-line bus system, it is easy to get a satisfactory performance with feedback control despite it being a single-objective method focusing on balancing space headways. \textcolor{black}{Model-based control outperforms RL-based controllers in terms of average travel time but underperforms compared to feedback control. This is because model-based control relies on a predefined model of bus propagation in its formulations, which may differ from real-time simulations due to the stochastic nature of passenger arrivals. In contrast, feedback control determines the holding duration solely based on real-time observations, allowing it to better adapt to the dynamic conditions of the simulation.}

Robust optimization and stochastic optimization underperform the LLM-enhanced RL, RL-local, and RL-local+global in this test. The reasons are summarized as follows:
\textcolor{black}{First, this test is based on a simulator that cannot be adequately modeled using only deterministic bus propagation models, as passenger arrivals are random. Both robust and stochastic optimization models are static in nature. While they can capture the distribution of random variables (e.g., passenger arrival rates, travel times between bus stops), they are not equipped to respond to real-time simulations. In contrast, all RL-based controllers, including the proposed paradigm, interact with the environment throughout the entire test. These controllers can dynamically adapt to the environment in an online manner, optimizing real-time performance at each action step.
Second, robust optimization focuses on improving the worst-case performance, which can lead to conservative actions in real-time control. This is because there may be a significant discrepancy between the actual conditions and the worst-case scenario in the environment. Stochastic optimization aims to optimize expected performance. However, in real-time tasks, the actual dynamic environment may deviate from the expected conditions, and stochastic optimization cannot effectively respond to such discrepancies due to its offline nature.
Third, exact algorithms are not effective in this context since the evaluation in objective functions is derived from simulation, not detailed formulations. PSO is applied here to solve the optimization problem. However, PSO cannot guarantee a solution close to the global optimum, especially given the complexity of the problem and the high dimensionality of the variables. As a result, PSO may converge to a local optimum rather than the global one.
Fourth, RL-based methods aim to maximize the cumulative reward over future steps rather than focusing solely on the immediate reward at the current step. This capability allows RL-based methods to estimate the agent's future performance, which is particularly beneficial in dynamic and fluctuating real-time environments. This forward-looking and online approach is a significant advantage of RL-based methods compared to offline optimization methods.
These factors contribute to the suboptimal performance of the optimization-based controllers in this study.}

\textcolor{black}{Concerning the limitations of LLMs (e.g., high cost and delayed response), we utilize LLMs in an offline setting to generate the reward function for the RL agent. Since the reward function does not require updates during the testing process, the trained RL agent does not need to interact with the LLMs in real-time during implements. Therefore, the delayed response of LLMs is not a significant issue in this setup. Furthermore, during the reward refinement process, the LLMs only needs to output the responses when the training and testing processes of the RL agent have been completed. Compared to directly using LLMs as controllers, the proposed paradigm reduces the need for frequent outputs from the LLMs, which effectively reduces the costs.
}

\begin{table}[h]
\centering
\caption{\label{table:baseline_case1}Evaluations of different control methods in Case 1}
\begin{tblr}{
  width = \linewidth,
  colspec = {Q[3.5]Q[2]Q[2]Q[2]Q[2]},
  cells = {c},
  hline{1-2,13} = {-}{},
}
\textbf{Control method} & \textbf{SD of time headways} & \textbf{Avg. travel time} & \textbf{Avg. waiting time} & \textbf{Avg. holding time}\\
GPT-4 + RL (cold start) & 88.14$\pm$21.88 & 1373.51$\pm$53.34 & 228.94$\pm$11.08 & 15.40$\pm$5.39\\
GPT-4 + RL (warm start) & \textbf{60.86}$\pm$8.78 & \textbf{1343.29}$\pm$31.29 & \textbf{220.73}$\pm$6.23 & 12.05$\pm$2.30\\
GPT-4 & 563.24$\pm$60.07 & 1899.6$\pm$100.71 & 636.90$\pm$87.74 & 13.85$\pm$1.30\\
Feedback control & 69.92$\pm$12.06 & 1349.12$\pm$33.37 & 223.09$\pm$7.20 & 12.31$\pm$2.25\\
Model-based control & 103.34$\pm$94.64 & 1376.79$\pm$112.64 & 242.71$\pm$82.89 & 12.34$\pm$2.91\\
RL - local & 150.29$\pm$67.71 & 1395.31$\pm$71.19 & 251.65$\pm$31.73 & 14.02$\pm$5.09\\
RL - global & 696.25$\pm$53.85 & 2352.71$\pm$96.88 & 803.71$\pm$87.16 & 89.67$\pm$0.25\\
RL - local + global & 303.78$\pm$219.55 & 1562.1$\pm$238.31 & 400.10$\pm$187.22 & 8.48$\pm$2.62\\
Robust optimization & 412.56$\pm$56.99 & 1758.58$\pm$63.93 & 440.18$\pm$62.86 & 36.91$\pm$4.41\\
Stochastic optimization & 465.34$\pm$62.58 & 1804.58$\pm$90.75 &  502.9$\pm$75.36 &  37.33$\pm$4.49\\
No holding & 411.15$\pm$61.24 & 1606.87$\pm$83.86 & 441.08$\pm$67.49 & 0$\pm$0
\end{tblr}
\end{table}

\begin{figure}[h]
\centering
\begin{subfigure}{0.52\textwidth}
    \includegraphics[width=\textwidth]{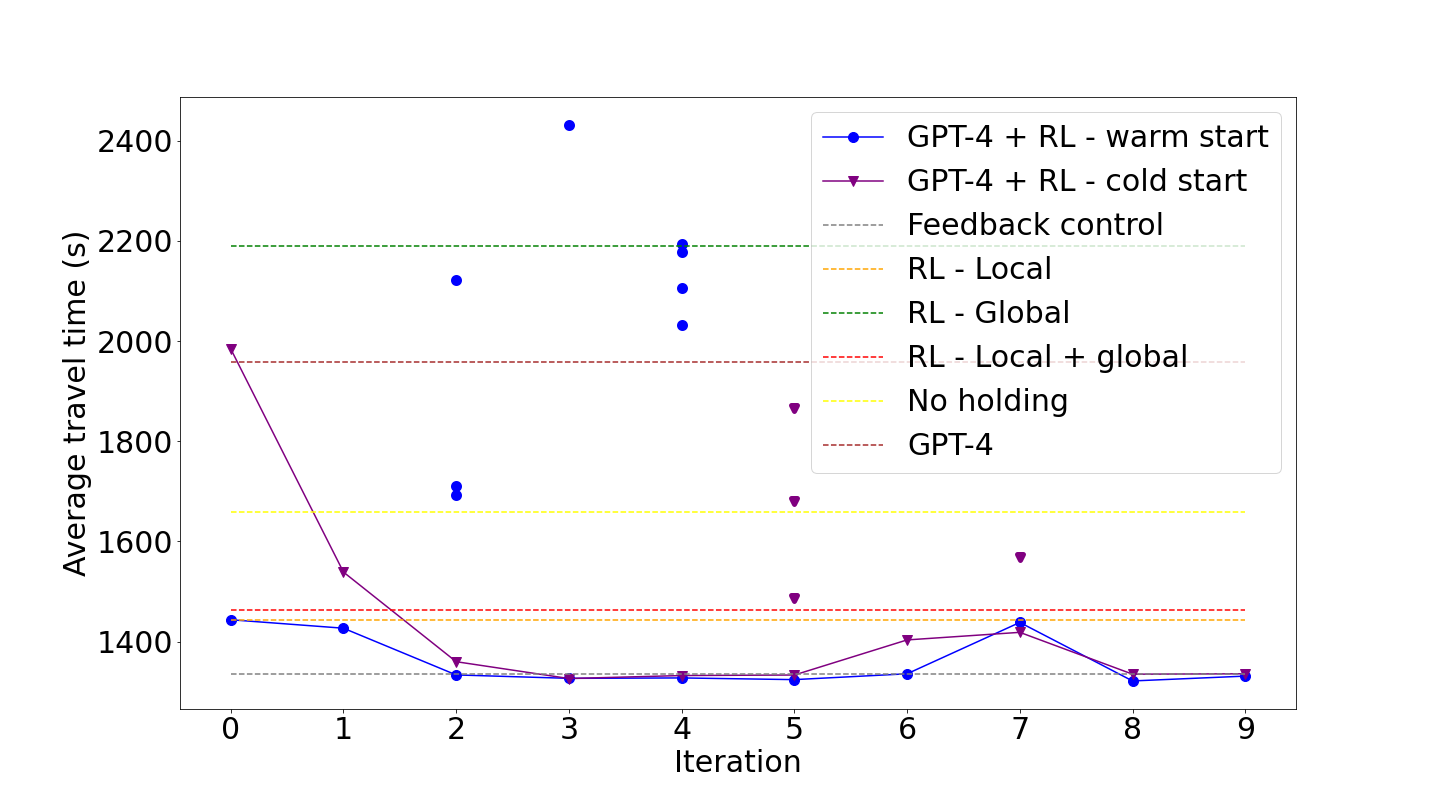}
    \caption{Average travel time.}
    \label{fig:case1_baseline_avg_travel_time}
\end{subfigure}\hspace{-16mm}
\hfill
\begin{subfigure}{0.52\textwidth}
    \includegraphics[width=\textwidth]{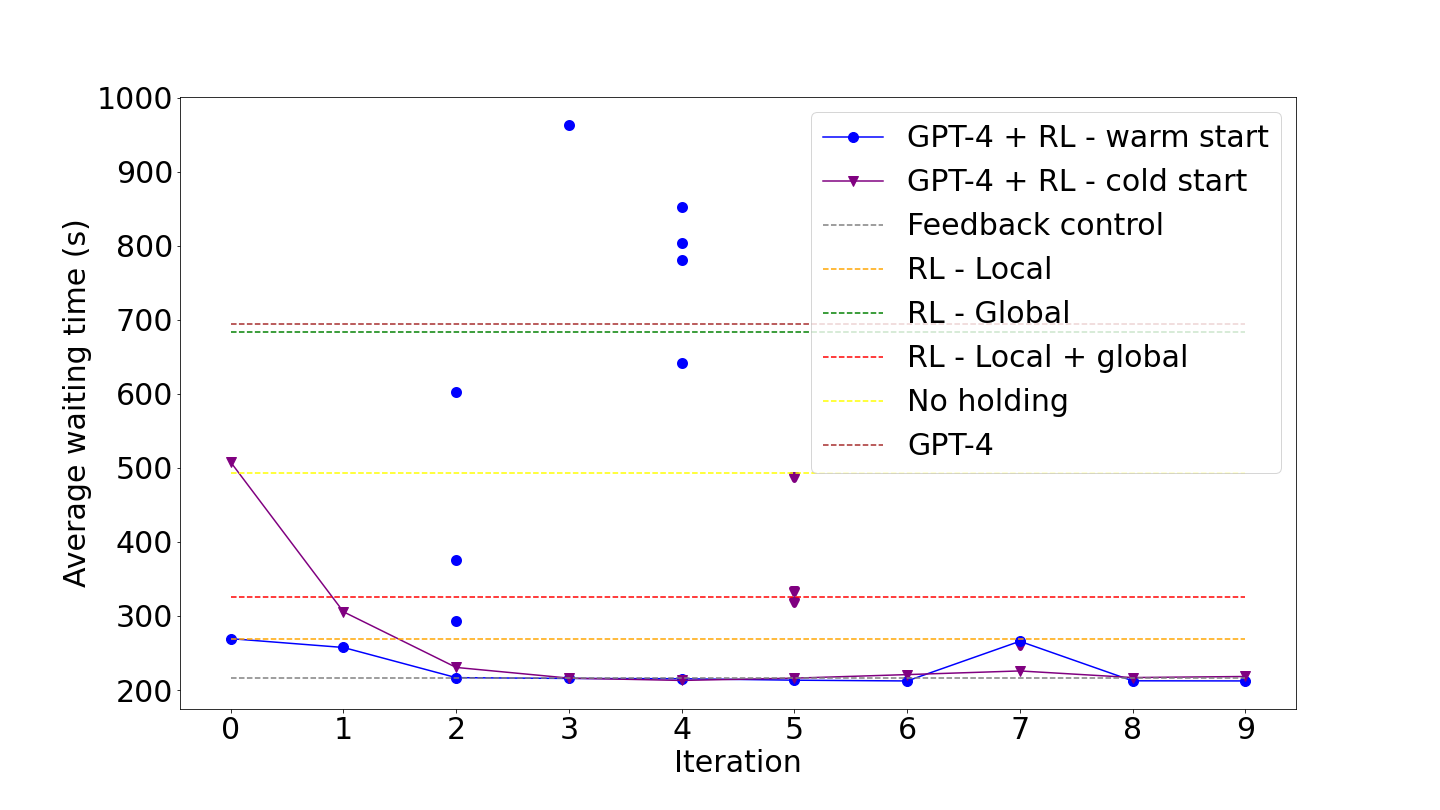}
    \caption{Average waiting time.}
    \label{fig:case1_baseline_avg_waiting_time}
\end{subfigure}
\hfill
\begin{subfigure}{0.52\textwidth}
    \includegraphics[width=\textwidth]{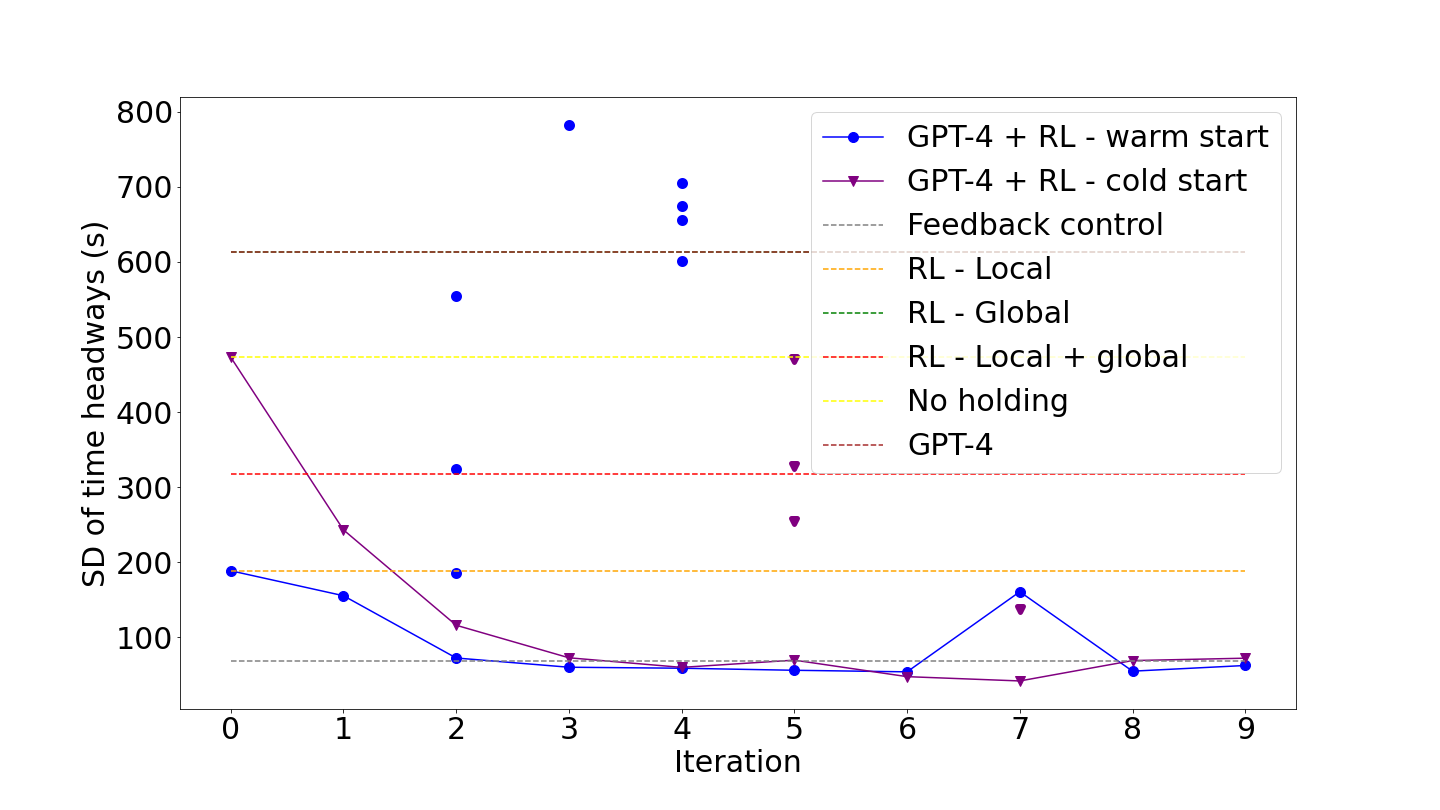}
    \caption{SD of time headways.}
    \label{fig:case1_baseline_SD_time_headways}
\end{subfigure}\hspace{-16mm}
\hfill
\begin{subfigure}{0.52\textwidth}
    \includegraphics[width=\textwidth]{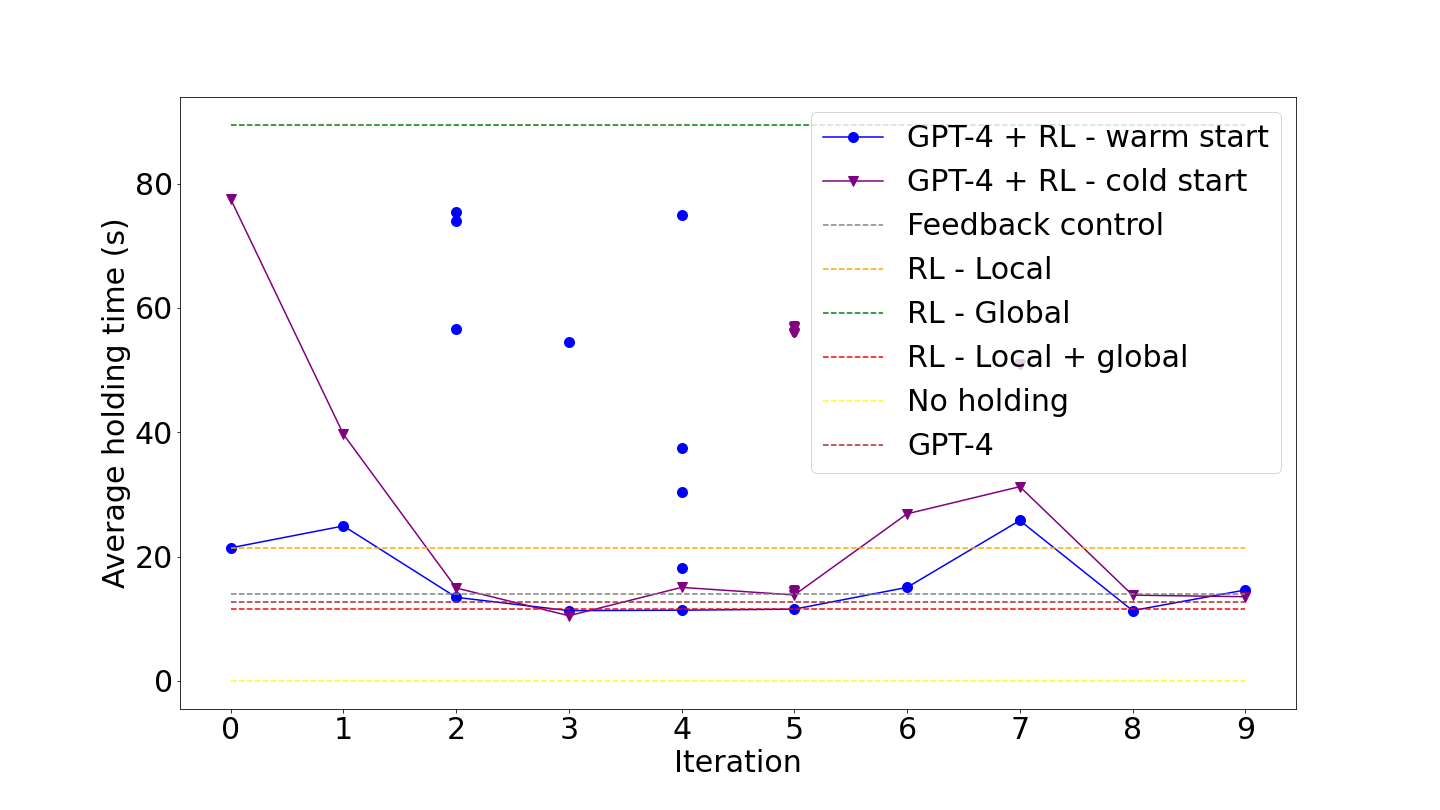}
    \caption{Average holding time.}
    \label{fig:case1_baseline_avg_holding_time}
\end{subfigure}
\caption{Evaluation comparisons with baselines.}
\label{fig:case1_baselines}
\end{figure}

Fig. \ref{fig:case1_baselines} illustrates the evolution of test performance for the proposed LLM-enhanced RL paradigm across iterations. \textcolor{black}{To simplify the presentation, all subsequent figures and additional analyses are based on the test results from the simulation using a random seed of 70.} The figure shows how the evaluation metrics change as iterations progress. The points outside the curves represent the test results from RL agents trained with reward functions that are eliminated by the reward refiner due to not meeting the performance criteria. The evaluations of average travel time, average waiting time, and SD of time headways demonstrate improvement and eventual convergence over iterations in both the cold and warm starts of the proposed paradigm, indicating that the iterative reward refinement process effectively enhances the control strategy.

\begin{figure}[h]
\centering
\begin{subfigure}{0.51\textwidth}
    \includegraphics[width=\textwidth]{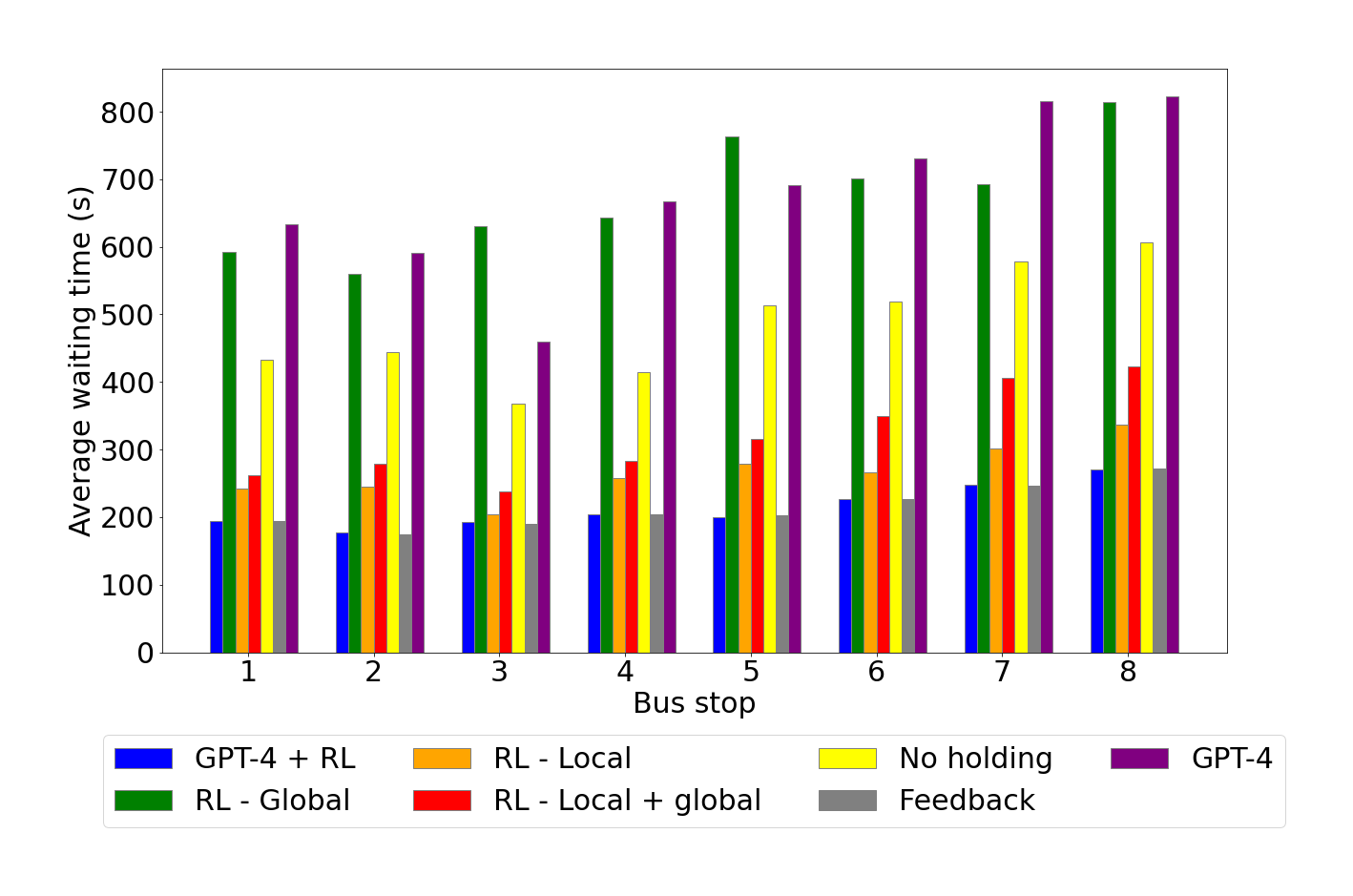}
    \caption{Average waiting time.}
    \label{fig:case1_distributed_waiting_time}
\end{subfigure}\hspace{-6mm}
\hfill
\begin{subfigure}{0.51\textwidth}
    \includegraphics[width=\textwidth]{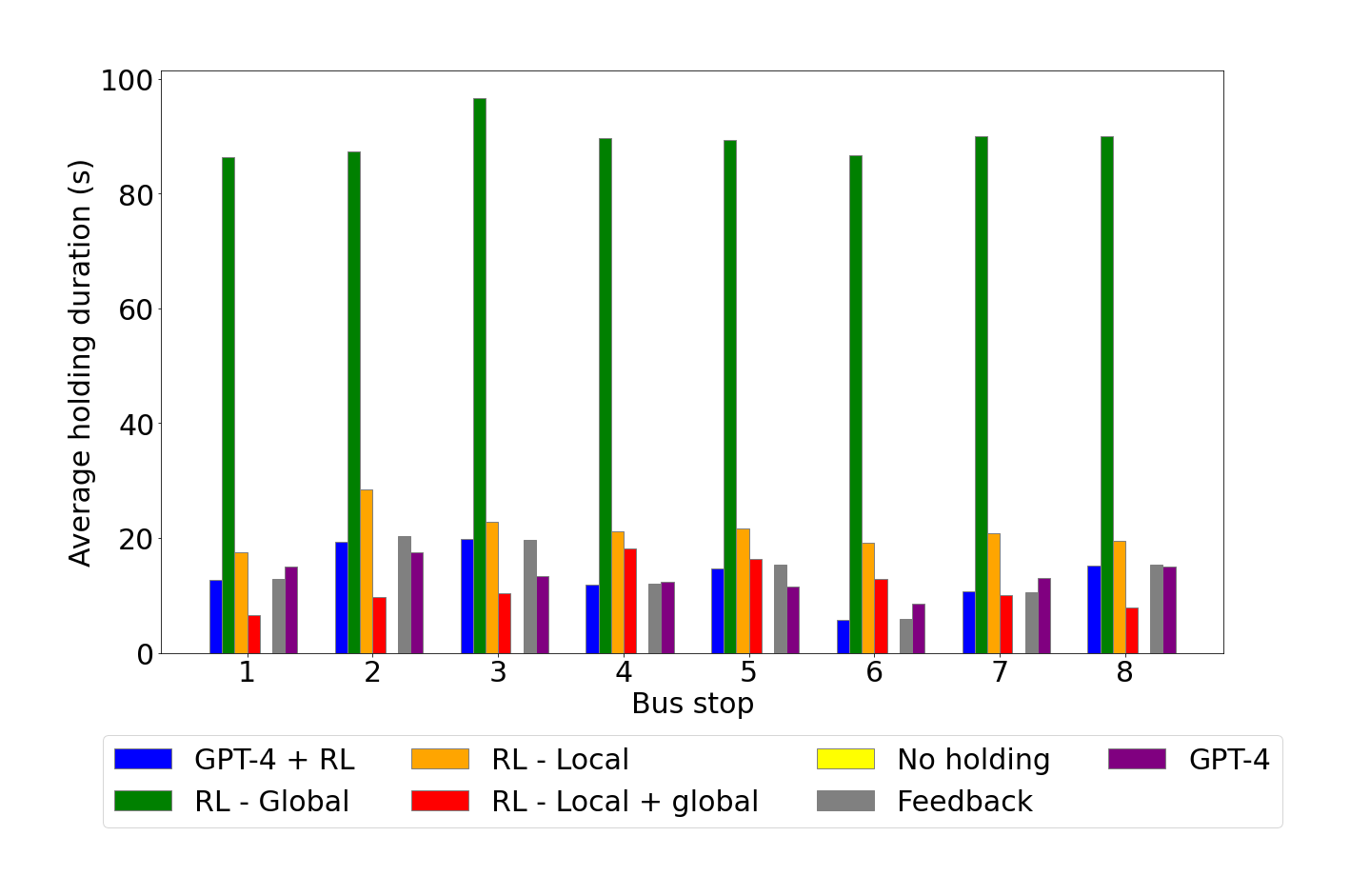}
    \caption{Average holding time.}
    \label{fig:case1_distributed_holding_time}
\end{subfigure}
\hfill
\begin{subfigure}{0.70\textwidth}
    \includegraphics[width=\textwidth]{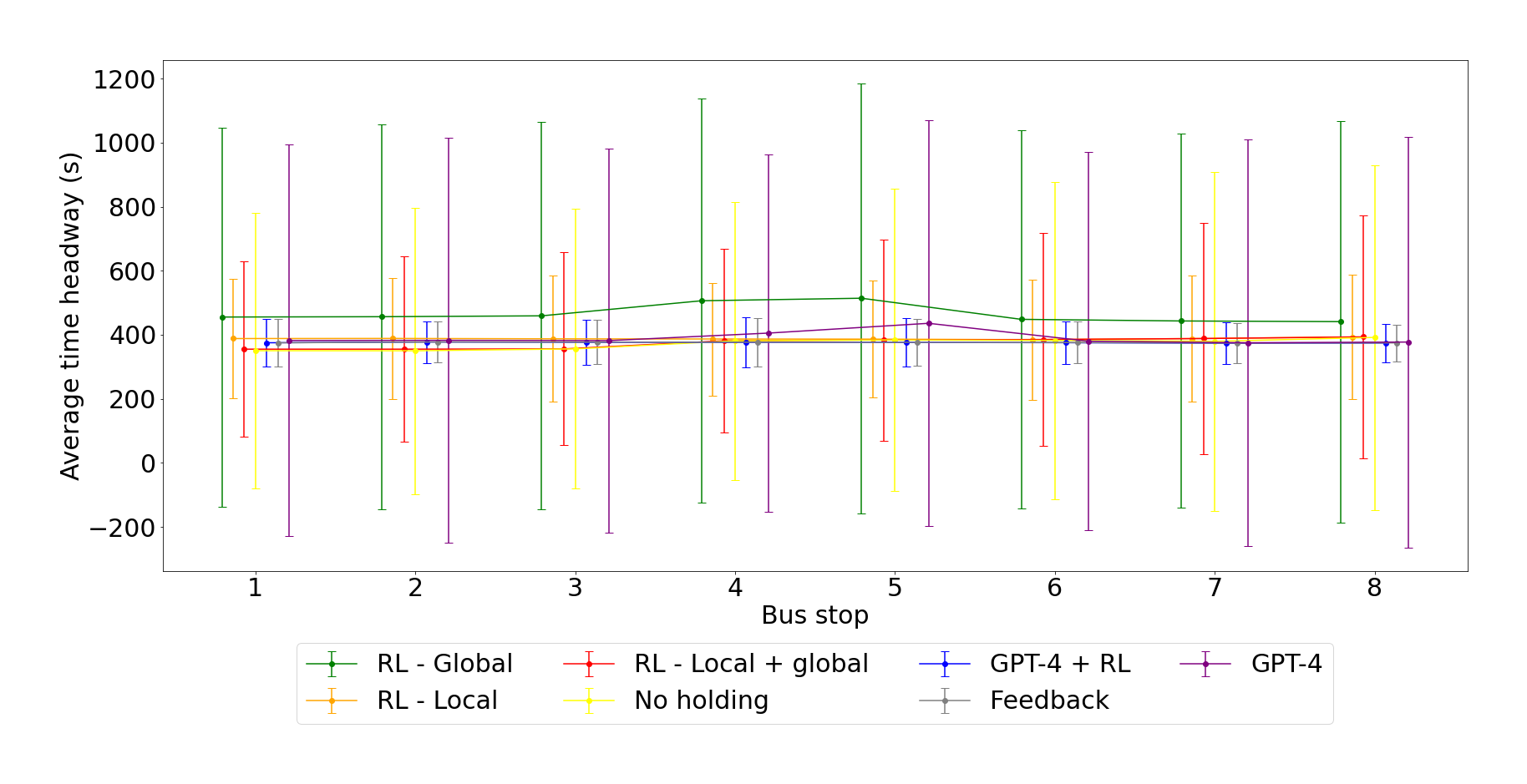}
    \caption{Average time headway.}
    \label{fig:case1_distrubuted_time_headways}
\end{subfigure}
\caption{Evaluation comparisons with baselines along each bus stop.}
\label{fig:case1_distributed_evaluations}
\end{figure}

The performance of each control strategy distributed across different bus stops is depicted in Fig. \ref{fig:case1_distributed_evaluations}. In Fig. \ref{fig:case1_distrubuted_time_headways}, each point represents the average time headway at a bus stop, with error bars indicating the SD of time headways. The distributed performance of each control strategy shows the same result as the overall performance. Feedback control and the proposed LLM-enhanced RL paradigm exhibit similar performances in terms of average waiting time and SD of time headways, outperforming the other strategies at each bus stop.

\begin{figure}[h]
\centering
\begin{subfigure}{0.52\textwidth}
    \includegraphics[width=\textwidth]{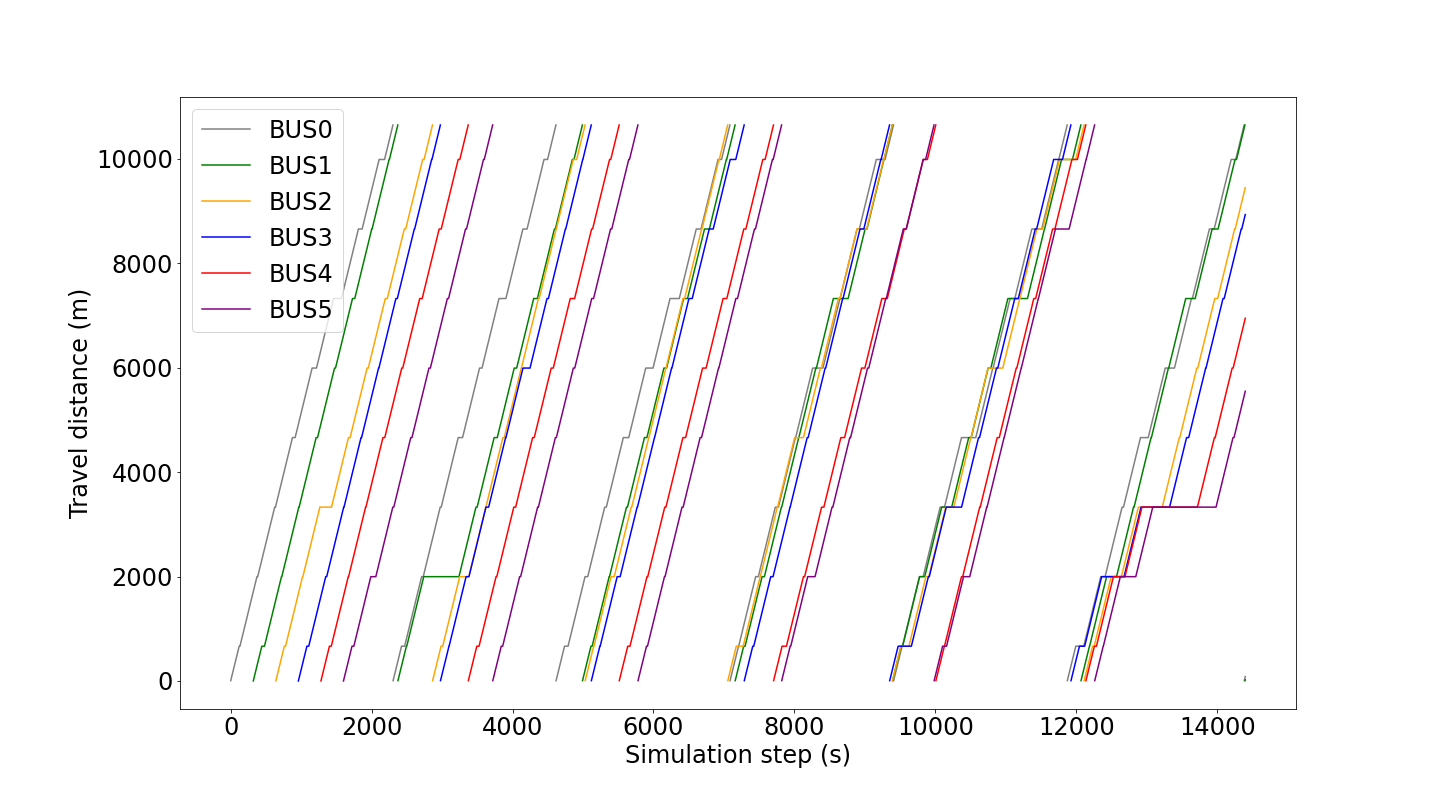}
    \caption{No holding.}
    \label{fig:case1_bus_trajectory_noholding}
\end{subfigure}\hspace{-16mm}
\hfill
\begin{subfigure}{0.52\textwidth}
    \includegraphics[width=\textwidth]{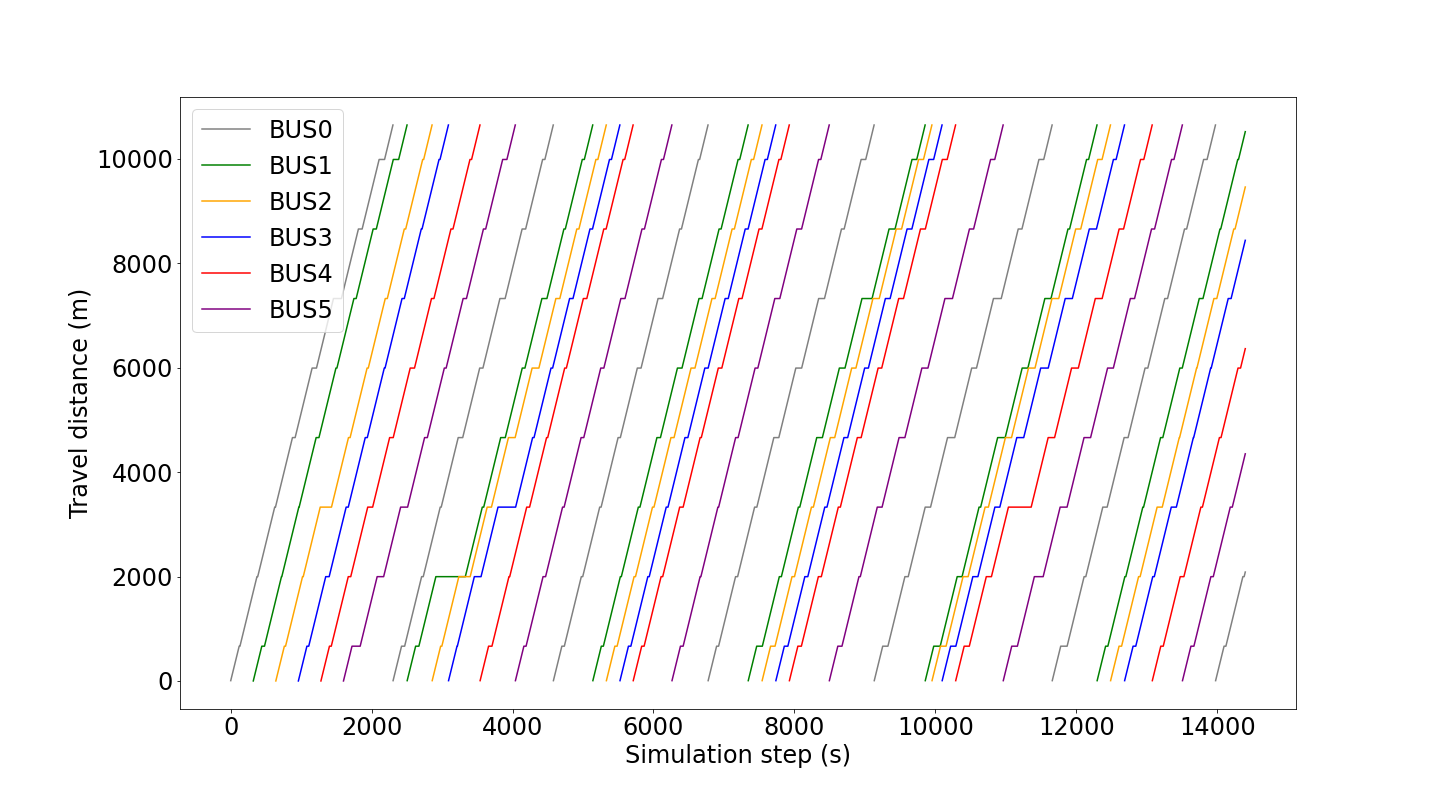}
    \caption{RL - local.}
    \label{fig:case1_bus_trajectory_RL_local}
\end{subfigure}
\hfill
\begin{subfigure}{0.52\textwidth}
    \includegraphics[width=\textwidth]{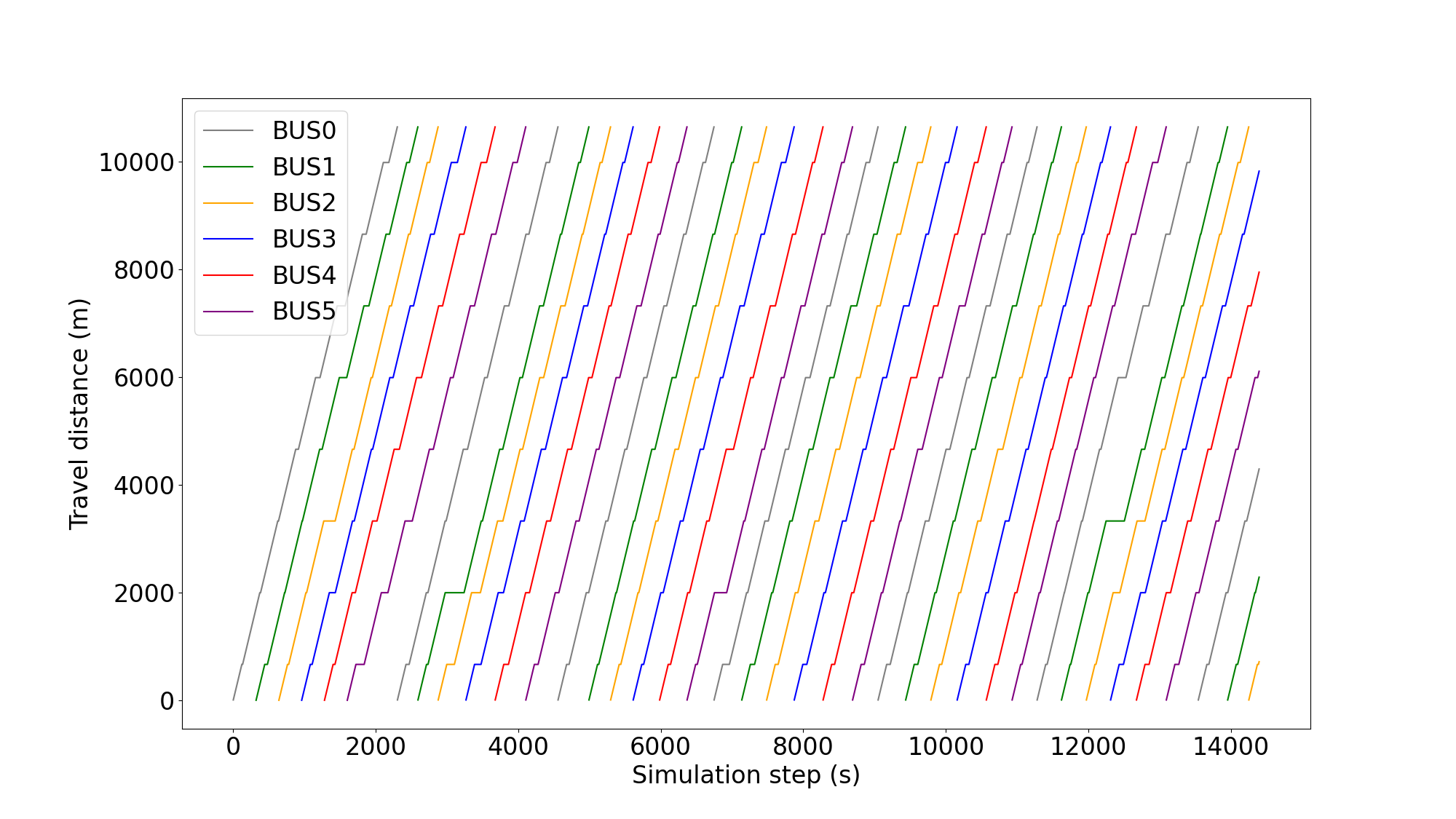}
    \caption{GPT4 + RL.}
    \label{fig:case1_bus_trajectory_gpt4_RL}
\end{subfigure}\hspace{-16mm}
\hfill
\begin{subfigure}{0.52\textwidth}
    \includegraphics[width=\textwidth]{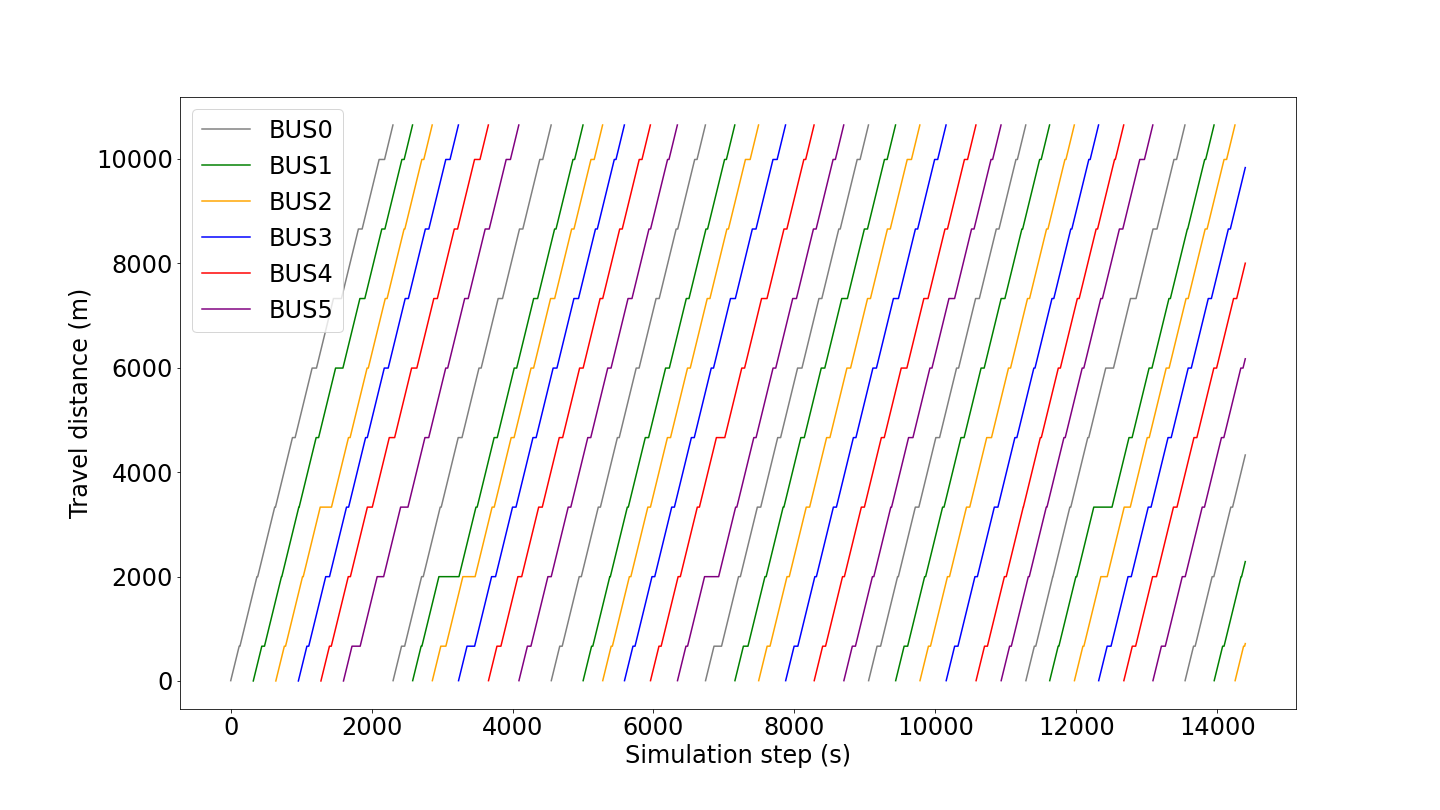}
    \caption{Feedback control.}
    \label{fig:case1_bus_trajectory_feedback_control}
\end{subfigure}
\caption{Bus trajectories in case study 1}
\label{fig:case1_bus_trajectories}
\end{figure}

Bus trajectories are displayed in Fig. \ref{fig:case1_bus_trajectories}, where two lines getting close or crossing each other indicate bus bunching. Severe bus bunchings are observed in the no-holding scenario. ``RL -local'' effectively mitigates bus bunching and reduces the SD of time headways compared to the no-holding scenario, but still presents some imbalance in headway distribution. Specifically, the LLM-enhanced RL reduces the SD of time headways by 59.50\% compared to ``RL - local'', while feedback control achieves a 53.48\% reduction.

\subsubsection{Sensitiveness on LLMs}
\label{subsection:sensitive_LLMs}

\begin{table}[h]
\centering
\caption{\label{table:various_llms}Evaluations of different LLMs}
\begin{tblr}{
  width = \linewidth,
  colspec = {Q[1.5]Q[1.2]Q[1.1]Q[1.25]Q[1.3]Q[1]},
  cells = {c},
  hline{1-2,7} = {-}{},
}
\textbf{LLM} & \textbf{ SD of time headways } & \textbf{ Avg. travel time } & \textbf{ Avg. waiting time } & \textbf{ Avg. holding time } & \textbf{ Num of errors }\\
GPT-4 & 72.38 & \textbf{1335.91} & \textbf{218.27} & 13.55 & \textbf{0/14}\\
Claude-Opus & 187.96 & 1376.18 & 256.24 & 15.04 & 7/32\\
Gemini-1.0-pro & 473.30 & 1658.91 & 492.72 & 0.00 & 4/22\\
GPT-3.5 & 99.56 & 1358.53 & 227.28 & 14.80 & 0/16\\
GPT-4o & \textbf{48.55} & 1372.83 & 221.96 & 27.68 & 0/38
\end{tblr}
\end{table}

We evaluate the performance of various LLMs in the proposed paradigm. Table \ref{table:various_llms} presents the test performances of the proposed paradigm empowered by GPT-4, Claude-Opus \citep{Cluade}, Gemini-1.0-pro \citep{team2023gemini}, GPT-3.5, and GPT-4o \citep{openai2023gpt-4}. Among these, GPT-4o achieves the minimum SD of time headways, while GPT-4 achieves the lowest average travel time and average waiting time. Furthermore, GPT-4 requires the fewest times of reward generations (including the output from both the reward modifier and refiner) to complete the required iterations, indicating more efficient reward exploration. The ``number of errors'' represents the occurrences of syntax errors in the total number of LLM-generated reward functions. Overall, the GPT family outperforms Claude-Opus and Gemini-1.0-pro in terms of control performance, reward exploration efficiency, and coding accuracy with this bus holding control scenario.

\begin{figure}[h]
\centering
\begin{subfigure}{0.52\textwidth}
    \includegraphics[width=\textwidth]{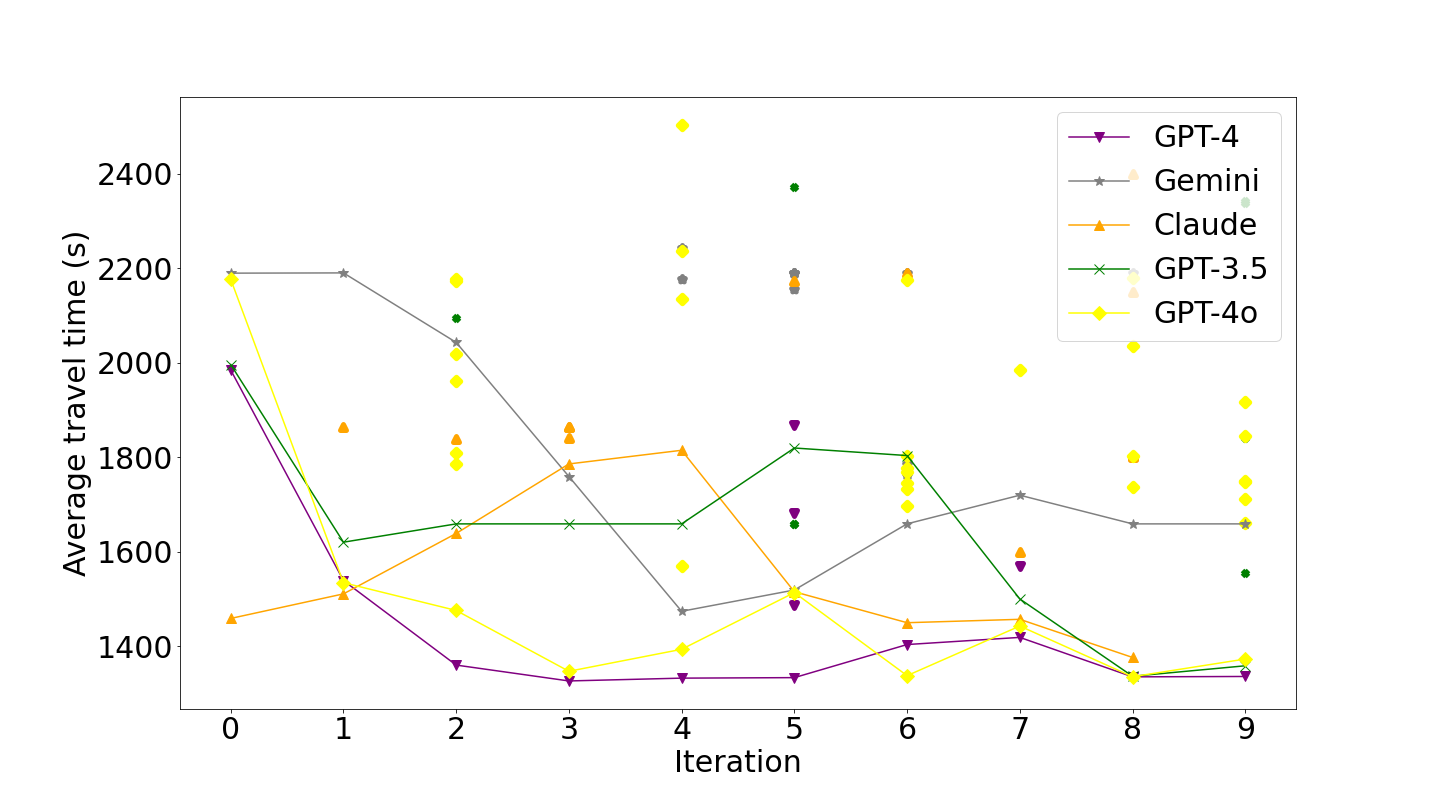}
    \caption{Average travel time.}
    \label{fig:case1_llms_avg_travel_time}
\end{subfigure}\hspace{-16mm}
\hfill
\begin{subfigure}{0.52\textwidth}
    \includegraphics[width=\textwidth]{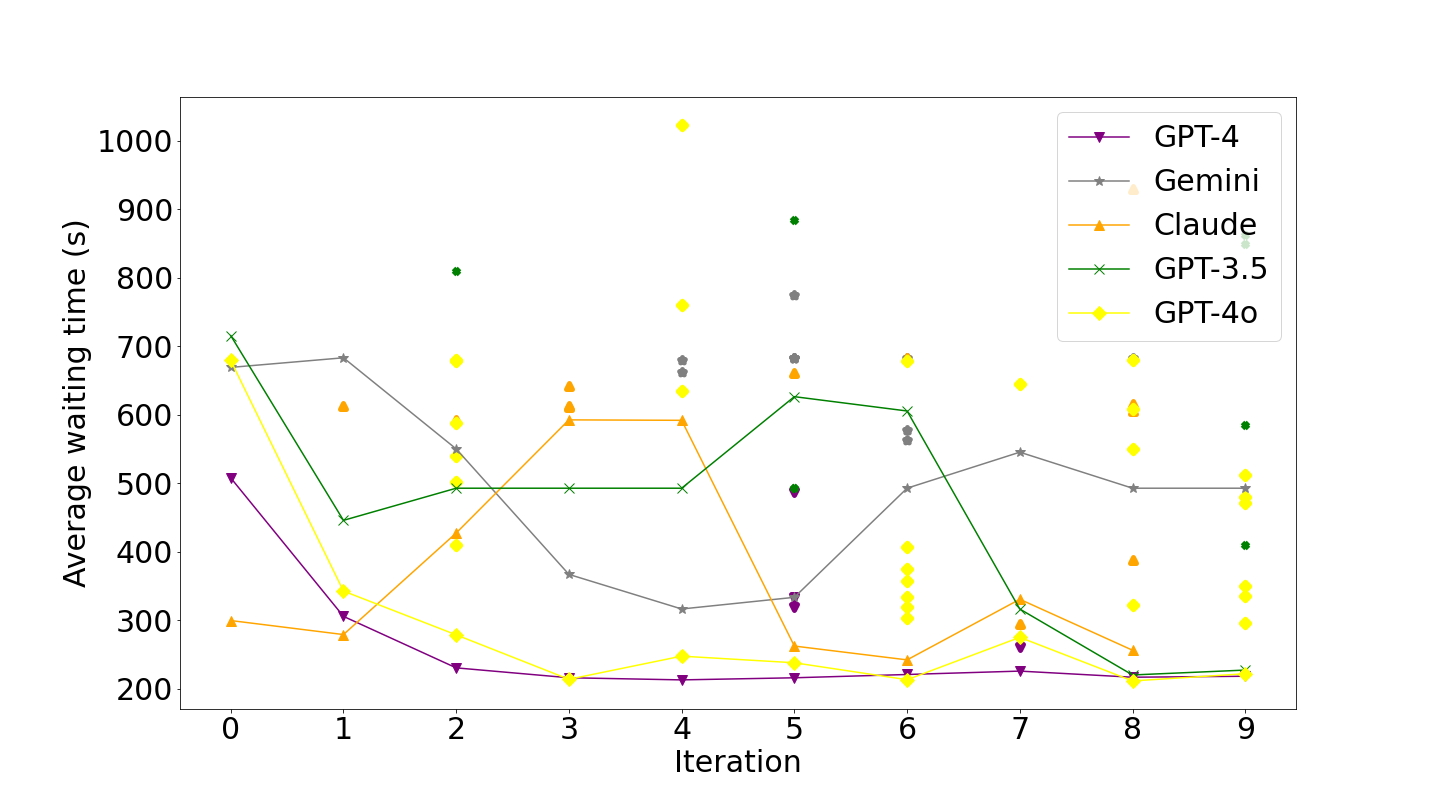}
    \caption{Average waiting time.}
    \label{fig:case1_llms_avg_waiting_time}
\end{subfigure}
\hfill
\begin{subfigure}{0.52\textwidth}
    \includegraphics[width=\textwidth]{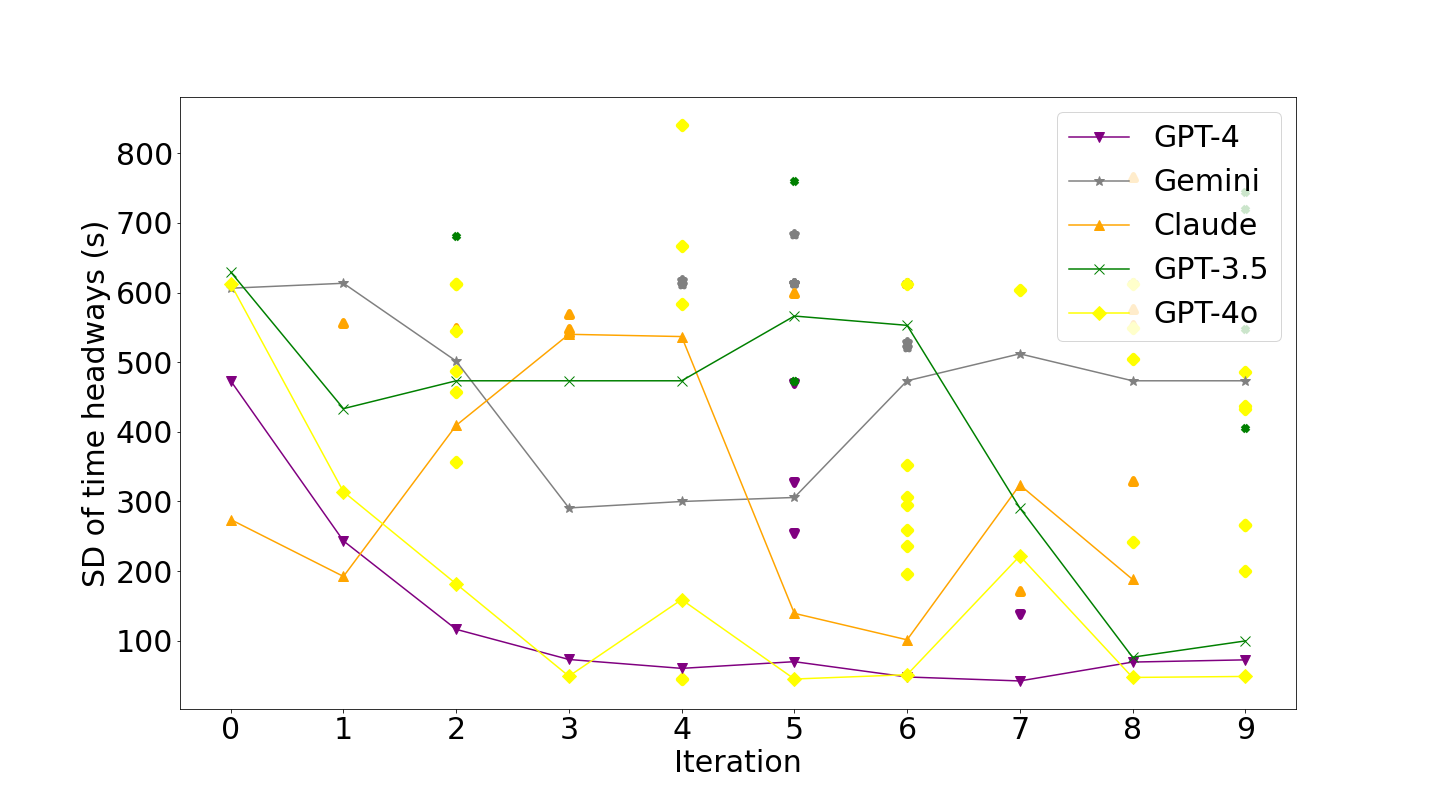}
    \caption{SD of time headways.}
    \label{fig:case1_llms_SD_time_headways}
\end{subfigure}\hspace{-16mm}
\hfill
\begin{subfigure}{0.52\textwidth}
    \includegraphics[width=\textwidth]{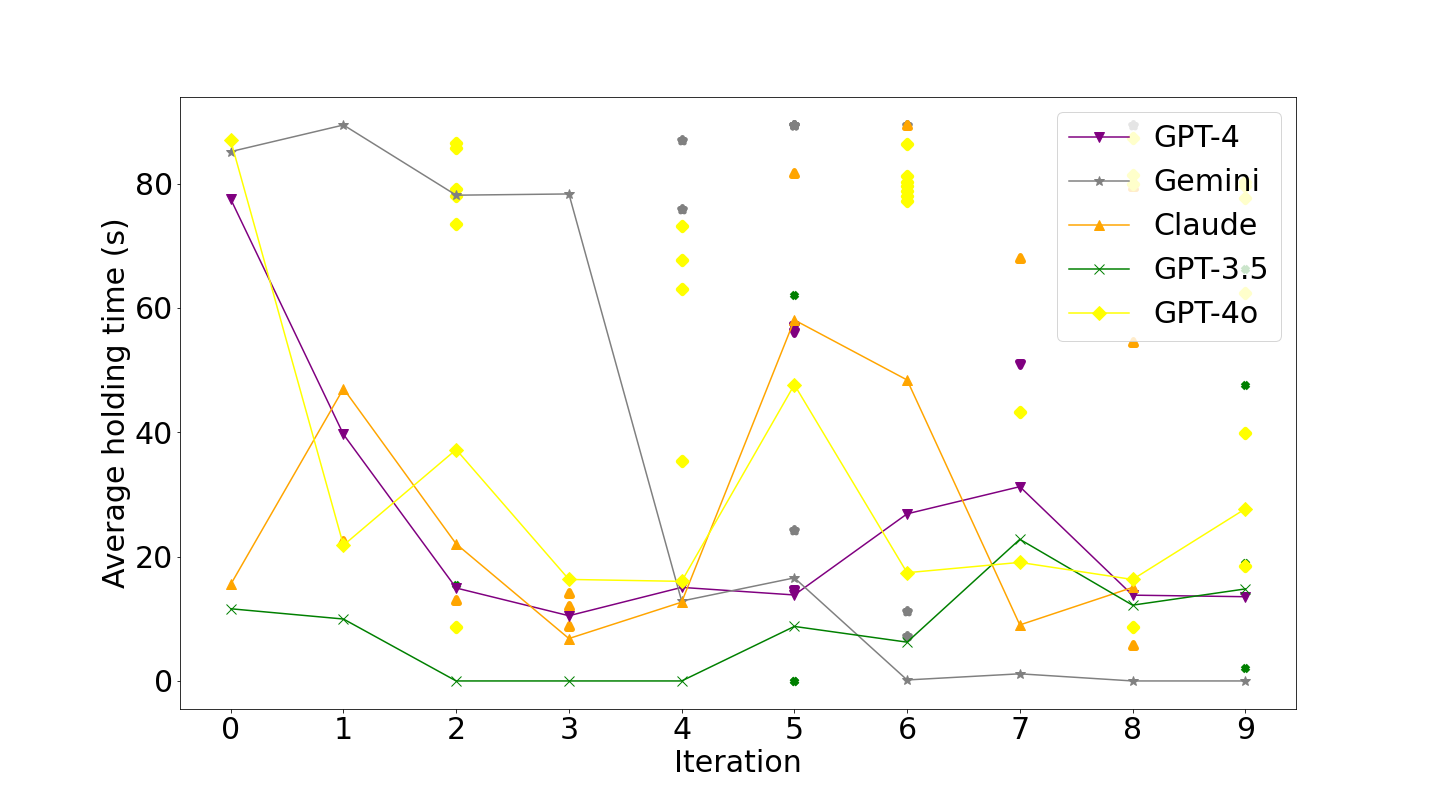}
    \caption{Average holding time.}
    \label{fig:case1_llms_avg_holding_time}
\end{subfigure}
\caption{Evaluation comparisons with other LLMs.}
\label{fig:case1_llms}
\end{figure}

Fig. \ref{fig:case1_llms} presents the evaluation trends of the proposed approach empowered by different LLMs over iterations. Claude-Opus does not achieve performance that meets the reward refiner's criteria within a reasonable number of explorations in the tenth iteration, resulting in only 9 iterations being shown for Claude-Opus in the figure. As depicted in Figs. \ref{fig:case1_llms_avg_travel_time}, \ref{fig:case1_llms_avg_waiting_time}, and \ref{fig:case1_llms_SD_time_headways}, GPT-4 and GPT-4o demonstrate more efficient improvements at the beginning of the iterations and exhibit more stable performance in subsequent iterations compared to the other LLMs.

\subsubsection{Comparisons with existing LLM-enhanced RL methods}

Three alternative LLM-enhanced RL methods: ``with filter'' \citep{yu2023preprint}, ``with tips'', and ``with critic'' \citep{li2024autoMC} are evaluated to compare with the proposed method. All three approaches utilize LLMs to generate reward functions for RL agents but employ different strategies instead of the \textit{reward refiner} module.

LLM-enhanced RL with filter involves the LLM generating multiple reward functions and selects the best-performing one for the RL agent \citep{yu2023preprint}. LLM-enhanced RL with tips incorporates insights from previously ineffective reward functions into the prompt (as tips) for the \textit{reward modifier} module, avoiding similar failures in future reward generation. In LLM-enhanced RL with critic, a \textit{reward critic} module evaluates and adjusts the reward function generated by the \textit{reward modifier} module at each iteration \citep{li2024autoMC}.

\begin{figure}[h]
\centering
\begin{subfigure}{0.52\textwidth}
    \includegraphics[width=\textwidth]{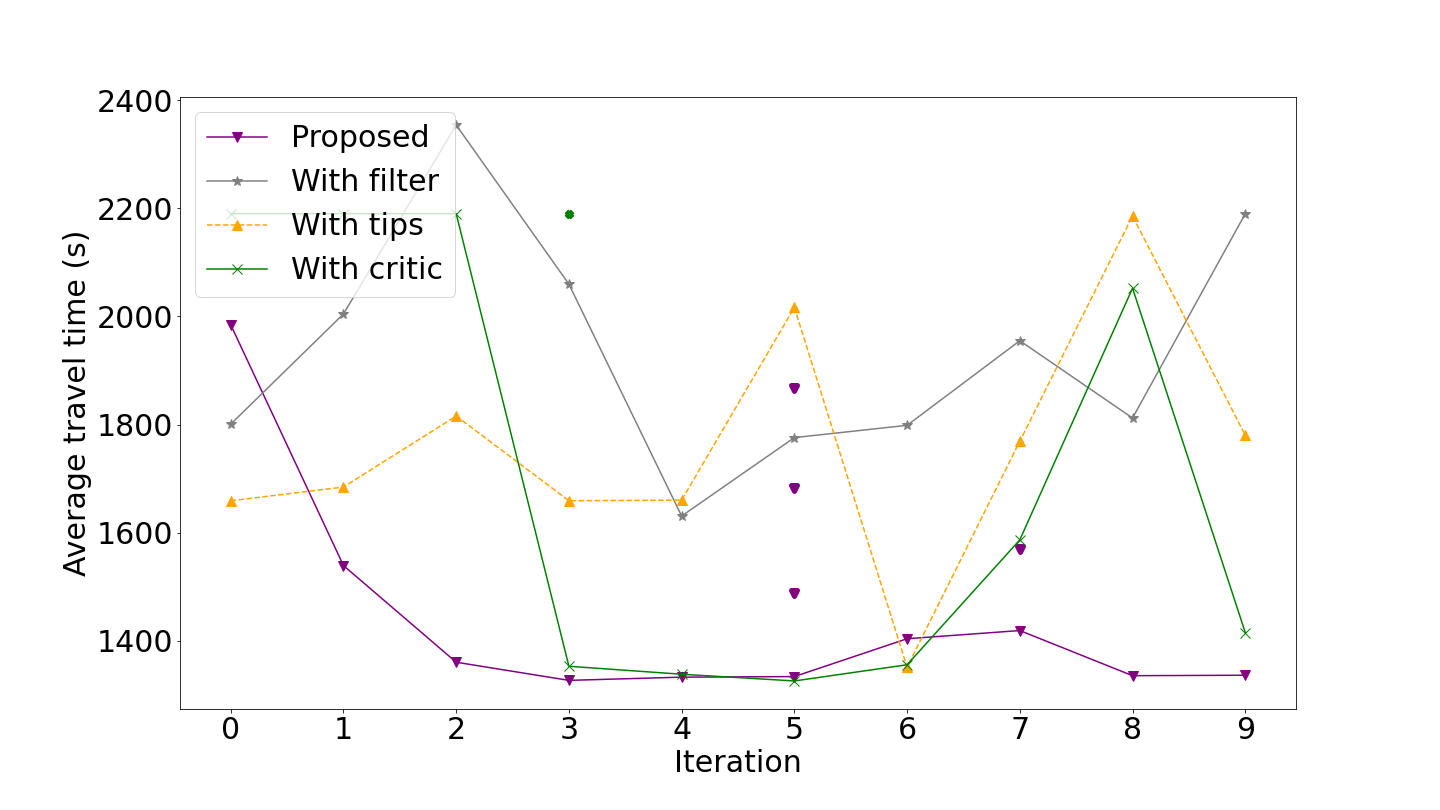}
    \caption{\textcolor{black}{Average travel time.}}
    \label{fig:case1_llm_RL_avg_travel_time}
\end{subfigure}\hspace{-16mm}
\hfill
\begin{subfigure}{0.52\textwidth}
    \includegraphics[width=\textwidth]{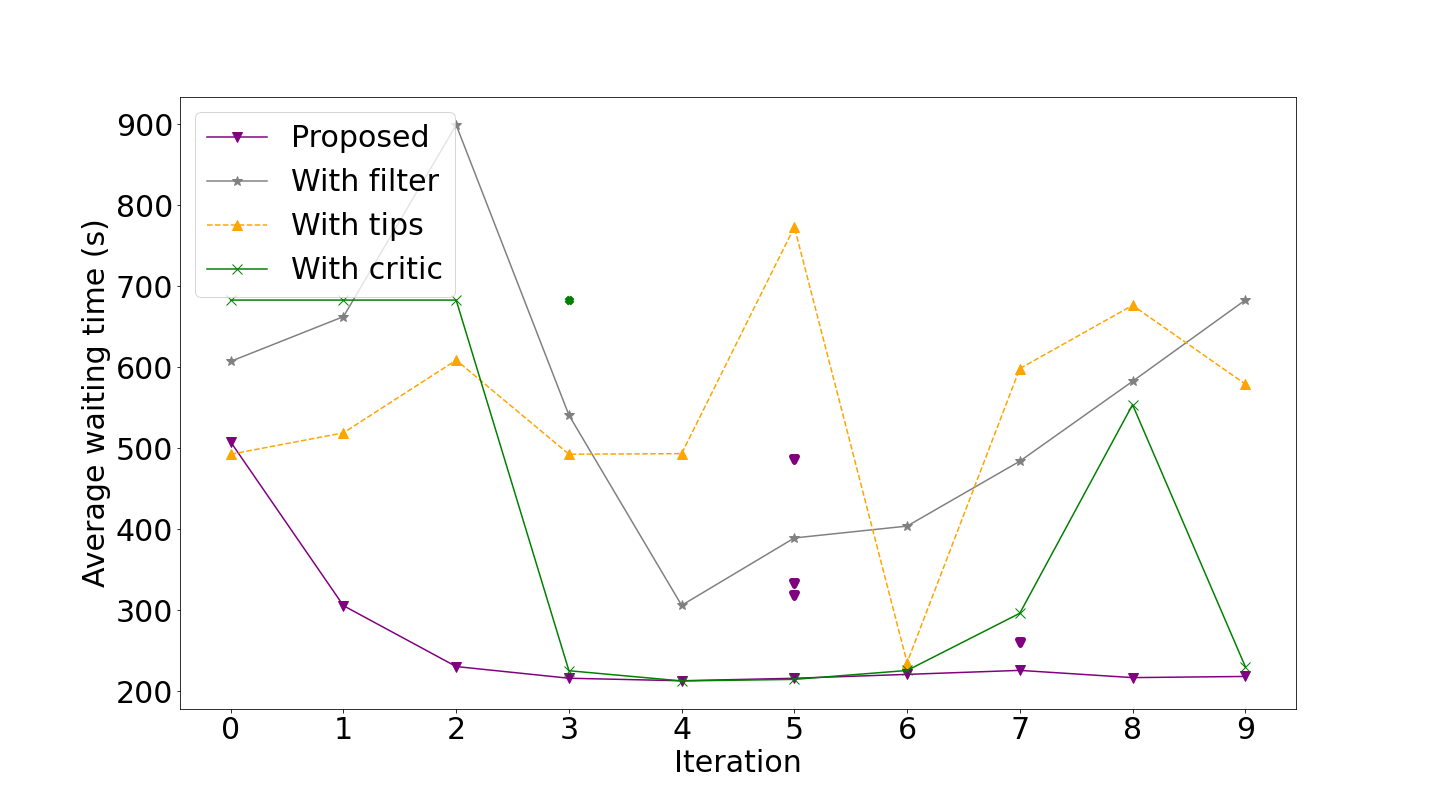}
    \caption{\textcolor{black}{Average waiting time.}}
    \label{fig:case1_llm_RL_avg_waiting_time}
\end{subfigure}
\hfill
\begin{subfigure}{0.52\textwidth}
    \includegraphics[width=\textwidth]{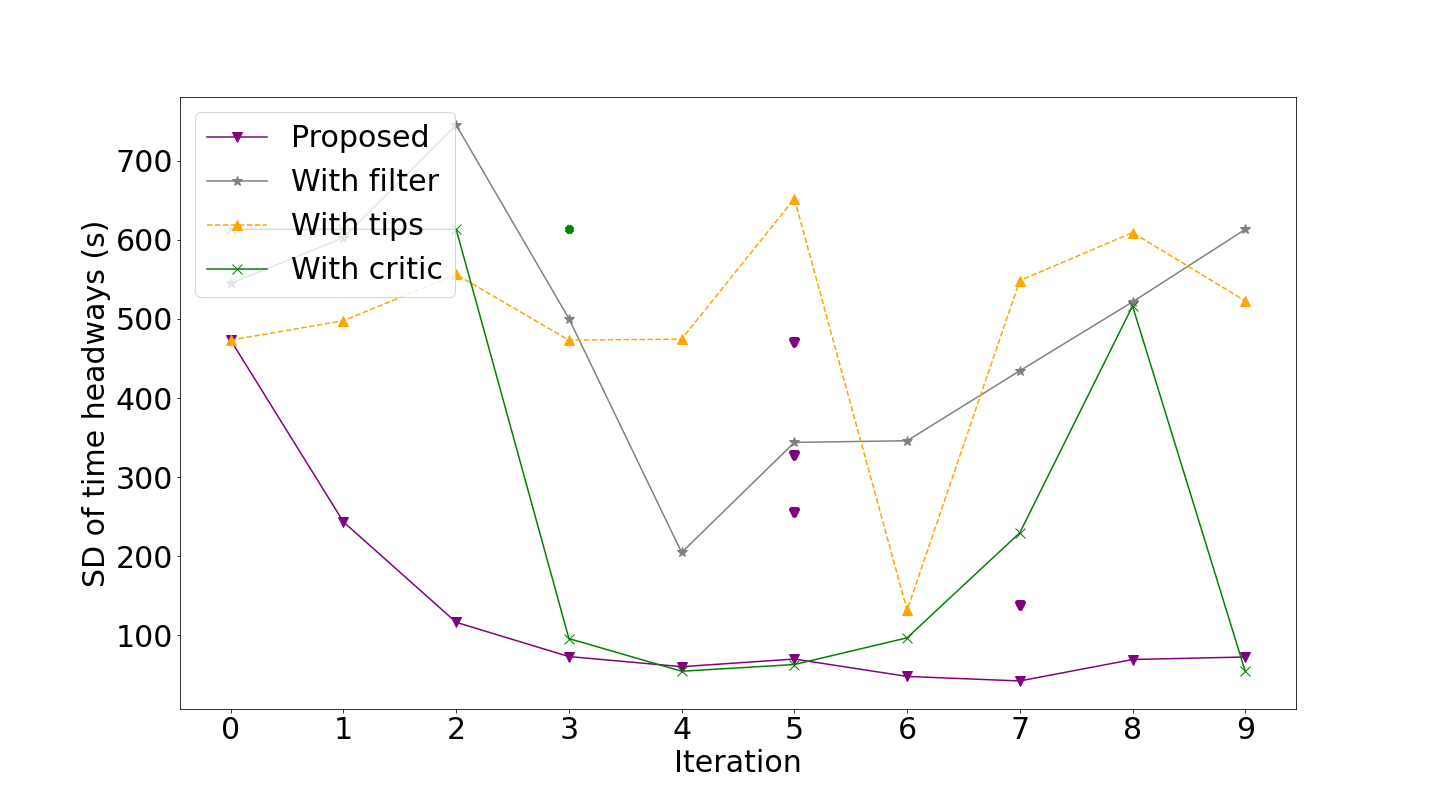}
    \caption{\textcolor{black}{SD of time headways.}}
    \label{fig:case1_llm_RL_SD_time_headways}
\end{subfigure}\hspace{-16mm}
\hfill
\begin{subfigure}{0.52\textwidth}
    \includegraphics[width=\textwidth]{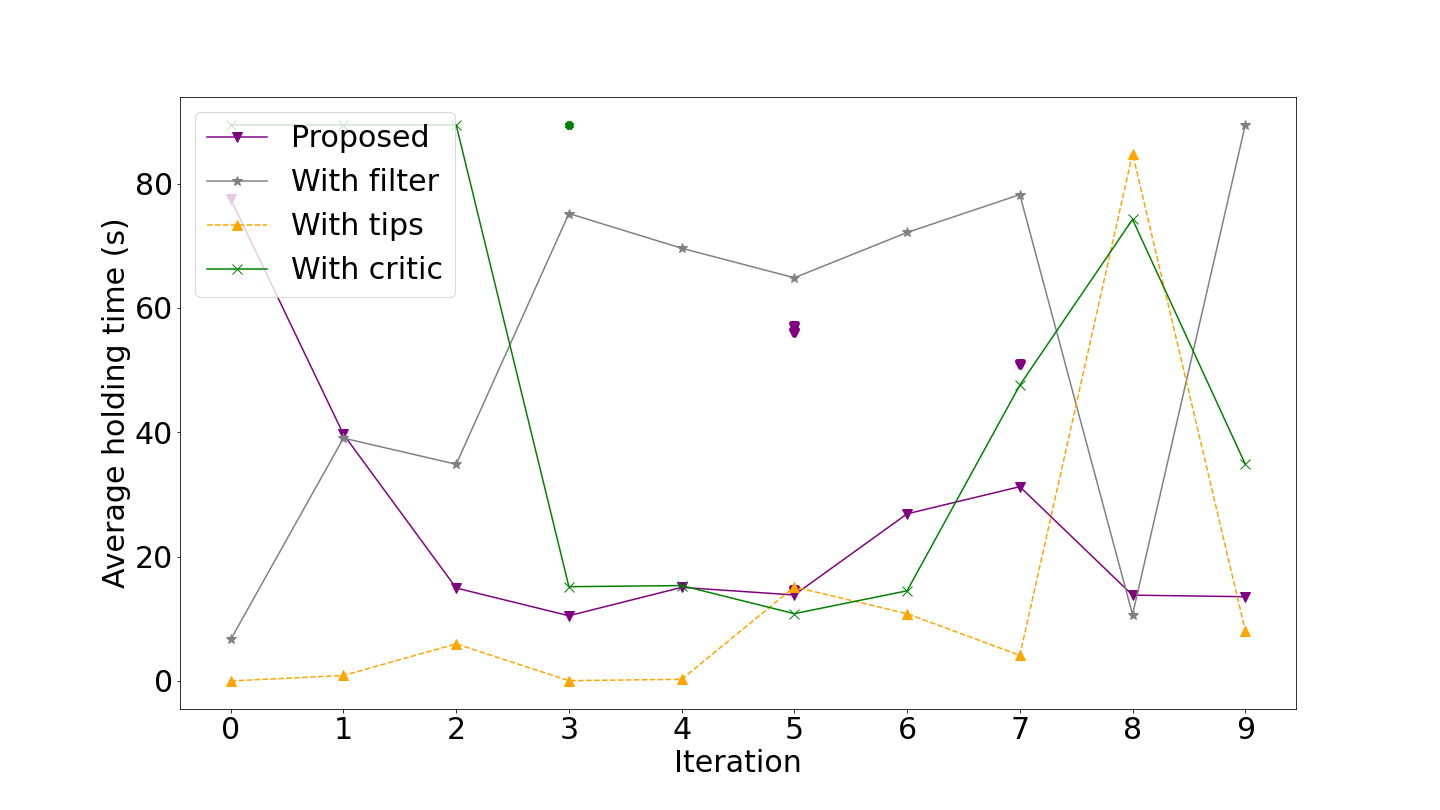}
    \caption{\textcolor{black}{Average holding time.}}
    \label{fig:case1_llm_RL_avg_holding_time}
\end{subfigure}
\caption{\textcolor{black}{Evaluation comparisons with existing LLM-enhanced RL methods.}}
\label{fig:case1_llm_RL}
\end{figure}

Table \ref{table:various_llm_RL} presents the test results. The LLM-enhanced RL with critic achieves the lowest SD of time headways, while the proposed method obtains the minimum average travel time and waiting time. Specifically, the proposed method reduces the average travel time and waiting time by 5.59\% and 5.30\%, respectively, compared to the best-performing baseline among the three existing LLM-enhanced RL methods. Fig. \ref{fig:case1_llm_RL} displays the performance evolutions of RL agents across iterations for each method. The results show that all three existing methods exhibit significant performance fluctuations and fail to ensure stable performance improvement over iterations. In contrast, the proposed method consistently enhances RL agent performance in a stable, stepwise manner and reliably produces an effective final reward function. These findings demonstrate that the proposed LLM-enhanced RL paradigm is more effective and reliable than the existing approaches.

\begin{table}[h]
\centering
\caption{\label{table:various_llm_RL}\textcolor{black}{Evaluations of different LLM-enhanced RL methods}}
\begin{tblr}{
  width = \linewidth,
  colspec = {Q[1.5]Q[1.2]Q[1.1]Q[1.25]Q[1.3]Q[1]},
  cells = {c},
  hline{1-2,6} = {-}{},
}
\textbf{Method} & \textbf{ SD of time headways } & \textbf{ Avg. travel time } & \textbf{ Avg. waiting time } & \textbf{ Avg. holding time }\\
The proposed & 72.38 & \textbf{1335.91} & \textbf{218.27} & 13.55\\
With filter & 613.39 & 2190.19 & 683.09 & 89.42\\
With tips & 522.85 & 1780.02 & 579.39 & 8.10\\
With critic & \textbf{54.08} & 1414.95 & 230.49 & 34.87
\end{tblr}
\end{table}

\subsubsection{Ablation studies}

The effectiveness of each LLM-based module is evaluated through ablation studies. Table \ref{table:ablation} summarizes the test results. The previous trial of the proposed paradigm with a warm start can be considered as an ablation of the reward initializer. The proposed paradigm with a cold start and ``RL -local'' serve as tests with and without all modules, respectively. For the ablation of the analyzer module, the training and test trajectories, along with the test results, are directly embedded in the input prompts of the reward modifier and reward refiner modules. When the reward refiner module is ablated, the reward function generated by the reward modifier is not checked against the refiner's criterion and is used directly in the next iteration.

\begin{table}[h]
\centering
\caption{\label{table:ablation}Evaluations of ablation study}
\begin{threeparttable}
\begin{tblr}{
  width = \linewidth,
  colspec = {Q[2]Q[1.8]Q[1.5]Q[1.5]Q[1.5]Q[2]Q[2]Q[2.1]Q[2]},
  cells = {c},
  hline{1,7} = {-}{0.08em},
  hline{2} = {-}{},
}
 & Initializer & Modifier & Refiner & Analyzer & SD-TH & Avg. TT & Avg. WT & Avg. HT\\
RL- local &  &  &  &  & 188.81 & 1443.77 & 269.38 & 21.41\\
Ablate analyzer & \Checkmark & \Checkmark & \Checkmark &  & 265.79 & 1543.91 & 320.82 & 25.62\\
Ablate refiner & \Checkmark & \Checkmark &  & \Checkmark & 161.90 & 1511.76 & 265.37 & 35.36\\
Warm start &  & \Checkmark & \Checkmark & \Checkmark & \textbf{62.66} & \textbf{1331.27} & \textbf{212.27} & 14.63\\
Cold start & \Checkmark & \Checkmark & \Checkmark & \Checkmark & 72.38 & 1335.91 & 218.27 & 13.55\\
\end{tblr}
\begin{tablenotes}
      \small
      \item Note: TT, WT, and HT represent travel time, waiting time, and holding time, respectively. SD-TH represents SD of time headways. 
    \end{tablenotes}
\end{threeparttable}
\end{table}

According to Table \ref{table:ablation}, the proposed approach with a warm start outperforms other strategies regarding SD of time headways, average travel time, and average waiting time. The cold start version also shows performance similar to the warm start. When ablating the analyzer and reward refiner, the average travel time increases by 15.97\% and 13.56\%, respectively, compared to the warm start version, indicating the necessity of the analyzer and reward refiner modules in the proposed LLM-enhanced RL paradigm. Regarding the reward initializer, its ablation does not impact the final test performance of the proposed paradigm, but this module is essential when there is no expert input to provide a starting reward function. Therefore, the reward initializer is critical for bypassing the extensive manual trial-and-error typically required in designing the reward function.

\begin{figure}[h]
\centering
\begin{subfigure}{0.52\textwidth}
    \includegraphics[width=\textwidth]{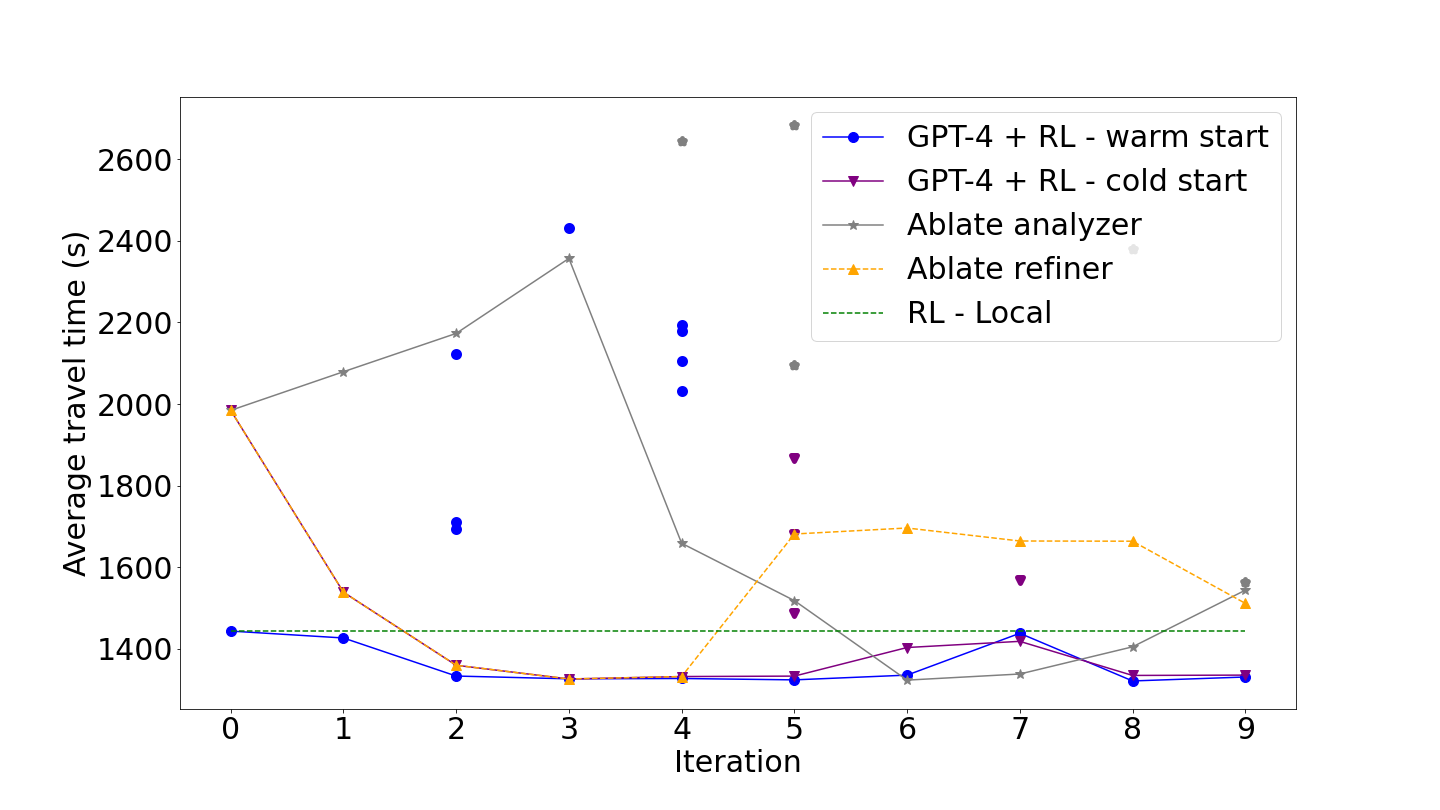}
    \caption{Average travel time.}
    \label{fig:case1_ablation_avg_travel_time}
\end{subfigure}\hspace{-16mm}
\hfill
\begin{subfigure}{0.52\textwidth}
    \includegraphics[width=\textwidth]{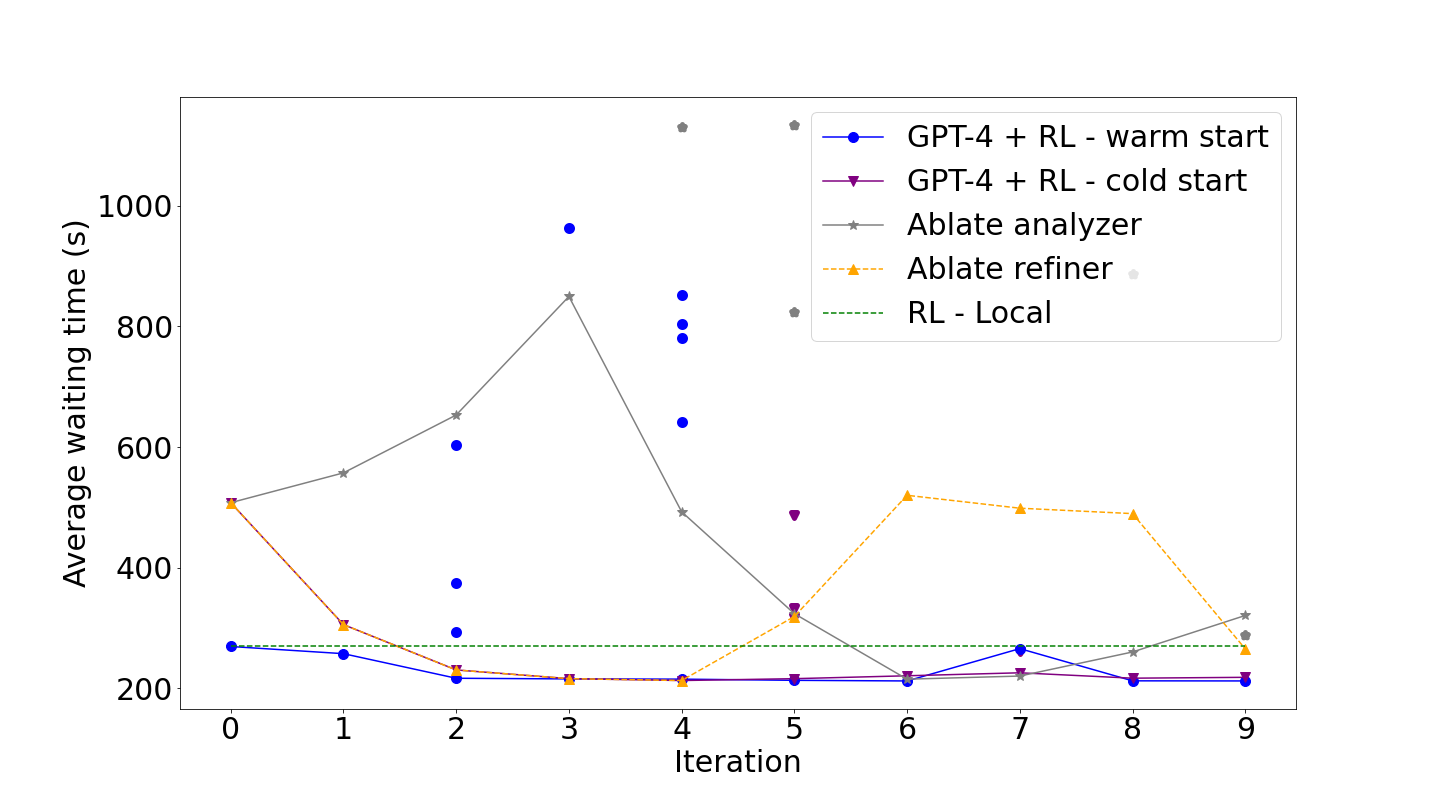}
    \caption{Average waiting time.}
    \label{fig:case1_ablation_avg_waiting_time}
\end{subfigure}
\hfill
\begin{subfigure}{0.52\textwidth}
    \includegraphics[width=\textwidth]{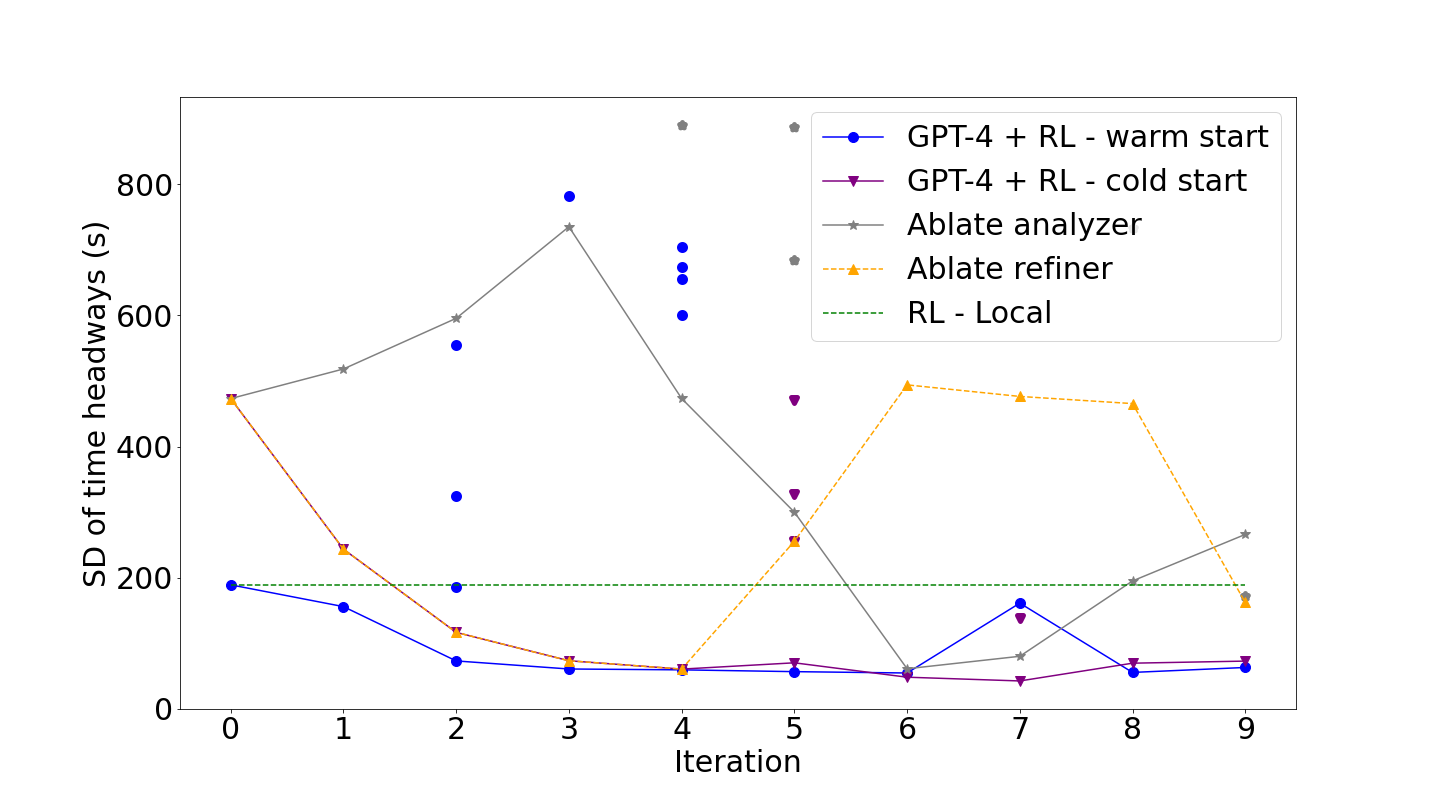}
    \caption{SD of time headways.}
    \label{fig:case1_ablation_SD_time_headways}
\end{subfigure}\hspace{-16mm}
\hfill
\begin{subfigure}{0.52\textwidth}
    \includegraphics[width=\textwidth]{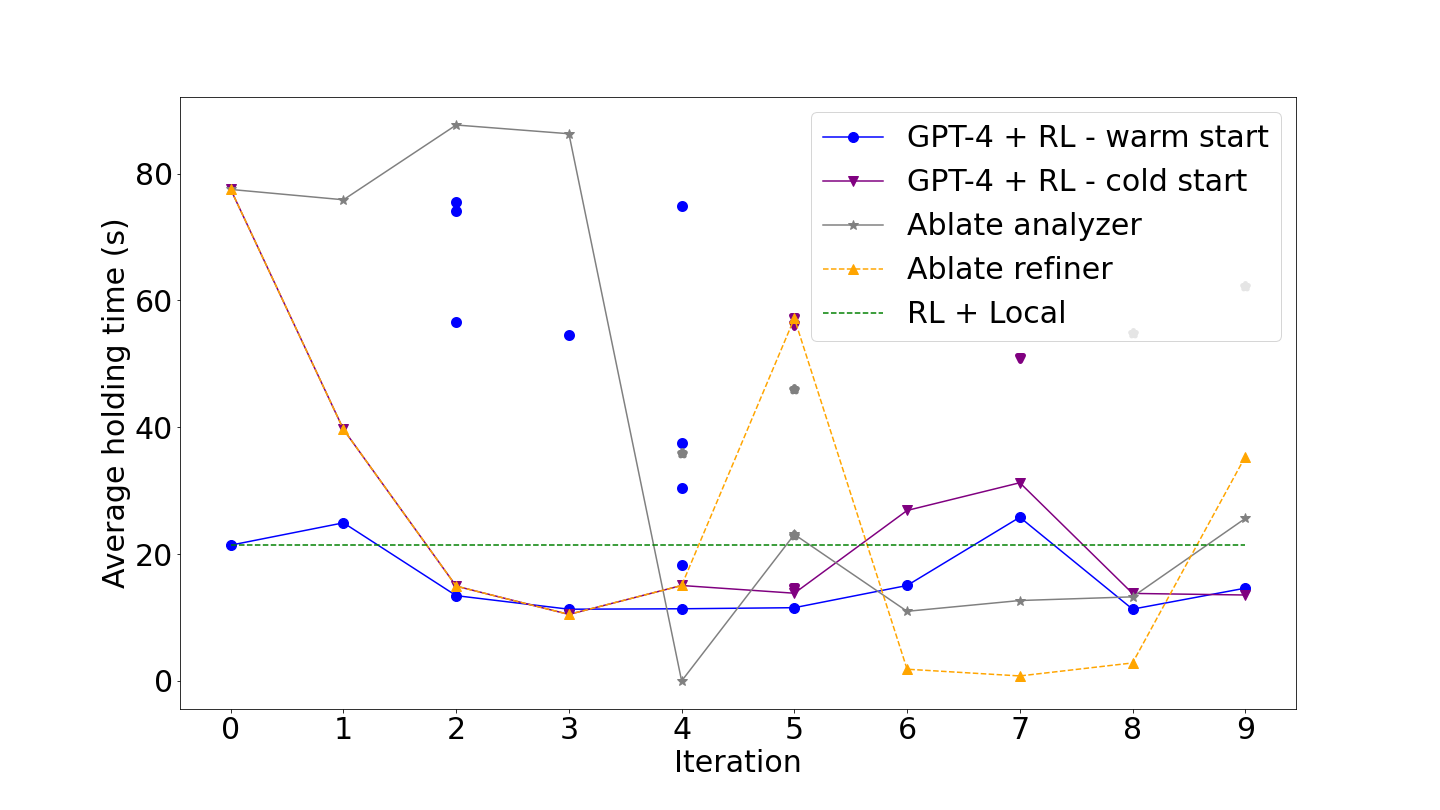}
    \caption{Average holding time.}
    \label{fig:case1_ablation_avg_holding_time}
\end{subfigure}
\caption{Evaluation comparisons with module ablation.}
\label{fig:case1_ablation}
\end{figure}

Fig. \ref{fig:case1_ablation} exhibits the performance curves of all strategies related to the ablation studies. Large fluctuations are observed when the analyzer and reward refiner are ablated, indicating that these modules are essential for ensuring stable improvement and convergence of the test performance.

\clearpage
\subsection{Case study 2: Two lines with shared corridor}

In real-world bus holding control scenarios, it is essential to consider the randomness of bus speed and travel time between adjacent stops. Additionally, the bus capacity and dwell time, influenced by passengers' boarding and alighting times, must be accurately modeled within the bus holding control strategy. This case study addresses these factors comprehensively and tests the effectiveness of the proposed paradigm in the multi-line system.

\subsubsection{Scenario description}

\begin{figure}[h]
\centering
\includegraphics[width=0.75\linewidth]{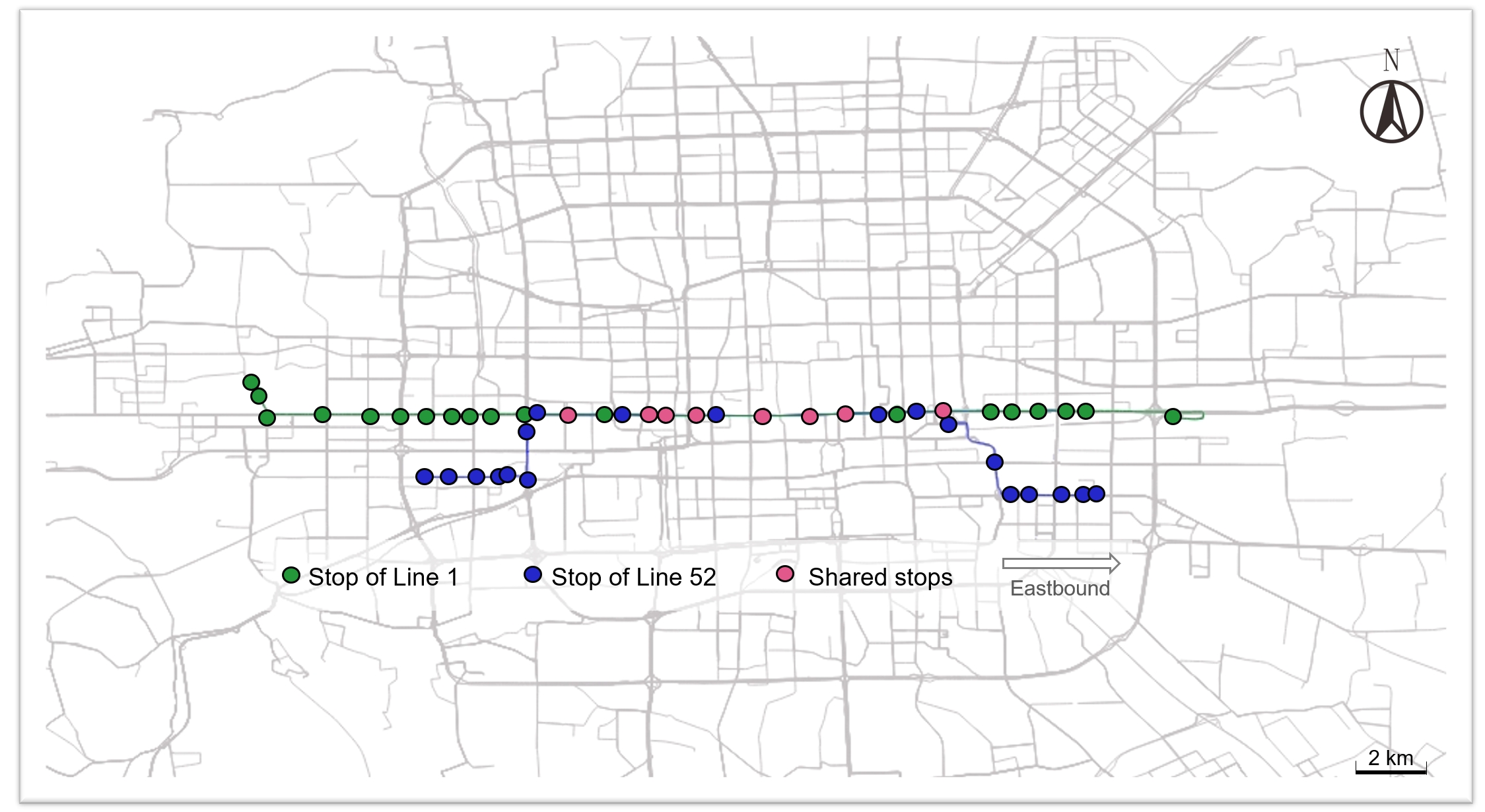}
\caption{\label{fig:busline_case2}Bus lines in Case 2}
\end{figure}

\begin{figure}
\centering
\begin{subfigure}{0.5\textwidth}
    \includegraphics[width=\textwidth]{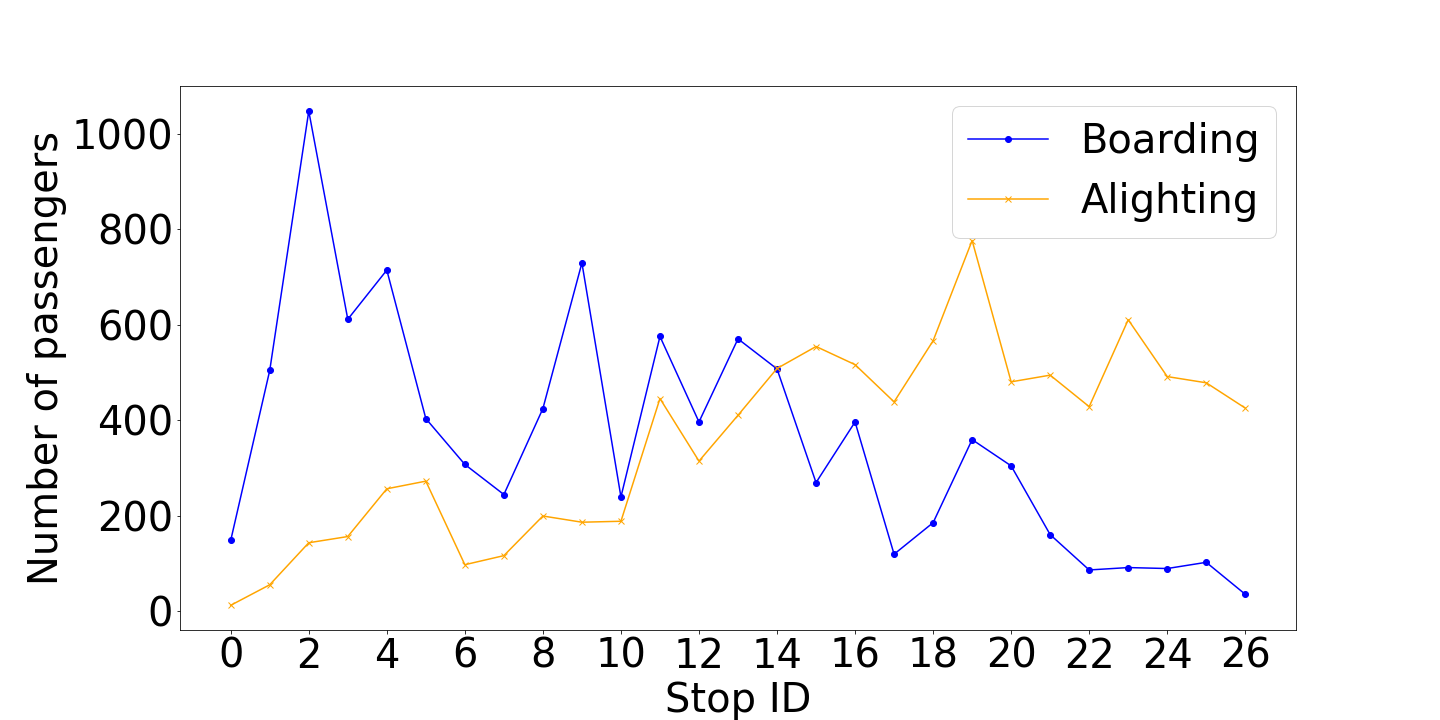}
    \caption{Bus line 1.}
    \label{fig:case2_passenger_demand_3436}
\end{subfigure}\hspace{-16mm}
\hfill
\begin{subfigure}{0.5\textwidth}
    \includegraphics[width=\textwidth]{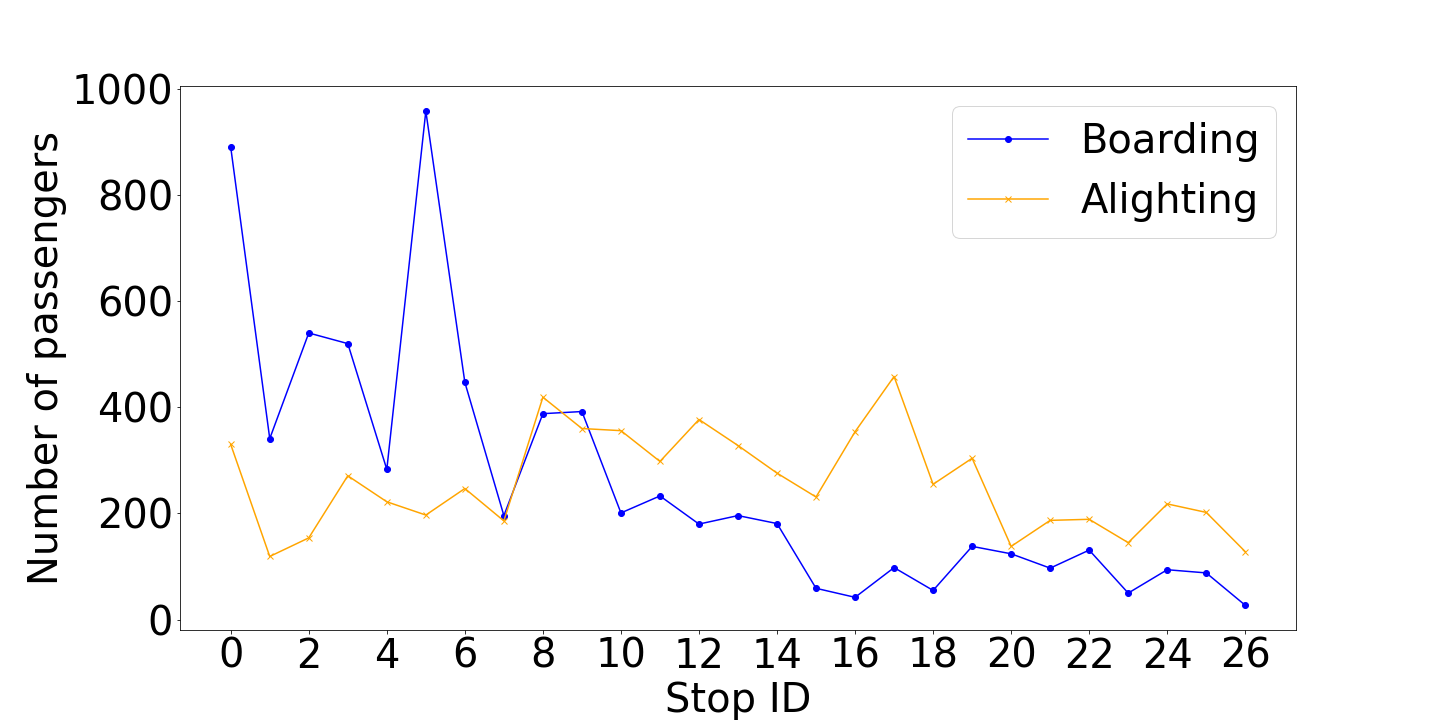}
    \caption{Bus line 52.}
    \label{fig:case2_passenger_demand_3986}
\end{subfigure}
\caption{Spatial distribution of passenger demand in Case 2.}
\label{fig:case2_passenger_demand}
\end{figure}

\begin{figure}[h]
\centering
\begin{subfigure}{0.5\textwidth}
    \includegraphics[width=\textwidth]{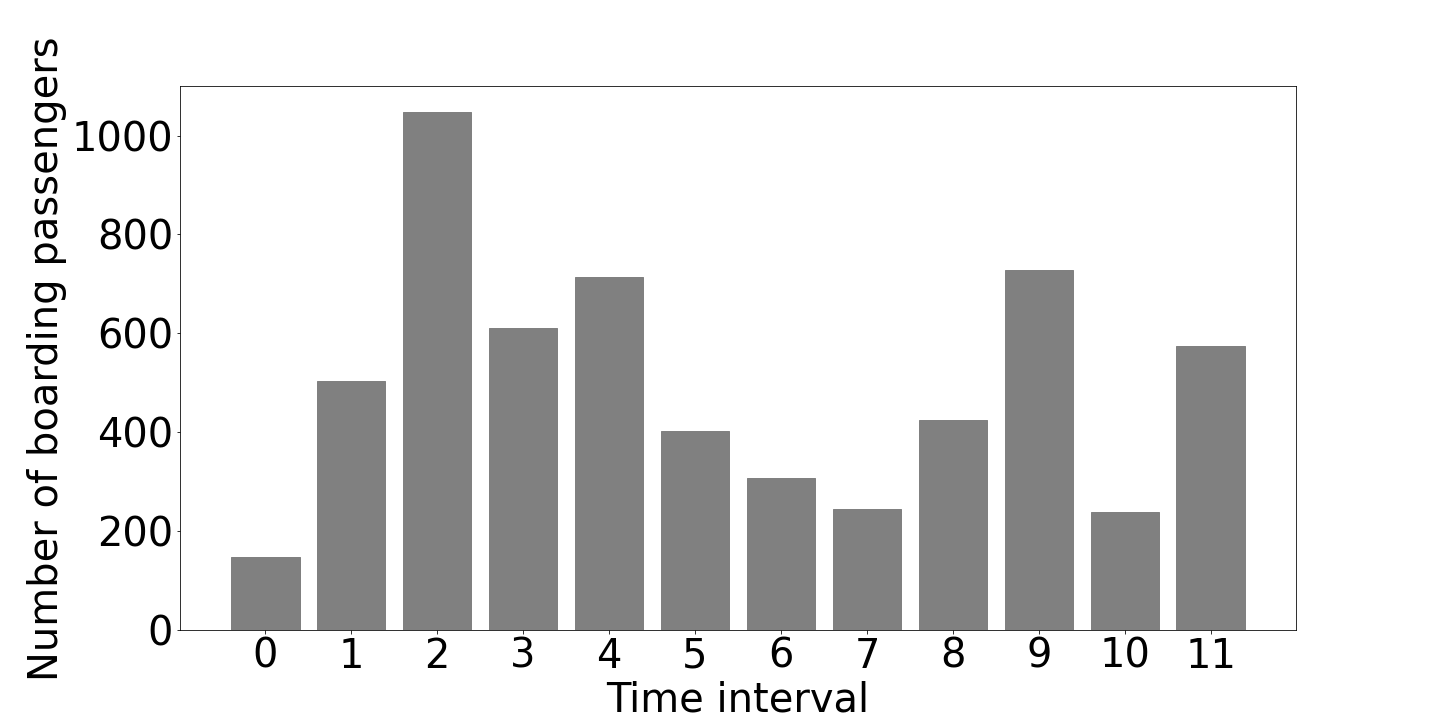}
    \caption{Bus line 1.}
    \label{fig:case2_passenger_demand_per_20min_3436}
\end{subfigure}\hspace{-16mm}
\hfill
\begin{subfigure}{0.5\textwidth}
    \includegraphics[width=\textwidth]{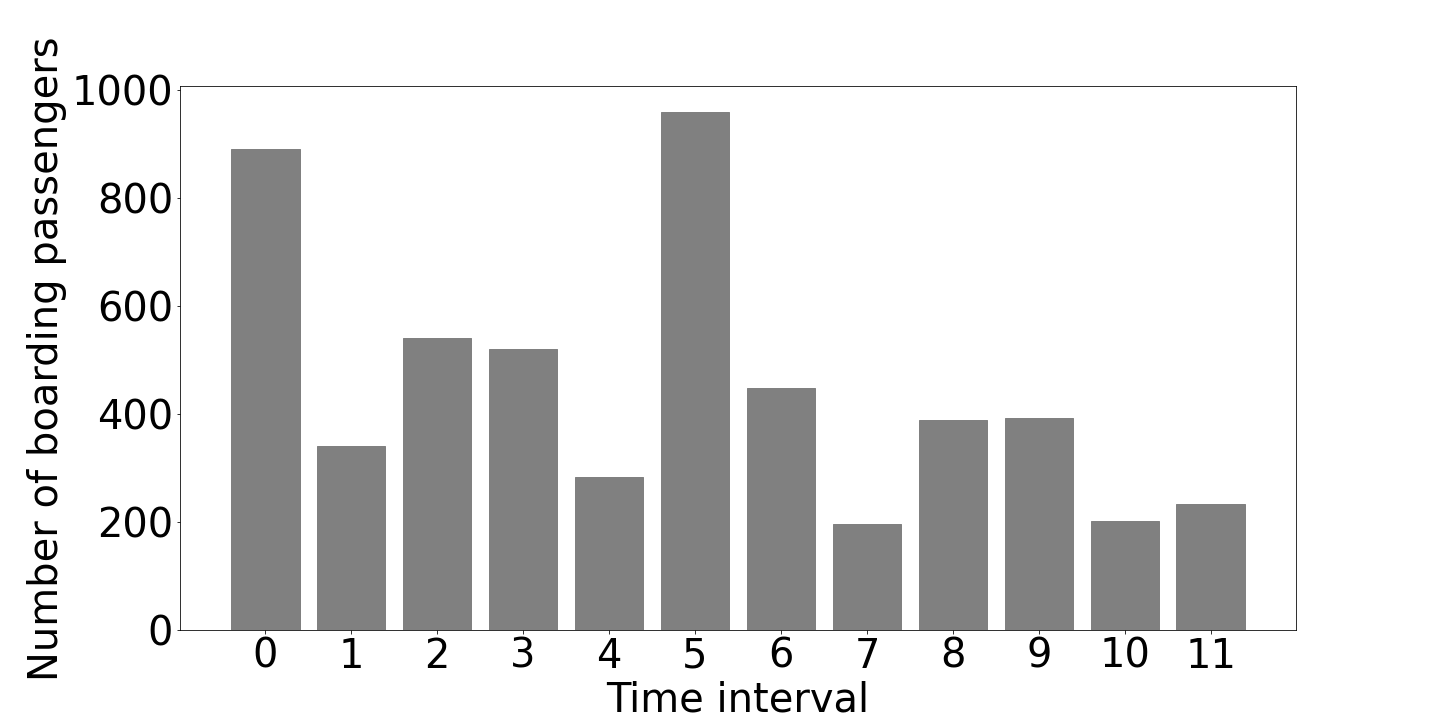}
    \caption{Bus line 52.}
    \label{fig:case2_passenger_demand_per_20min_3986}
\end{subfigure}
\caption{Temporal distribution of passenger demand in Case 2.}
\label{fig:case2_passenger_demand_per_20min}
\end{figure}

In case study 2, we test the proposed LLM-enhanced RL paradigm in a real-world bus system located in Beijing, China. Fig. \ref{fig:busline_case2} depicts the structure of the bus system. It consists of two bus lines (Line 1 and Line 52) with a partially shared corridor. There are 27 bus stops on each bus line and 8 shared stops for the two lines. The length of Line 1 and Line 52 are 24.95 km and 22.58 km, respectively. Buses depart every 5 min within each line. We utilize eastbound passengers' IC card data collected during the morning rush hour (from 6 am to 10 am) on a weekday (Tuesday, May 14th, 2019) to accomplish this numerical test. The number of boarding and alighting passengers at each stop is displayed in Fig. \ref{fig:case2_passenger_demand}. The temporal distribution of boarding passenger demand is shown in Fig. \ref{fig:case2_passenger_demand_per_20min}, in which each time interval represents 20 minutes. 

The capacity of buses is 120 pax \citep{wang2023partc}. The buses' travel time between two adjacent stops is randomly generated following a Gamma distribution with the mean travel time derived from real-world data \citep{berrebi2018partc}. The travel times for different buses are generated independently and distinct from each other. Overtaking is allowed in this scenario. Passengers' arrivals follow Poisson processes, with arrival rates randomly generated within an interval 10\% higher or lower than the real-world demand. The random seeds for passengers' demand and buses' travel time differ across episodes. The origin and destination stops of passengers are based on real-world data. Passengers with both origin and destination at shared bus stops are categorized as shared passengers. In our simulation, shared passengers can take a bus from either bus line to complete their trip. The bus dwell time is determined by the maximum of passengers' boarding and alighting times. Boarding and alighting times are proportional to the number of boarding and alighting passengers, with each boarding passenger taking 3 seconds and each alighting passenger taking 1.8 seconds \citep{wang2020partc}.

\subsubsection{Comparisons with baselines}

Table \ref{table:baseline_case2} summarizes the test performances of the proposed paradigm with a cold start and baseline strategies. Each control method was evaluated using 30 different random seeds to mitigate the effects of stochasticity. The final reward function derived from the proposed LLM-enhanced RL paradigm is detailed and explained in \ref{appendix:reward_function_case2}. The proposed LLM-enhanced RL paradigm achieves minimum average travel time for each bus line and the overall system. Compared to the no-holding scenario, ``GPT-4 + RL'', feedback control, and ``RL - local'' reduce the overall average travel time by 3.14\%, 1.69\%, and 1.30\%, respectively. ``GPT-4 + RL'' significantly enhances improvement compared to vanilla RL controllers and feedback control. Notably, the average holding times of Line 1 and Line 52 in the proposed control paradigm are 6.52 and 6.61 seconds, respectively, which are 43.45\% and 36.81\% less than those in feedback control. These comparisons between ``GPT-4 + RL'' and feedback control indicate that the proposed LLM-enhanced RL paradigm is a more efficient holding control strategy, performing better on average travel time with fewer holding interventions in the bus system. ``RL - global'' and ``RL - local + global'' fail to control the system and obtain the same or similar performance with the no-holding scenario. 

\begin{figure}[h]
\centering
\begin{subfigure}{0.35\textwidth}
    \includegraphics[width=\textwidth]{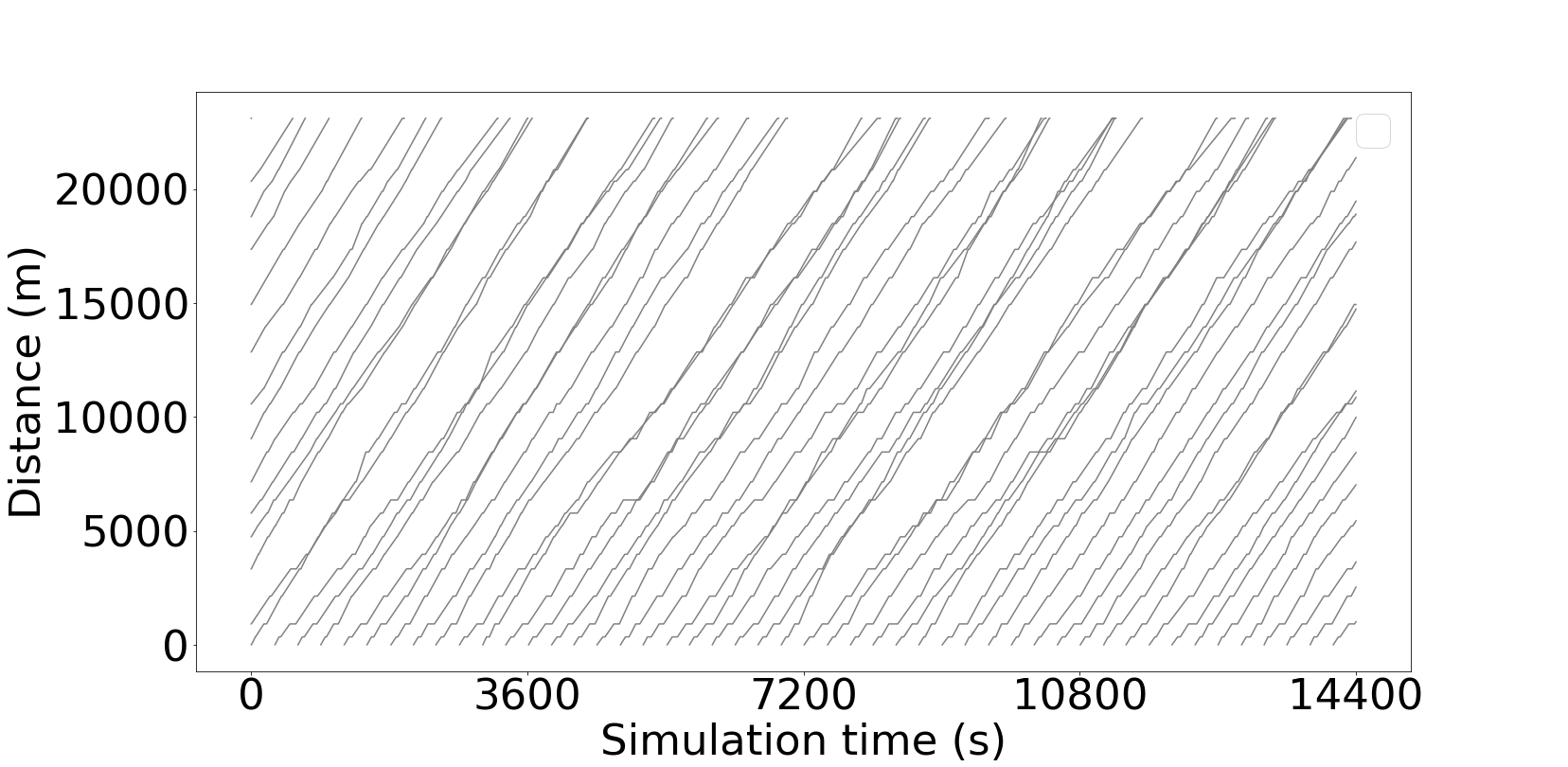}
    \caption{No holding (Line 1).}
    \label{fig:case2_noholding_trajectory_3436}
\end{subfigure}\hspace{-8mm}
\hfill
\begin{subfigure}{0.35\textwidth}
    \includegraphics[width=\textwidth]{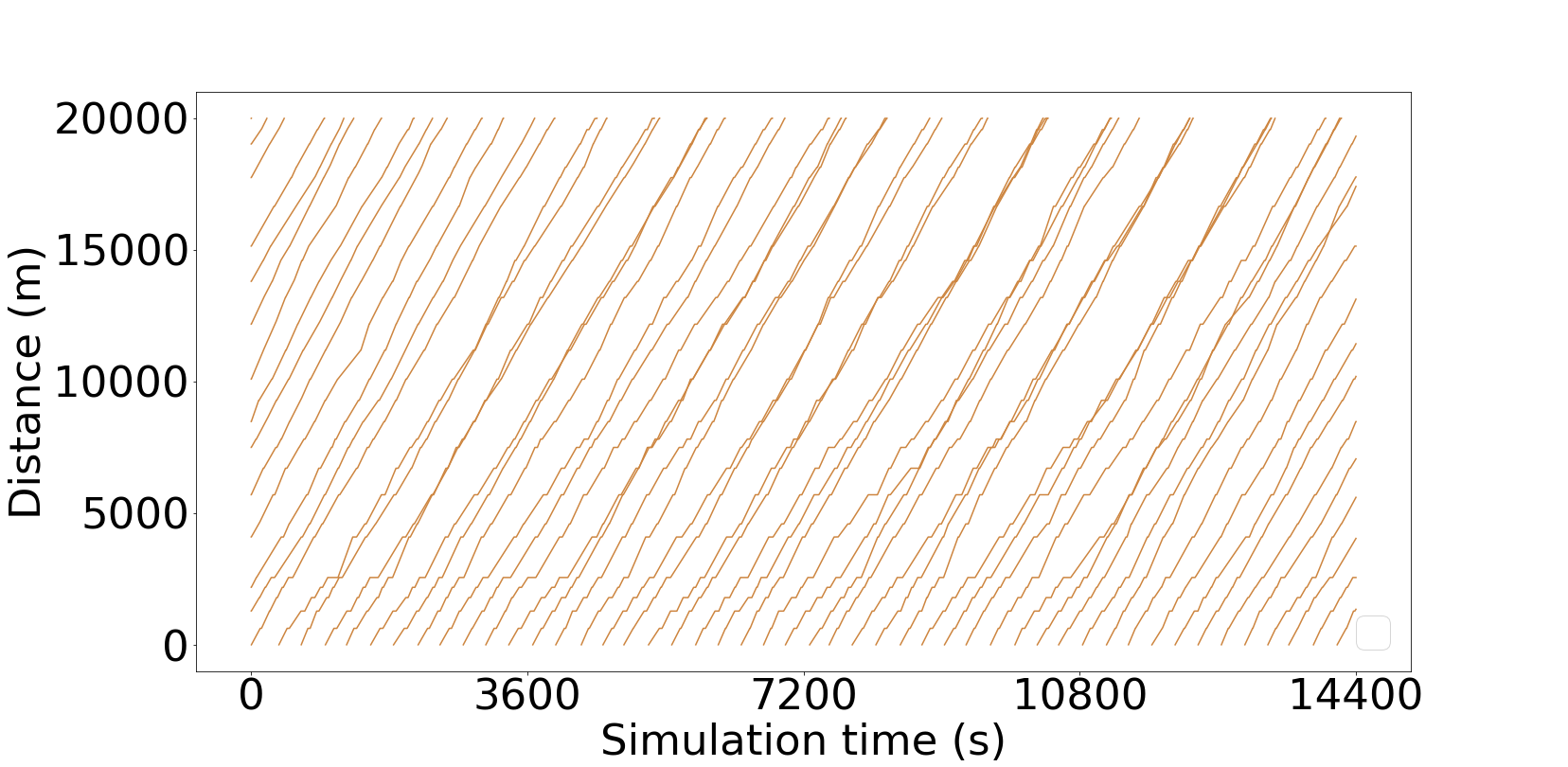}
    \caption{No holding (Line 52).}
    \label{fig:case2_noholding_trajectory_3986}
\end{subfigure}\hspace{-8mm}
\hfill
\begin{subfigure}{0.35\textwidth}
    \includegraphics[width=\textwidth]{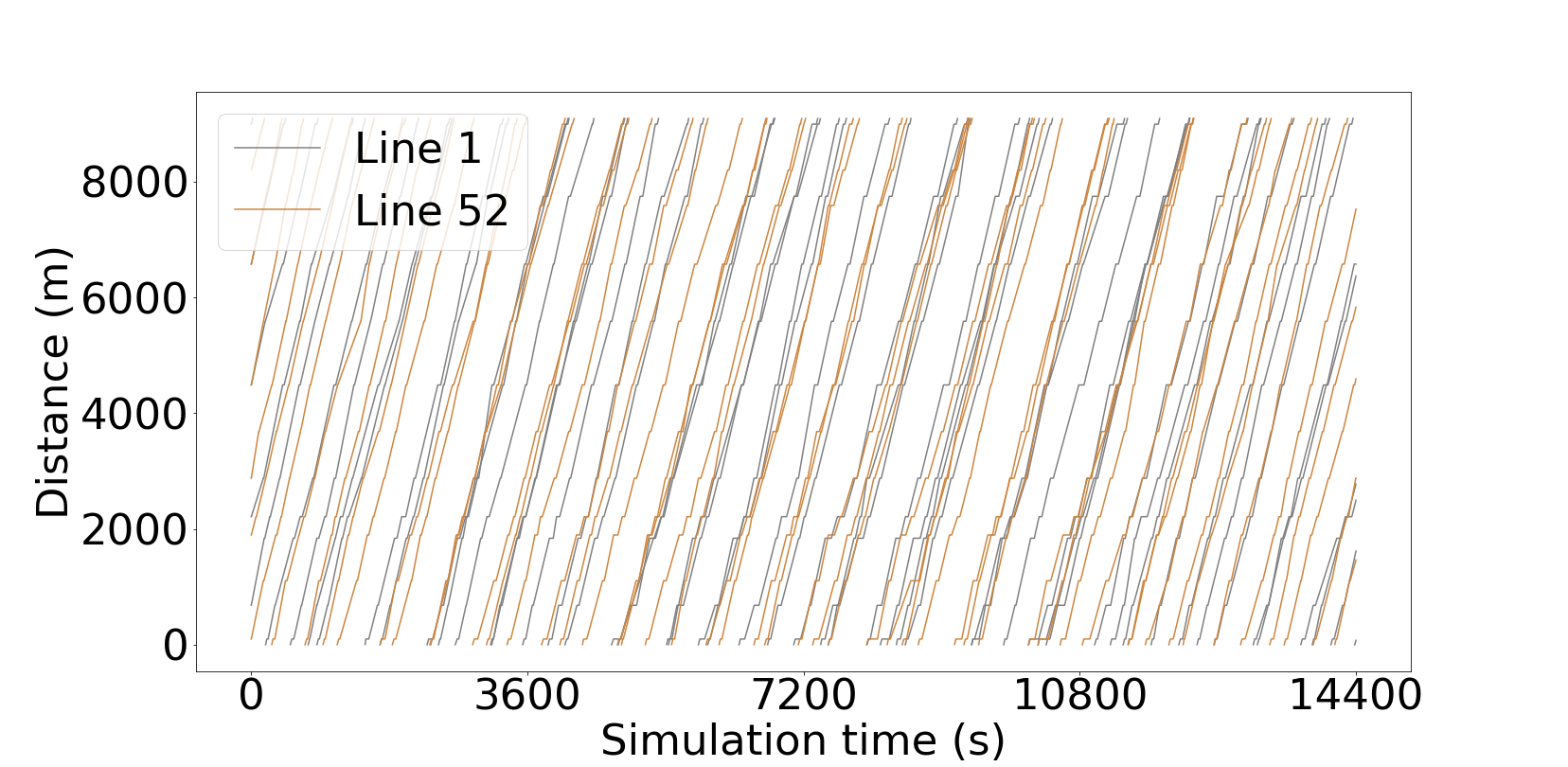}
    \caption{No holding (shared corridor).}
    \label{fig:case2_noholding_common_line_trajectory}
\end{subfigure}
\hfill
\begin{subfigure}{0.35\textwidth}
    \includegraphics[width=\textwidth]{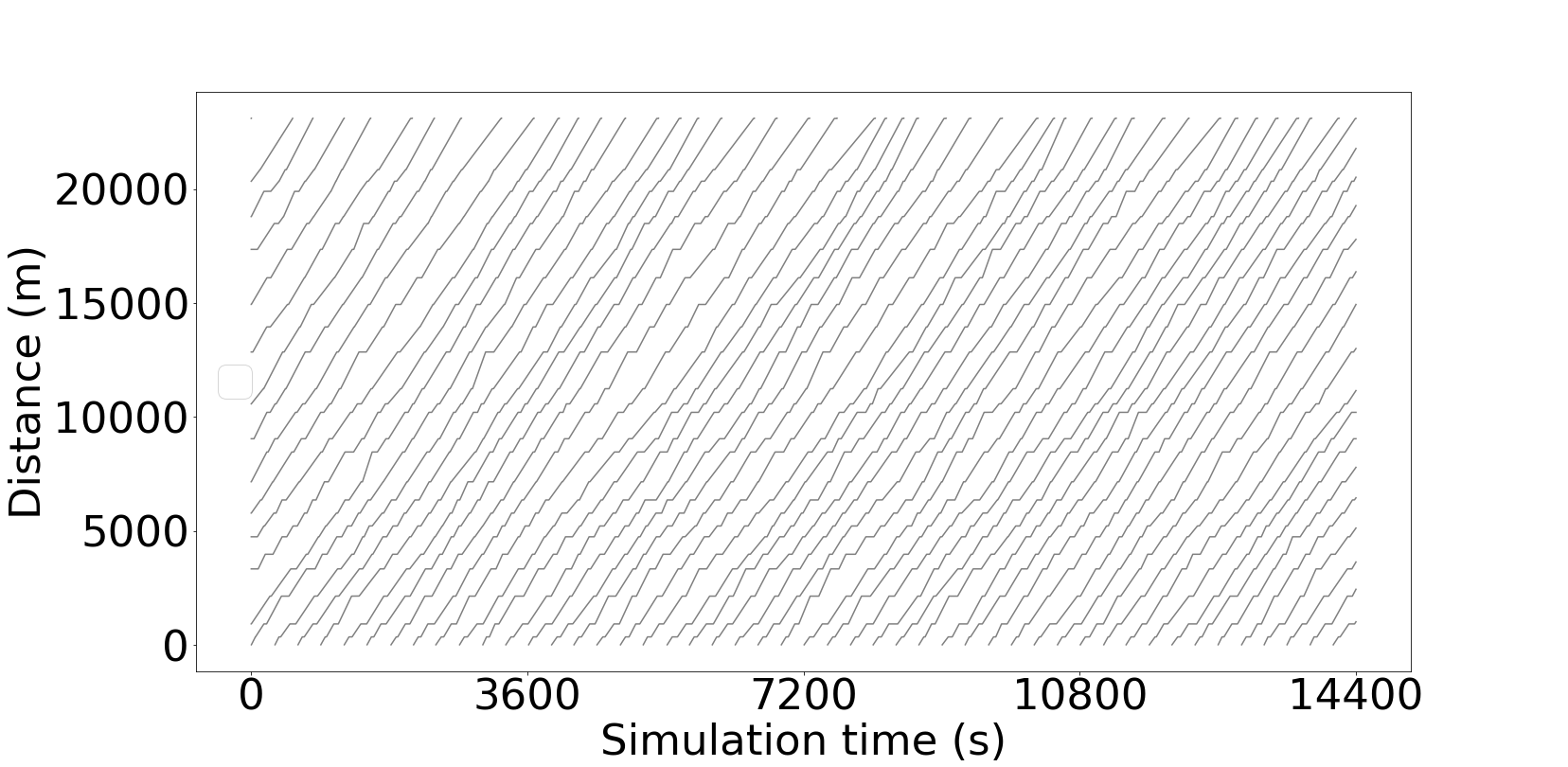}
    \caption{Feedback control (Line 1).}
    \label{fig:case2_feedback_control_trajectory_3436}
\end{subfigure}\hspace{-8mm}
\hfill
\begin{subfigure}{0.35\textwidth}
    \includegraphics[width=\textwidth]{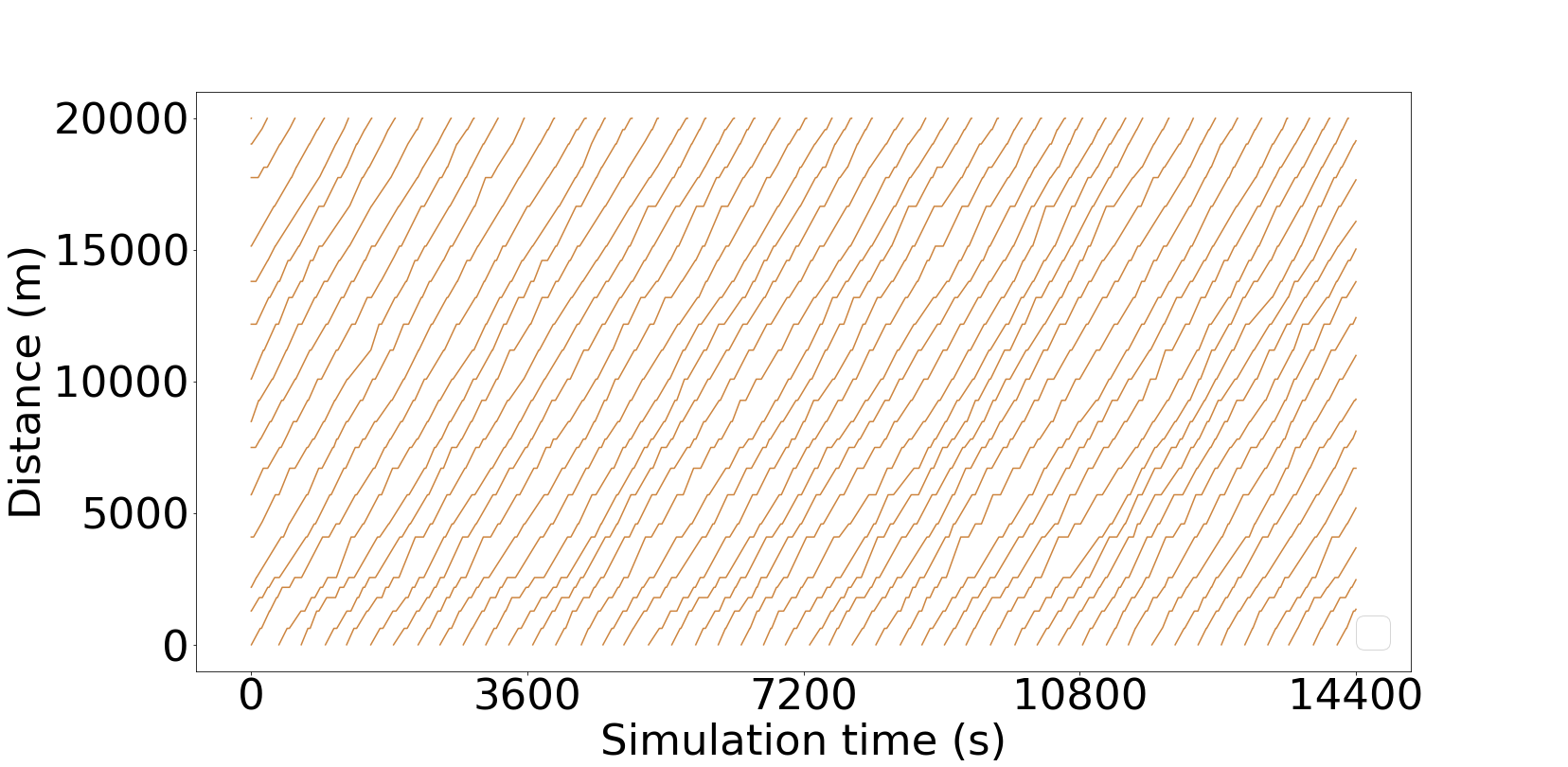}
    \caption{Feedback control (Line 52).}
    \label{fig:case2_feedback_control_trajectory_3986}
\end{subfigure}\hspace{-8mm}
\hfill
\begin{subfigure}{0.35\textwidth}
    \includegraphics[width=\textwidth]{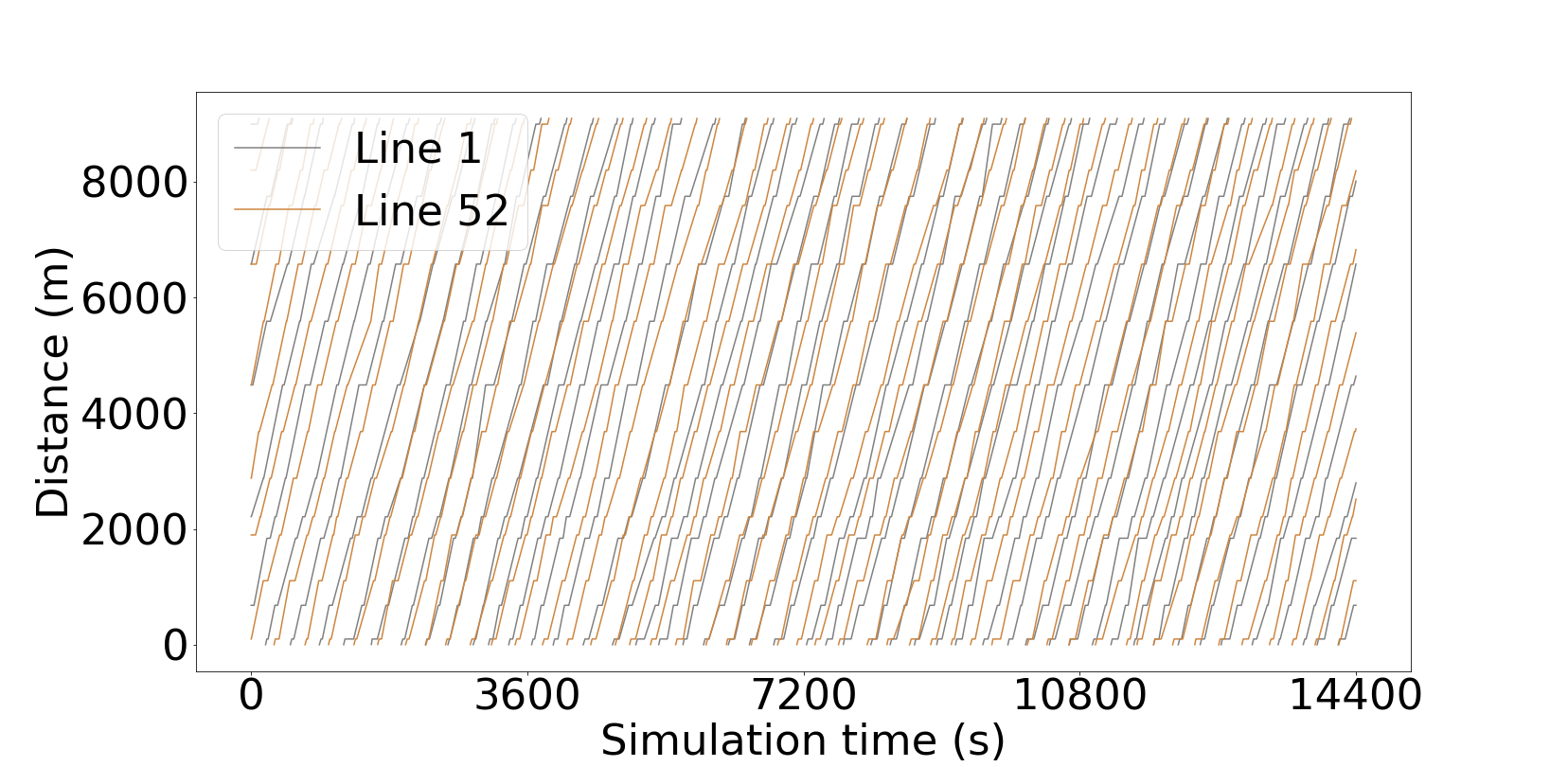}
    \caption{Feedback control (shared corridor).}
    \label{fig:case2_feedback_control_common_line_trajectory}
\end{subfigure}
\hfill
\begin{subfigure}{0.35\textwidth}
    \includegraphics[width=\textwidth]{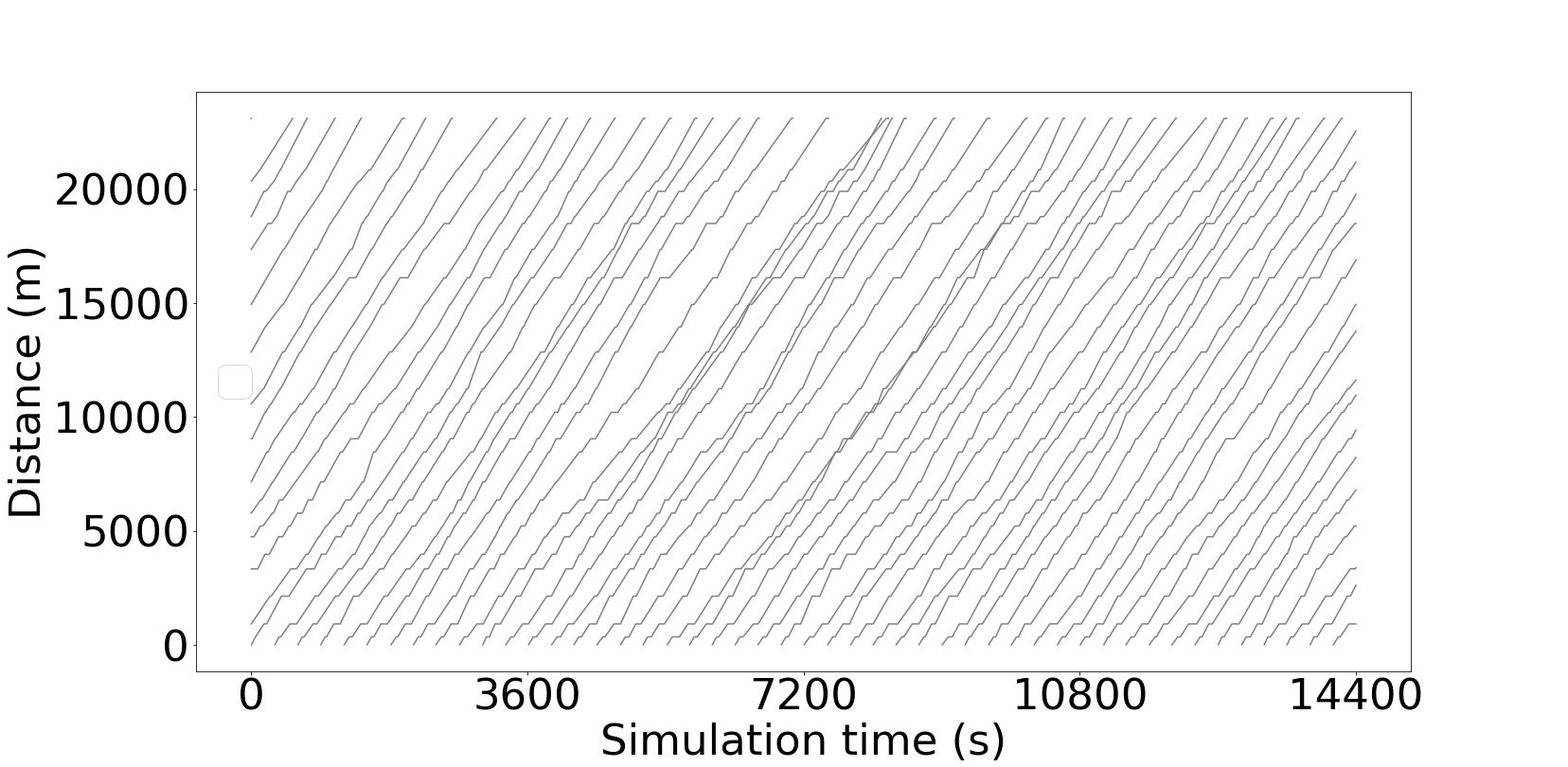}
    \caption{GPT4 + RL (Line 1).}
    \label{fig:case2_gpt4_RL_trajectory_3436}
\end{subfigure}\hspace{-8mm}
\hfill
\begin{subfigure}{0.35\textwidth}
    \includegraphics[width=\textwidth]{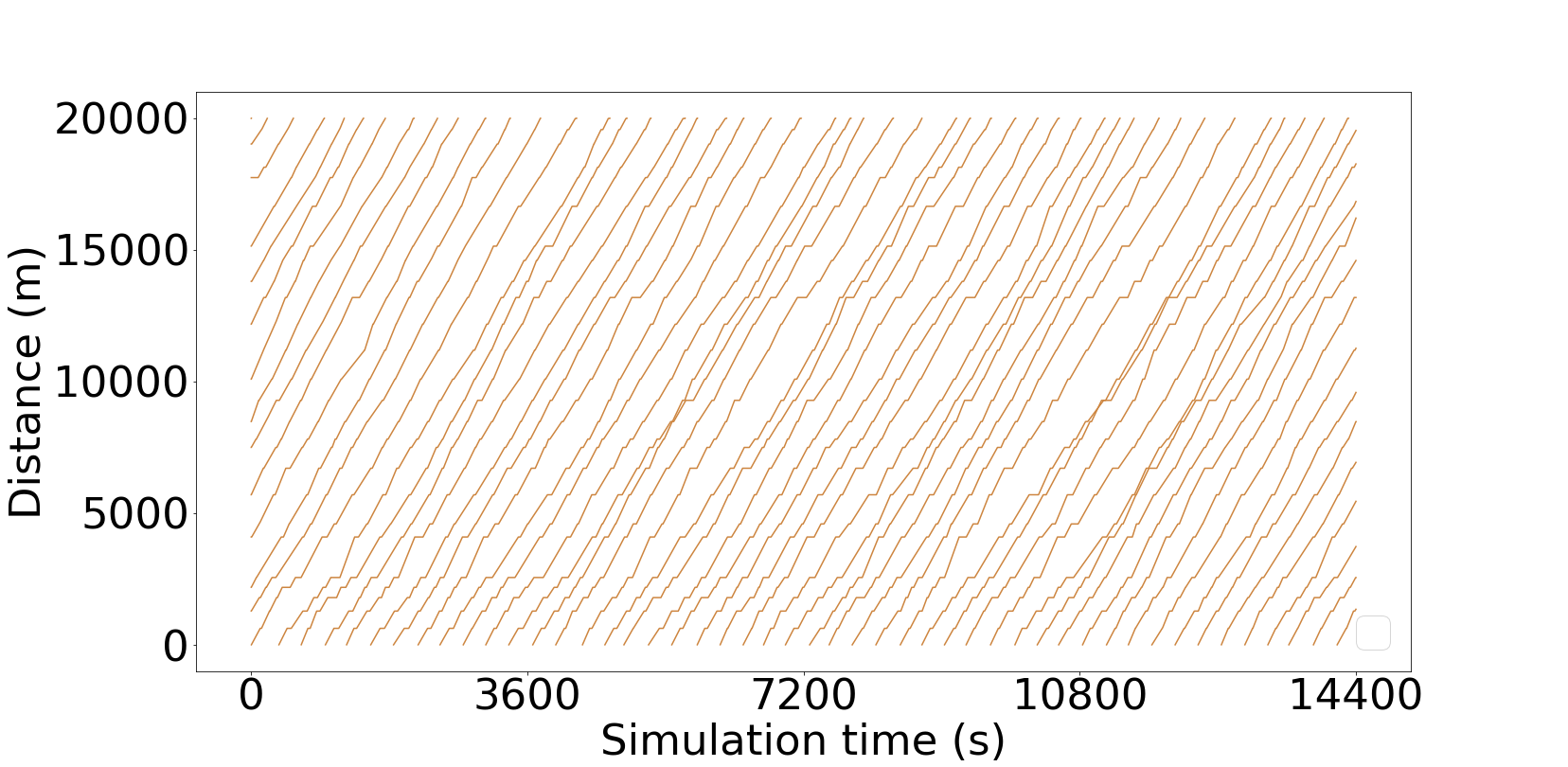}
    \caption{GPT4 + RL (Line 52).}
    \label{fig:case2_gpt4_RL_trajectory_3986}
\end{subfigure}\hspace{-8mm}
\hfill
\begin{subfigure}{0.35\textwidth}
    \includegraphics[width=\textwidth]{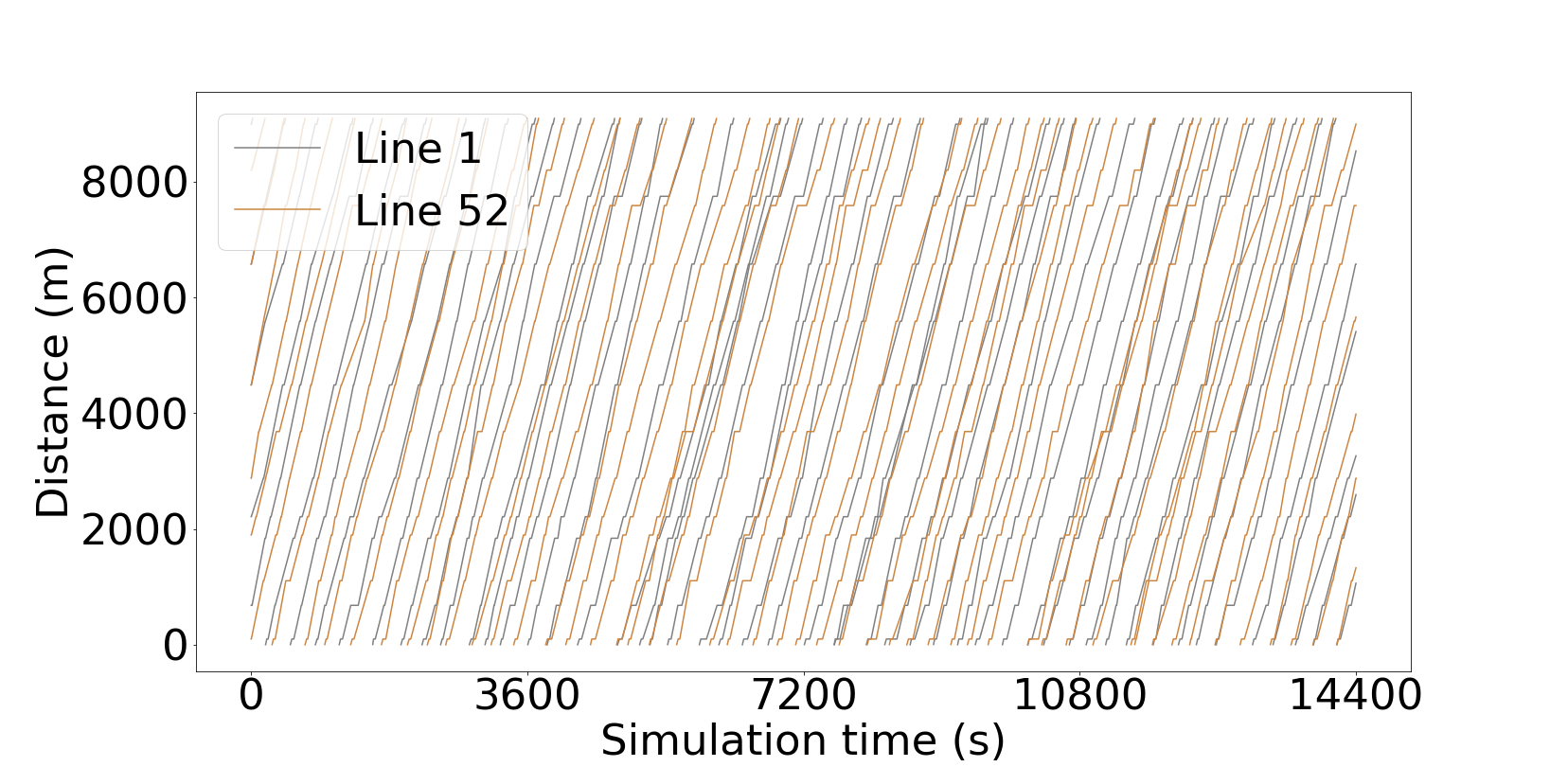}
    \caption{GPT4 + RL (shared corridor).}
    \label{fig:case2_gpt4_RL_common_line_trajectory}
\end{subfigure}
\caption{Evaluation comparisons with module ablation.}
\label{fig:case2_bus_trajectories}
\end{figure}

Regarding the separate performance of Line 1 and Line 52, ``GPT-4 + RL'' still achieves the minimum average travel time for both lines. Since feedback control focuses solely on the single objective of headway balancing, it provides the minimum SD of time headways and thus the average waiting time for both lines, as expected. Fig. \ref{fig:case2_bus_trajectories} exhibits the bus trajectories for the no-holding scenario, feedback control, and ``GPT-4 + RL''. \textcolor{black}{To simplify the presentation, all subsequent figures and sensitive analysis are based on the test results from the simulation using a random seed of 70.} As the statistical analysis suggests, feedback control can homogenize the headways perfectly. The proposed LLM-enhanced RL paradigm does not regulate the headways as \textcolor{black}{balanced} as feedback control but finds a better trade-off between headway consistency and holding delay, resulting in the minimum average travel time for the bus system. Regarding the shared corridor and shared passengers, it is counter-intuitive that the no-holding scenario achieves the minimum average travel time. None of the holding control strategies can improve the performance of the shared corridor.

\textcolor{black}{As outlined in \ref{appendix:reward_function_case2}, the reward function in the proposed paradigm consists of five components, and only two components directly address headway balancing. The RL agent's goal in the proposed paradigm is not only to balance headways but also to find a tradeoff between balancing headways and minimizing holding delays. As a result, occasional bus bunching is observed in the bus trajectories (Fig. \ref{fig:case2_gpt4_RL_trajectory_3436} and \ref{fig:case2_gpt4_RL_trajectory_3986}). In contrast, the feedback control mechanism focuses solely on balancing headways, so bus bunching does not appear in its trajectories (Fig. \ref{fig:case2_feedback_control_trajectory_3436} and \ref{fig:case2_feedback_control_trajectory_3986}). Although the proposed approach results in more instances of bus bunching compared to feedback control, it still significantly outperforms the no-holding scenario in mitigating bus bunching and slightly reduces the average travel time compared to feedback control.}

Normally, with effective holding control strategies, the waiting time of passengers can be reduced compared to the no-holding scenario, but the in-vehicle travel time might increase due to the holding delay. However, compared to the no-holding scenario, the average travel time in ``GPT-4 + RL'' is reduced significantly more than the waiting time. This may be attributed to the large dwell delay in the no-holding scenario. Specifically, severe bus bunching in the no-holding scenario causes large numbers of passengers to board delayed buses, which are always highly occupied. Consequently, the dwell delay affects the in-vehicle travel time for more passengers in the no-holding scenario, resulting in a larger average travel time. 

Besides the displayed indicators, it's notable that 12.03\% of passengers are unable to complete their trip in the no-holding scenario during the 4-hour simulation, due to large time headways and bus capacity limits. In contrast, the percentage of passengers that can not complete their trip is significantly lower, at only 1.07\%, in the proposed LLM-enhanced RL paradigm. This improvement indicates a substantial enhancement in the overall service quality of the bus system.


\begin{sidewaystable}[h]
\centering
\caption{\label{table:baseline_case2}\textcolor{black}{Evaluations of different control methods in Case 2}}
\begin{threeparttable}
\begin{tblr}{
  width = \linewidth,
  colspec = {Q[160]Q[120]Q[125]Q[92]Q[83]Q[27]Q[120]Q[125]Q[92]Q[83]},
  cells = {c},
  cell{1}{1} = {r=2}{},
  cell{2}{2} = {c=4}{0.401\linewidth},
  cell{2}{7} = {c=4}{0.384\linewidth},
  cell{10}{2} = {c=4}{0.401\linewidth},
  cell{10}{7} = {c=4}{0.384\linewidth},
  hline{1,10,18} = {-}{},
  hline{2,11} = {2-5,7-10}{},
  hline{3} = {1-5,7-10}{},
}
\textbf{Control method} & \textbf{SD of time headways} & \textbf{Avg. TT} & \textbf{Avg. WT} & \textbf{Avg. HT} & ~ & \textbf{SD of time headways} & \textbf{Avg. TT} & \textbf{Avg. WT} & \textbf{Avg. HT}\\
 & \textbf{Line 1} &  &  &  & ~ & \textbf{Line 52} &  &  & \\
\textbf{GPT-4 + RL} & 101.05$\pm$7.09 & \textbf{1642.52}$\pm$22.74 & 163.69$\pm$3.59 & 6.52$\pm$0.60 & ~ & 86.08$\pm$7.04 & \textbf{1189.79}$\pm$15.04 & 153.65$\pm$2.37 & 6.61$\pm$0.59\\
\textbf{Feedback} & \textbf{58.45}$\pm$3.06 & 1666.37$\pm$31.66 & \textbf{156.55}$\pm$3.10 & 11.53$\pm$0.54 & ~ & \textbf{51.51}$\pm$2.86 & 1208.63$\pm$13.41 & \textbf{149.15}$\pm$1.76 & 10.46$\pm$0.49\\
\textbf{Model-based control} & 120.86$\pm$9.03 & 1681.69$\pm$35.91 & 172.46$\pm$3.92 & 4.41$\pm$0.39 & ~ & 103.10$\pm$11.28 & 1212.30$\pm$16.75 & 157.65$\pm$2.38 & 6.11$\pm$0.42\\
\textbf{No holding} & 216.79$\pm$19.72 & 1723.33$\pm$58.27 & 194.58$\pm$6.97 & 0$\pm$0 & ~ & 201.75$\pm$16.09 & 1197.23$\pm$22.30 & 173.69$\pm$4.60 & 0$\pm$0\\
\textbf{RL-local} & 82.49$\pm$6.61 & 1661.80$\pm$38.65 & 161.13$\pm$3.56 & 16.28$\pm$0.75 & ~ & 69.24$\pm$5.27 & 1227.20$\pm$14.25 & 151.49$\pm$1.45 & 22.76$\pm$0.51 \\
\textbf{RL-global} & 223.40$\pm$21.40 & 1727.91$\pm$58.90 & 197.28$\pm$7.84 & 1.46$\pm$0.15 & \textbf{~} & 199.68$\pm$15.88 & 1194.48$\pm$23.50 & 172.67$\pm$4.76 & 1.58$\pm$0.20\\
\textbf{RL-local+global} & 216.79$\pm$19.72 & 1723.33$\pm$58.27 & 194.58$\pm$6.97 & 0$\pm$0 & \textbf{~} & 201.72$\pm$16.07 & 1197.23$\pm$22.31 & 173.68$\pm$4.60 & 0.01$\pm$0.02\\
\textbf{~} & \textbf{Overall} &  &  &  & ~ & \textbf{Shared stops} &  &  & \\
\textbf{GPT-4 + RL} & - & \textbf{1443.13}$\pm$14.58 & 159.27$\pm$2.41 & - & ~ & 102.55$\pm$5.69 & \textbf{1282.96}$\pm$58.83 & 111.61$\pm$4.97 & -\\
\textbf{Feedback} & - & 1464.64$\pm$18.49 & \textbf{153.29}$\pm$2.06 & - & ~ & 97.10$\pm$6.24 & 1318.15$\pm$59.41 & 109.12$\pm$5.54 & -\\
\textbf{Model-based control} & - & 1474.67$\pm$24.06 & 165.93$\pm$2.36 & - & ~ & 104.54$\pm$6.15 & 1312.87$\pm$54.69 & 116.15$\pm$5.75 & - \\
\textbf{No holding} & - & 1489.85$\pm$38.94 & 185.30$\pm$4.87 & - & ~ & 132.42$\pm$7.09 & 1305.03$\pm$65.23 & 135.19$\pm$7.32 & - \\
\textbf{RL-local} & - & 1470.52$\pm$22.33 & 156.89$\pm$2.22 & - & ~ & \textbf{94.17}$\pm$4.40 & 1283.75$\pm$48.72 & \textbf{107.58}$\pm$3.65 & - \\
\textbf{RL-global} & - & 1490.92$\pm$39.44 & 186.34$\pm$5.23 & - & ~ & 131.34$\pm$7.66 & 1309.49$\pm$70.50 & 135.29$\pm$6.94 & -\\
\textbf{RL-local+global} & - & 1489.85$\pm$38.95 & 185.30$\pm$4.87 & - & \textbf{~} &132.42$\pm$7.09 & 1305.03$\pm$65.23 & 135.19$\pm$7.32 & -\\
\end{tblr}
\begin{tablenotes}
      \small
      \item Note: TT, WT, and HT represent travel time, waiting time, and holding time, respectively. SD-TH represents SD of time headways. 
    \end{tablenotes}
\end{threeparttable}
\end{sidewaystable}

\subsubsection{Sensitive analysis}

Sensitive analyses are conducted to evaluate the robustness of the RL agents trained with the reward function provided by the proposed paradigm. These analyses focus on two aspects: shared passenger percentages and passenger demand levels.

\textbf{(1) Percentage of shared passengers}

\begin{figure}[h]
\centering
\begin{subfigure}{0.52\textwidth}
    \includegraphics[width=\textwidth]{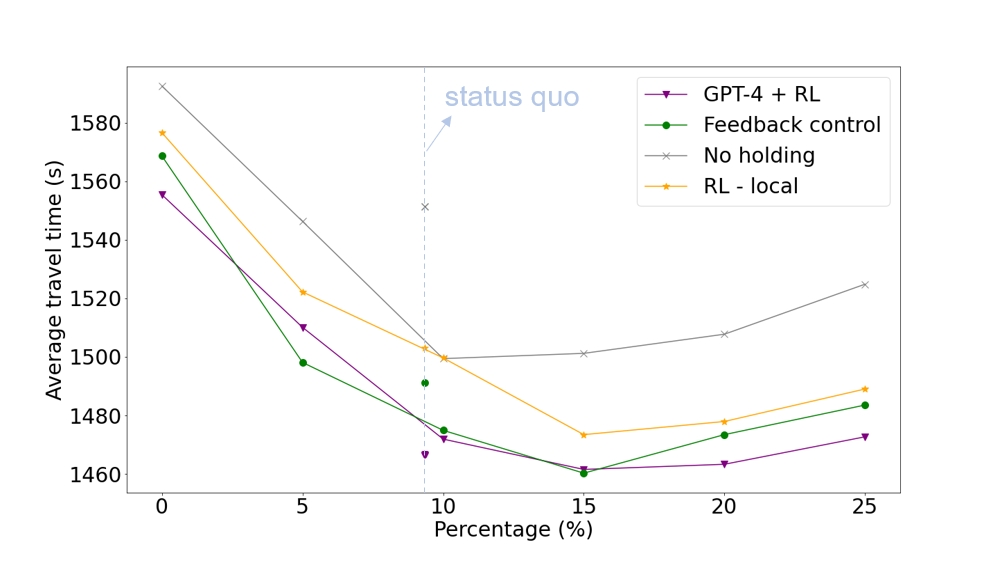}
    \caption{Bus line 1.}
    \label{fig:case2_sensitive_shared_pax_avg_travel_time}
\end{subfigure}\hspace{-16mm}
\hfill
\begin{subfigure}{0.52\textwidth}
    \includegraphics[width=\textwidth]{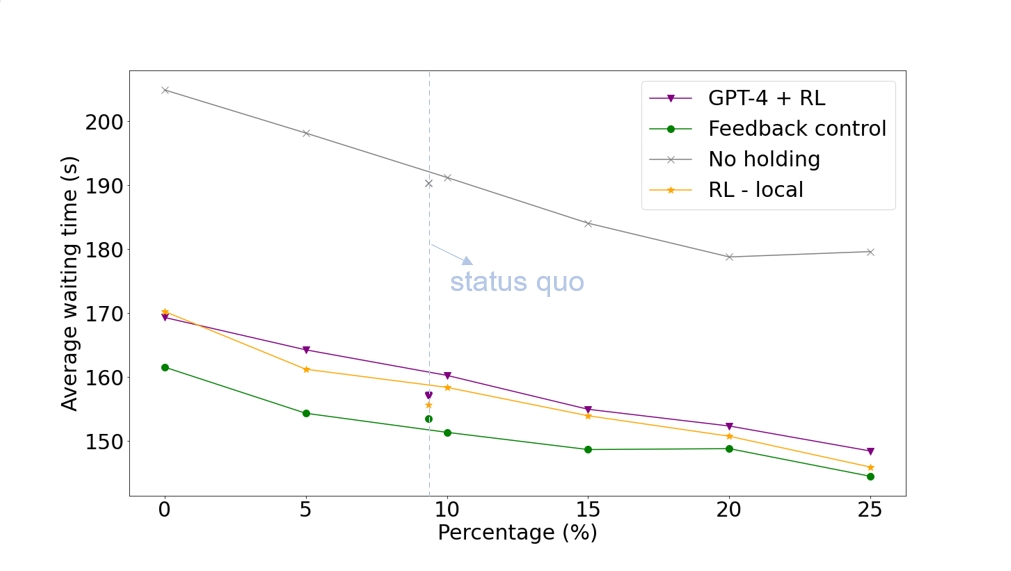}
    \caption{Bus line 52.}
    \label{fig:case2_sensitive_shared_pax_avg_waiting_time}
\end{subfigure}
\caption{Test results with different percentages of shared passengers.}
\label{fig:case2_sensitive_shared_pax}
\end{figure}

Fig. \ref{fig:case2_sensitive_shared_pax} displays the average travel time and average waiting time of several control strategies change with different percentages of shared passengers, ranging from 0\% to 25\% in increments of 5\%. The total number of passengers remains constant while the percentage of shared passengers varies. The shared passenger percentage of status quo is 9.35\%. As the proportion of shared passengers increases, the average waiting time decreases due to more passengers being able to board buses from either line. The average travel time initially decreases up to a shared passenger percentage of 15\% before rising with further increases in shared passenger percentage. In most cases, ``GPT-4 + RL'' consistently outperforms other control strategies in average travel time, except when the shared passenger percentage is 5\%. This indicates that the RL agents trained with the reward function provided by the proposed paradigm exhibit robustness across different passenger composition patterns.

\textbf{(2) Passenger demand level}

\begin{figure}
\centering
\begin{subfigure}{0.52\textwidth}
    \includegraphics[width=\textwidth]{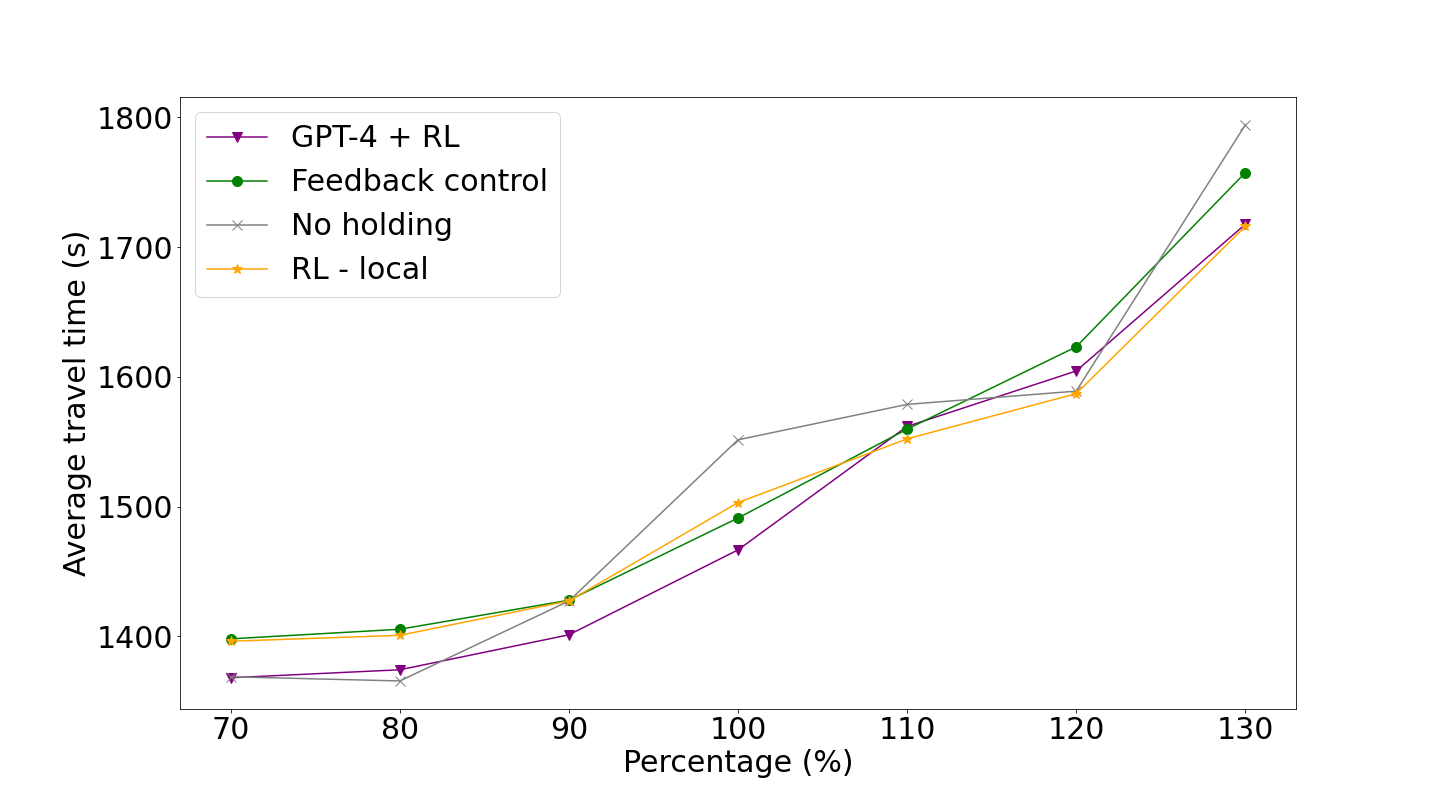}
    \caption{Bus line 1.}
    \label{fig:case2_sensitive_demand_level_avg_travel_time}
\end{subfigure}\hspace{-16mm}
\hfill
\begin{subfigure}{0.52\textwidth}
    \includegraphics[width=\textwidth]{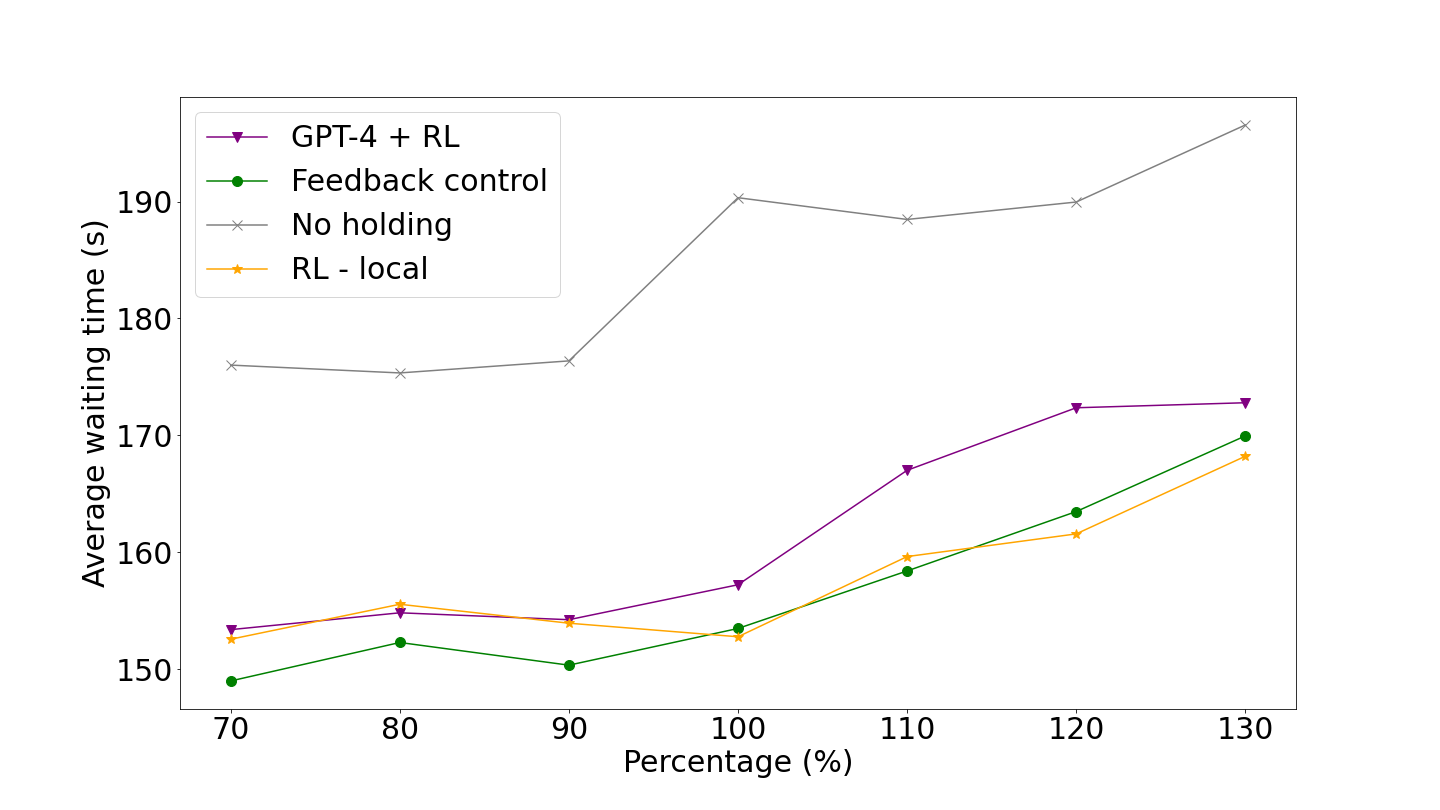}
    \caption{Bus line 52.}
    \label{fig:case2_sensitive_demand_level_avg_waiting_time}
\end{subfigure}
\caption{Test results with different passenger demand levels.}
\label{fig:case2_sensitive_demand_level}
\end{figure}

Fig. \ref{fig:case2_sensitive_demand_level} presents the average travel time and average waiting time for various control strategies under different passenger demand levels, ranging from 70\% to 130\% of the status quo, in increments of 10\%. As the total demand increases, both the average travel time and average waiting time rise accordingly. The RL agents trained with the final reward function from the proposed paradigm consistently outperform feedback control in terms of average travel time across all tested demand levels. Notably, within the demand range of 90\% to 110\%, which corresponds to the range used during the RL agents' training, ``GPT-4 + RL'' surpasses all other strategies. This demonstrates the stability and effectiveness of the proposed LLM-enhanced RL paradigm in managing varying levels of passenger demand. For optimal performance across a wider range of demand levels, the RL agents may need to be trained with a correspondingly broader range of demand scenarios.

\clearpage

\subsection{Case study 3: Multiple lines}

To further demonstrate the adaptiveness and generalization capability of the proposed paradigm, a larger real-world system comprising six bus lines is tested using the same reward function and trained agents from Case 2, as the reward function in Case 2 already accounts for bus stops serving both single bus lines and multiple bus lines.

\subsubsection{Scenario description}

\begin{figure}[h]
\centering
\includegraphics[width=0.8\linewidth]{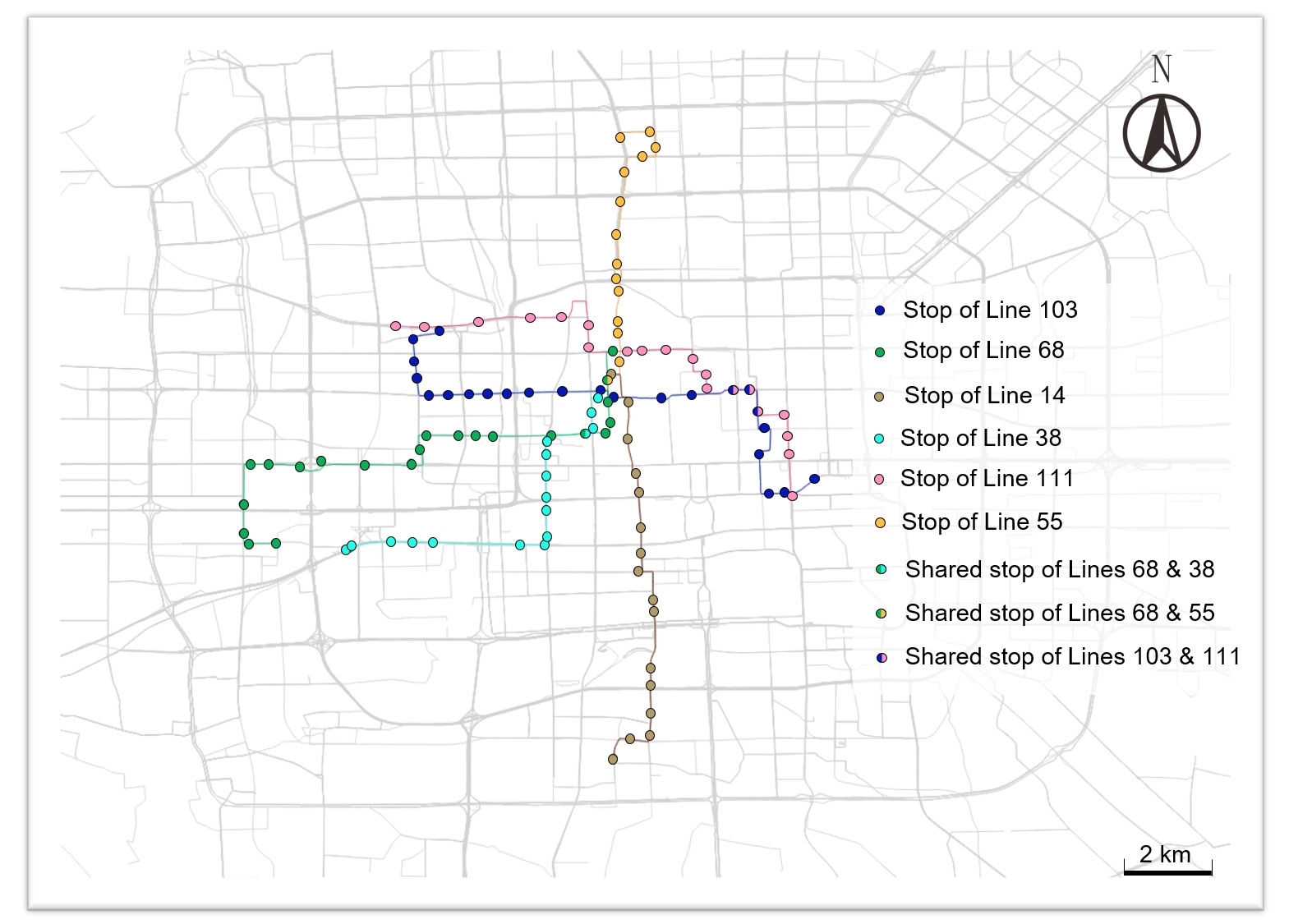}
\caption{\label{fig:busline_case3}Bus lines in Case 3}
\end{figure}

\begin{figure}[h]
\centering
\begin{subfigure}{0.5\textwidth}
    \includegraphics[width=\textwidth]{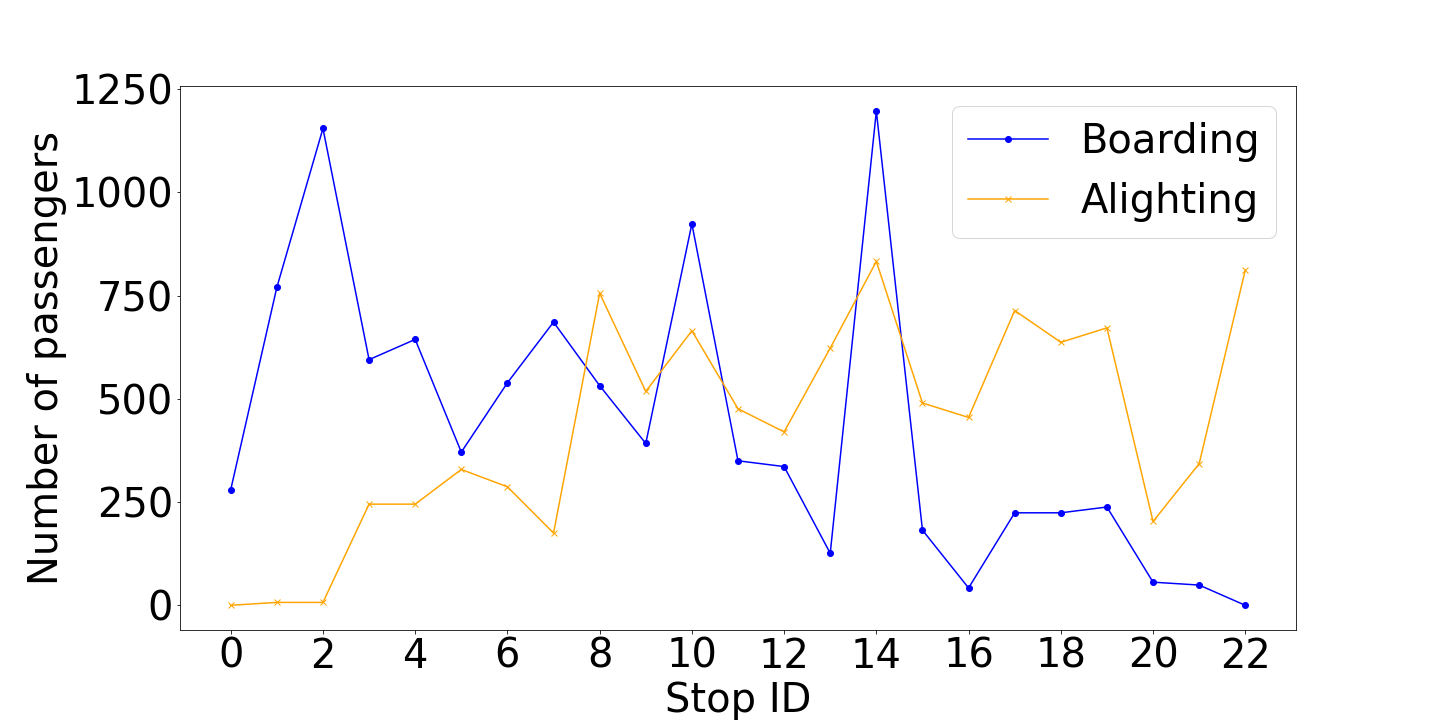}
    \caption{Bus line 103.}
    \label{fig:case3_passenger_demand_103}
\end{subfigure}\hspace{-16mm}
\hfill
\begin{subfigure}{0.5\textwidth}
    \includegraphics[width=\textwidth]{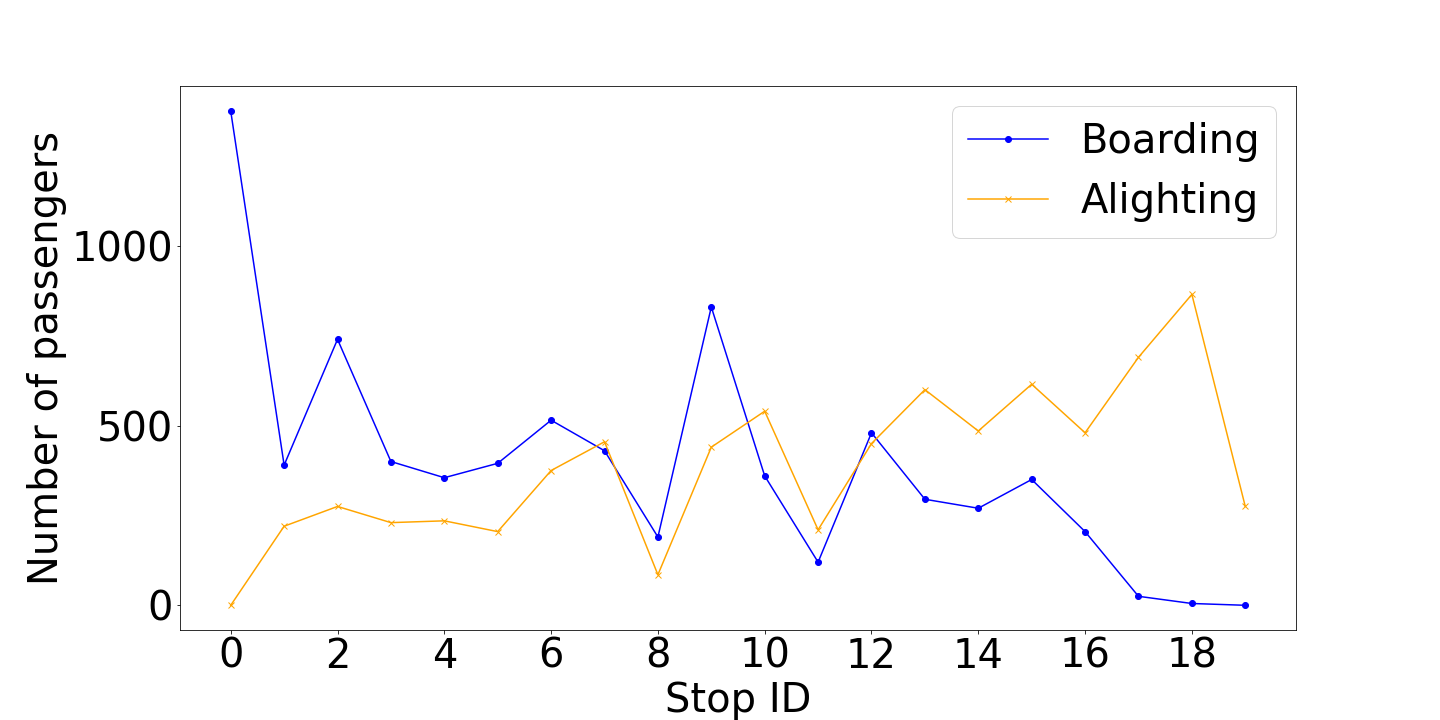}
    \caption{Bus line 111.}
    \label{fig:case3_passenger_demand_111}
\end{subfigure}
\hfill
\begin{subfigure}{0.5\textwidth}
    \includegraphics[width=\textwidth]{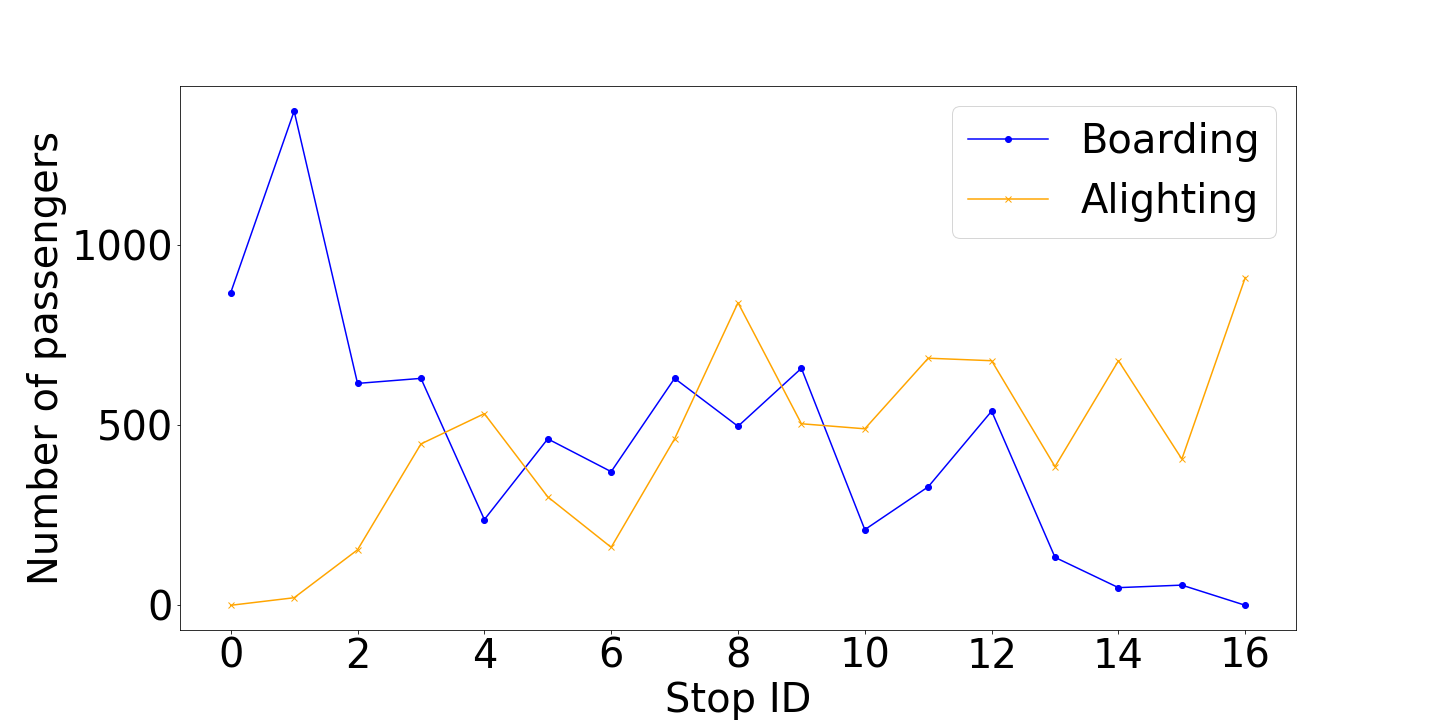}
    \caption{Bus line 14.}
    \label{fig:case3_passenger_demand_14}
\end{subfigure}\hspace{-16mm}
\hfill
\begin{subfigure}{0.5\textwidth}
    \includegraphics[width=\textwidth]{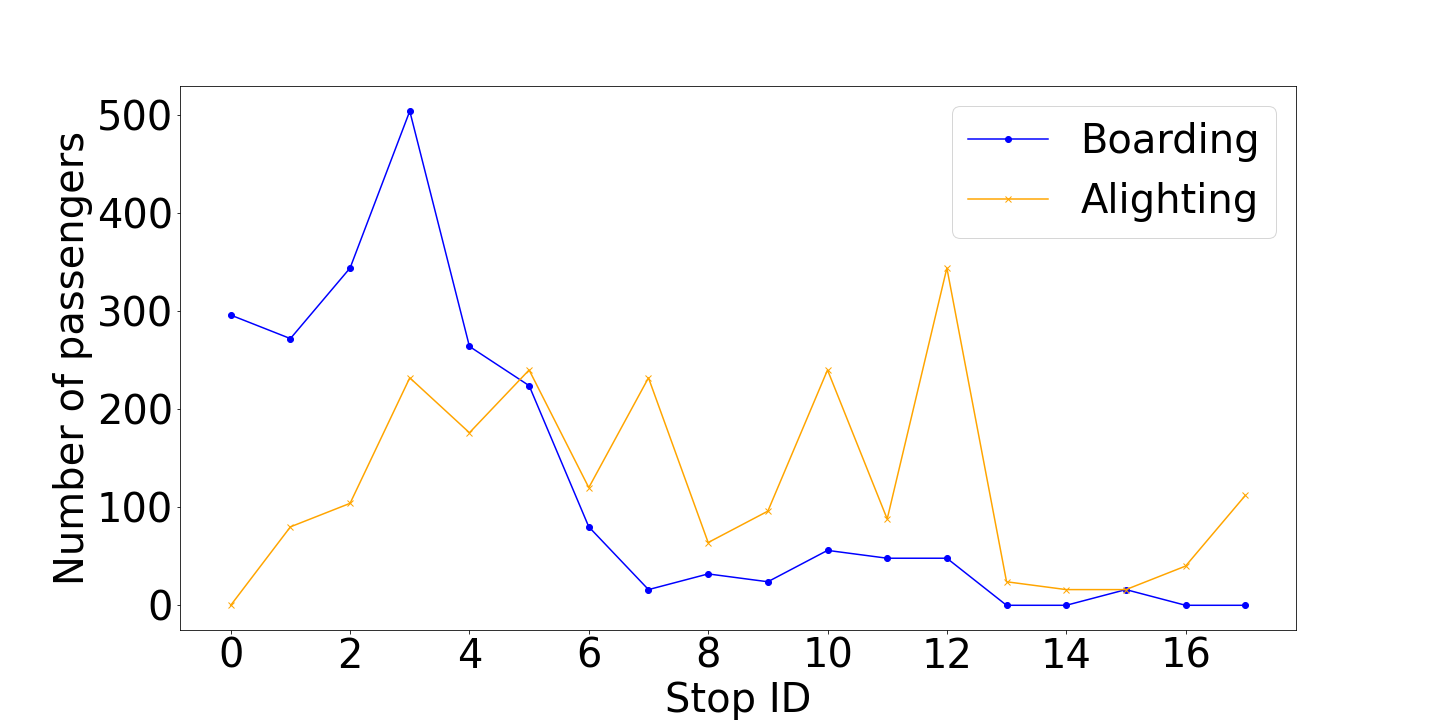}
    \caption{Bus line 38.}
    \label{fig:case3_passenger_demand_38}
\end{subfigure}
\hfill
\begin{subfigure}{0.5\textwidth}
    \includegraphics[width=\textwidth]{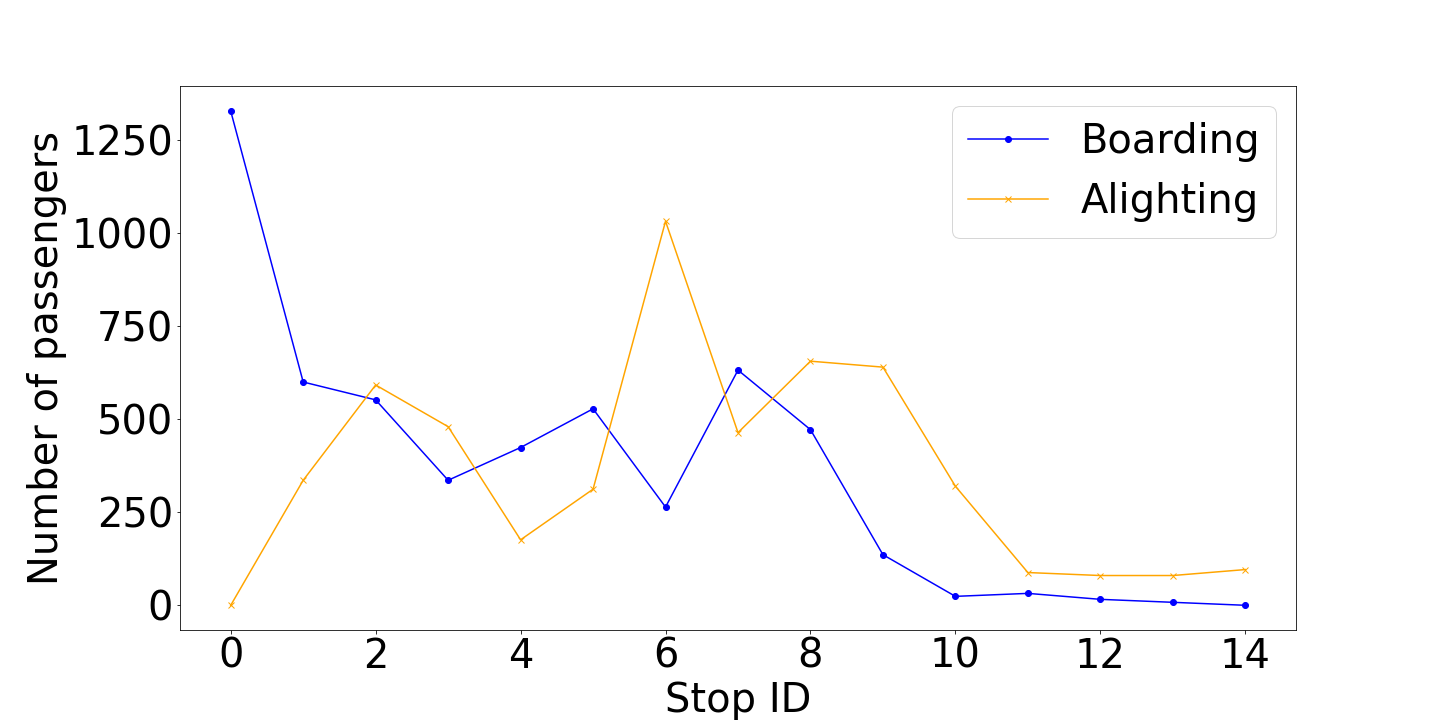}
    \caption{Bus line 55.}
    \label{fig:case3_passenger_demand_55}
\end{subfigure}\hspace{-16mm}
\hfill
\begin{subfigure}{0.5\textwidth}
    \includegraphics[width=\textwidth]{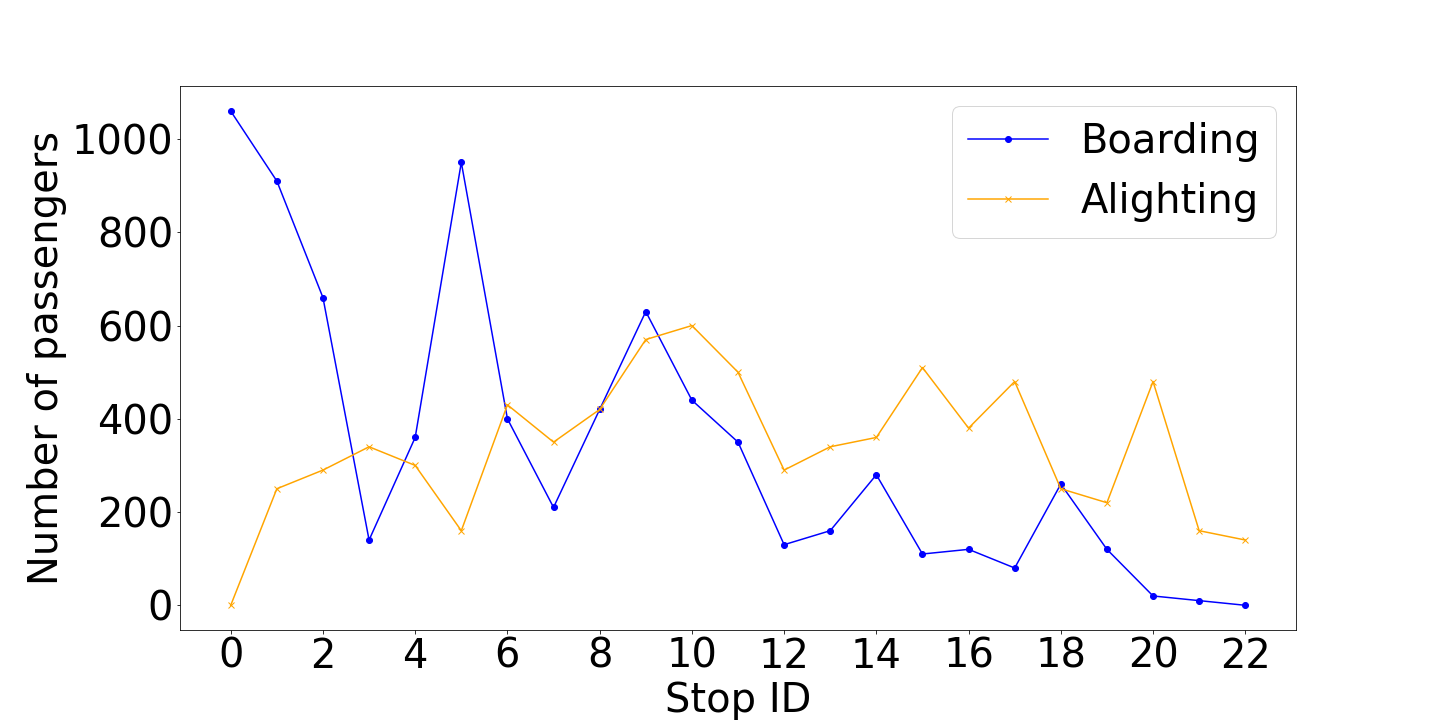}
    \caption{Bus line 68.}
    \label{fig:case3_passenger_demand_58}
\end{subfigure}
\caption{\textcolor{black}{Spatial distribution of passenger demand in Case 3.}}
\label{fig:case3_passenger_demand}
\end{figure}

Figure \ref{fig:busline_case3} illustrates the bus system in Case 3, where six bus lines interact within the same region, sharing some bus stops. The lengths of bus lines 103, 68, 14, 38, 111, and 55 are 13.01 km, 12.76 km, 11.08 km, 9.02 km, 12.66 km, and 8.21 km, respectively. The boarding and alighting passenger demand during the morning rush hours (6:00 to 10:00 AM) on a weekday (Tuesday, May 14th, 2019) for each bus line is depicted in Figure \ref{fig:case3_passenger_demand}. To simulate busy lines where bus holding is necessary, the real IC data-derived demand was scaled up accordingly. The departure interval for buses on each line is set to 5 minutes. The travel time between two adjacent stops for buses is randomly generated following a Gamma distribution, with the mean travel time based on real-world data \citep{berrebi2018partc}. Other simulation settings remain consistent with those in Case 2.

\subsubsection{Comparisons with baselines}

The proposed LLM-enhanced RL paradigm is compared against feedback control, RL-local, and no-holding scenarios. The test results are summarized in Table \ref{table:baseline_case3}. Each control method was evaluated using 30 different random seeds to mitigate the effects of stochasticity.

In terms of overall performance, the proposed paradigm achieves the lowest average travel time, reducing it by 13.41 seconds, 4.62 seconds, and 29.07 seconds compared to the no-holding, feedback control, and RL-local scenarios, respectively.

For the performance distributed across individual bus lines, the proposed paradigm also delivers the lowest average travel time in most lines, with the exception of Line 55. Notably, the feedback control method achieves the smallest SD of time headways and the lowest average waiting time for most bus lines. However, this does not correspond to the minimum average travel time.

This finding suggests that achieving perfectly balanced bus headways is not necessarily the optimal solution for minimizing the total travel time of passengers. Instead, a more considerate objective, like the reward function from the proposed paradigm, is better suited for addressing the complexities of real-world traffic systems.

\begin{sidewaystable}[h]
\centering
\caption{\label{table:baseline_case3}\textcolor{black}{Evaluations of different control methods in Case 3}}
\begin{threeparttable}
\begin{tblr}{
  width = \linewidth,
  colspec = {Q[94]Q[125]Q[102]Q[92]Q[83]Q[27]Q[125]Q[90]Q[90]Q[79]},
  cells = {c},
  cell{1}{1} = {r=2}{},
  cell{2}{2} = {c=4}{0.401\linewidth},
  cell{2}{7} = {c=4}{0.384\linewidth},
  cell{7}{2} = {c=4}{0.401\linewidth},
  cell{7}{7} = {c=4}{0.384\linewidth},
  cell{12}{2} = {c=4}{0.401\linewidth},
  cell{12}{7} = {c=4}{0.384\linewidth},
  cell{17}{2} = {c=4}{0.401\linewidth},
  hline{1,7,12,17,22} = {-}{},
  hline{2,8,13,18} = {2-5,7-10}{},
  hline{3} = {1-5,7-10}{},
}
\textbf{Control method} & \textbf{SD of time headways} & \textbf{Avg. TT} & \textbf{Avg. WT} & \textbf{Avg. HT} & ~ & \textbf{SD of time headways} & \textbf{Avg. TT} & \textbf{Avg. WT} & \textbf{Avg. HT}\\
 & \textbf{Line 14} &  &  &  & ~ & \textbf{Line 111} &  &  & \\
\textbf{LLM+RL} & 84.07$\pm$7.39 & \textbf{998.91}$\pm$62.38 & 171.13$\pm$14.80 & 6.75$\pm$0.62 & ~ & 84.59$\pm$6.13 & \textbf{851.94}$\pm$7.80 & 156.62$\pm$2.35 & 4.59$\pm$0.48\\
\textbf{Feedback} & \textbf{59.79}$\pm$5.62 & 1002.72$\pm$63.30 & \textbf{166.68}$\pm$12.49 & 8.11$\pm$0.67 & ~ & 64.32$\pm$3.61 & 853.97$\pm$7.39 & 153.32$\pm$1.25 & 5.65$\pm$0.44\\
\textbf{RL-local} & 67.82$\pm$6.10 & 1014.09$\pm$62.49 & 168.54$\pm$13.14 & 8.62$\pm$0.77 & \textbf{~} & \textbf{61.55}$\pm$4.32 & 870.37$\pm$9.65 & \textbf{153.11}$\pm$1.48 & 8.97$\pm$0.88\\
\textbf{No holding} & 166.39$\pm$11.49 & 1003.47$\pm$62.35 & 187.33$\pm$12.85 & 0$\pm$0 & ~ & 178.87$\pm$14.29 & 873.54$\pm$22.71 & 178.79$\pm$5.86 & 0$\pm$0\\
\textbf{~} & \textbf{Line 55} &  &  &  & ~ & \textbf{Line 38} &  &  & \\
\textbf{LLM+RL} & 80.21$\pm$7.71 & 535.05$\pm$6.93 & 154.41$\pm$2.29 & 5.62$\pm$0.64 & ~ & 56.80$\pm$4.60 & \textbf{590.57}$\pm$10.66 & 152.07$\pm$2.44 & 3.00$\pm$0.43\\
\textbf{Feedback} & \textbf{69.04}$\pm$5.51 & 538.38$\pm$6.53 & \textbf{154.04}$\pm$1.89 & 7.14$\pm$0.63 & \textbf{~} & \textbf{44.97}$\pm$3.27 & 599.17$\pm$10.57 & \textbf{151.61}$\pm$2.45 & 5.88$\pm$0.37\\
\textbf{RL-local} & 75.35$\pm$5.86 & 547.92$\pm$6.56 & 154.71$\pm$2.22 & 9.06$\pm$0.60 & ~ & 54.01$\pm$4.89 & 596.61$\pm$10.62 & 151.65$\pm$2.30 & 5.22$\pm$0.44\\
\textbf{No holding} & 145.60$\pm$15.58 & \textbf{529.13}$\pm$6.53 & 159.66$\pm$2.91 & 0$\pm$0 & ~ & 85.03$\pm$10.25 & 580.24$\pm$10.08 & 152.62$\pm$2.41 & 0$\pm$0\\
\textbf{~} & \textbf{Line 103} &  &  &  & ~ & \textbf{Line 68} &  &  & \\
\textbf{LLM+RL} & 131.03$\pm$8.38 & \textbf{1217.55}$\pm$69.46 & 196.46$\pm$8.71 & 7.40$\pm$0.68 & ~ & 79.06$\pm$7.46 & \textbf{777.87}$\pm$8.69 & 158.64$\pm$1.89 & 5.69$\pm$0.76\\
\textbf{Feedback} & \textbf{106.16}$\pm$9.45 & 1224.46$\pm$100.88 & \textbf{195.49}$\pm$14.54 & 9.94$\pm$0.57 & ~ & \textbf{59.82}$\pm$5.05 & 783.38$\pm$7.99 & \textbf{155.95}$\pm$1.52 & 7.68$\pm$0.59\\
\textbf{RL-local} & 132.02$\pm$11.67 & 1292.23$\pm$41.41 & 200.80$\pm$10.76 & 14.72$\pm$0.92 & ~ & 80.13$\pm$7.49 & 789.83$\pm$10.58 & 156.80$\pm$1.58 & 7.94$\pm$0.94\\
\textbf{No holding} & 252.14$\pm$18.22 & 1253.79$\pm$76.39 & 222.96$\pm$11.2 & 0$\pm$0 & ~ & 191.20$\pm$18.95 & 781.44$\pm$12.49 & 176.98$\pm$6.25 & 0$\pm$0\\
\textbf{~} & \textbf{Overall } &  &  &  & ~ & ~ & ~ & ~ & ~\\
\textbf{LLM+RL} & - & \textbf{901.31}$\pm$18.63 & 169.07$\pm$2.83 & - & ~ & ~ & ~ & ~ & ~\\
\textbf{Feedback} & - & 905.93$\pm$30.03 & \textbf{166.75}$\pm$4.12 & - & ~ & ~ & ~ & ~ & ~\\
\textbf{RL-local} & - & 930.38$\pm$15.32 & 168.62$\pm$4.07 & - & ~ & ~ & ~ & ~ & ~\\
\textbf{No holding} & - & 914.72$\pm$20.62 & 187.12$\pm$4.37 & - & ~ & ~ & ~ & ~ & ~
\end{tblr}
\begin{tablenotes}
      \small
      \item Note: TT, WT, and HT represent travel time, waiting time, and holding time, respectively. SD-TH represents SD of time headways. 
    \end{tablenotes}
\end{threeparttable}
\end{sidewaystable}

\subsubsection{Adaptiveness and generalization analysis}

Across all three test scenarios, the proposed LLM-enhanced paradigm consistently achieves the best performance in minimizing average travel time. Its consistently strong results demonstrate that the proposed paradigm is effective for both synthetic and real-world bus systems. It can adapt directly to large-scale bus systems with the experience of small systems without retraining and provides a unified, generic reward function for bus holding at stops serving both single and multiple lines.

The promising performance of the proposed paradigm can be attributed to two key features. First, the generic reward function is designed to operate seamlessly across different stop configurations, eliminating the need to distinguish between varying bus systems. This design grants the paradigm greater adaptability and stronger generalization compared to methods that require specific and separate reward functions for different types of bus stops. Second, the proposed paradigm's self-supervised (enabled by the \textit{reward refiner} module) and self-improving (enabled by the combination of the \textit{agent performance analyzer} and \textit{reward modifier} modules) workflow is inherently well-suited for zero-shot tasks. This enables the paradigm to generate effective reward functions and achieve promising RL agent performance with minimal iterations of self-adjustment and fine-tuning in previously unseen control scenarios.

Regarding conventional RL methods, RL-local consistently achieves the best performance among its three vanilla RL methods, highlighting the importance of dense reward functions in bus holding control problems. Feedback control, on the other hand, reliably achieves the smallest SD of time headways and the lowest average waiting time. In the synthetic, simple bus line system (Case 1), perfectly balanced headways result in similarly low average travel times. However, in more complex bus systems (Cases 2 and 3), focusing solely on headway balancing is insufficient to achieve significant reductions in total travel time, underscoring the necessity of a more considerate approach like the proposed paradigm.

\clearpage

\subsection{Generic tests}

To better illustrate the generalization capability of the proposed paradigm, we randomly generate 30 bus systems with diverse structures, varying in the number of bus lines, stops, and passenger demand, for performance evaluation. Specifically, the number of bus lines ranges from 2 to 10, with each line having 16 to 25 stops randomly assigned. Shared bus stop locations are also randomly selected. The number of passengers departing from each stop is randomly set between 100 and 1,100 over a four-hour period. The passengers' arrival follows Poisson processes, with destinations randomly selected from downstream stops. The key characteristics of each bus system and their corresponding test results are summarized in Table \ref{table:generic_tests}. The number of stops in each scenario is represented as a list, where each entry corresponds to the number of stops for a specific bus line. Number of passengers is the total number of passengers of all bus lines. 
The reward function utilized in the proposed approach is the same as that used in Case Studies 2 and 3, as the LLM-generated reward function is supposed to accommodate generic scenarios.

According to the test results across 30 different scenarios, the proposed paradigm outperforms feedback control in 90\% of the cases. On average, it achieves a 4.51\% reduction in average travel time compared to the no-holding scenario. In scenarios 3, 9, and 21, feedback control slightly outperforms the proposed paradigm, though the difference is minimal. Overall, the proposed paradigm reduces average travel time by 29.82 seconds compared to feedback control, demonstrating its effectiveness across most scenarios.

These extensive randomized tests confirm that the proposed paradigm consistently delivers promising performance across various bus system structures, highlighting its ability to provide generic and effective bus holding control strategies.

\begin{table}
\footnotesize
\centering
\caption{\label{table:generic_tests}Performance comparisons for generic tests}
\begin{tblr}{
  width = \linewidth,
  colspec = {Q[58]Q[81]Q[167]Q[113]Q[108]Q[62]Q[63]Q[62]Q[63]Q[60]Q[63]},
  row{2} = {c},
  row{3} = {c},
  row{34} = {c},
  column{2} = {c},
  column{4} = {c},
  column{5} = {c},
  cell{1}{1} = {r=3}{c},
  cell{1}{2} = {r=3}{},
  cell{1}{3} = {r=3}{},
  cell{1}{4} = {r=3}{},
  cell{1}{5} = {r=3}{},
  cell{1}{6} = {c=6}{0.373\linewidth,c},
  cell{2}{6} = {c=2}{0.125\linewidth},
  cell{2}{8} = {c=2}{0.125\linewidth},
  cell{2}{10} = {c=2}{0.123\linewidth},
  cell{4}{1} = {c},
  cell{4}{6} = {c},
  cell{4}{7} = {c},
  cell{4}{8} = {c},
  cell{4}{9} = {c},
  cell{4}{10} = {c},
  cell{4}{11} = {c},
  cell{5}{1} = {c},
  cell{5}{6} = {c},
  cell{5}{7} = {c},
  cell{5}{8} = {c},
  cell{5}{9} = {c},
  cell{5}{10} = {c},
  cell{5}{11} = {c},
  cell{6}{1} = {c},
  cell{6}{6} = {c},
  cell{6}{7} = {c},
  cell{6}{8} = {c},
  cell{6}{9} = {c},
  cell{6}{10} = {c},
  cell{6}{11} = {c},
  cell{7}{1} = {c},
  cell{7}{6} = {c},
  cell{7}{7} = {c},
  cell{7}{8} = {c},
  cell{7}{9} = {c},
  cell{7}{10} = {c},
  cell{7}{11} = {c},
  cell{8}{1} = {c},
  cell{8}{6} = {c},
  cell{8}{7} = {c},
  cell{8}{8} = {c},
  cell{8}{9} = {c},
  cell{8}{10} = {c},
  cell{8}{11} = {c},
  cell{9}{1} = {c},
  cell{9}{6} = {c},
  cell{9}{7} = {c},
  cell{9}{8} = {c},
  cell{9}{9} = {c},
  cell{9}{10} = {c},
  cell{9}{11} = {c},
  cell{10}{1} = {c},
  cell{10}{6} = {c},
  cell{10}{7} = {c},
  cell{10}{8} = {c},
  cell{10}{9} = {c},
  cell{10}{10} = {c},
  cell{10}{11} = {c},
  cell{11}{1} = {c},
  cell{11}{6} = {c},
  cell{11}{7} = {c},
  cell{11}{8} = {c},
  cell{11}{9} = {c},
  cell{11}{10} = {c},
  cell{11}{11} = {c},
  cell{12}{1} = {c},
  cell{12}{6} = {c},
  cell{12}{7} = {c},
  cell{12}{8} = {c},
  cell{12}{9} = {c},
  cell{12}{10} = {c},
  cell{12}{11} = {c},
  cell{13}{1} = {c},
  cell{13}{6} = {c},
  cell{13}{7} = {c},
  cell{13}{8} = {c},
  cell{13}{9} = {c},
  cell{13}{10} = {c},
  cell{13}{11} = {c},
  cell{14}{1} = {c},
  cell{14}{6} = {c},
  cell{14}{7} = {c},
  cell{14}{8} = {c},
  cell{14}{9} = {c},
  cell{14}{10} = {c},
  cell{14}{11} = {c},
  cell{15}{1} = {c},
  cell{15}{6} = {c},
  cell{15}{7} = {c},
  cell{15}{8} = {c},
  cell{15}{9} = {c},
  cell{15}{10} = {c},
  cell{15}{11} = {c},
  cell{16}{1} = {c},
  cell{16}{6} = {c},
  cell{16}{7} = {c},
  cell{16}{8} = {c},
  cell{16}{9} = {c},
  cell{16}{10} = {c},
  cell{16}{11} = {c},
  cell{17}{1} = {c},
  cell{17}{6} = {c},
  cell{17}{7} = {c},
  cell{17}{8} = {c},
  cell{17}{9} = {c},
  cell{17}{10} = {c},
  cell{17}{11} = {c},
  cell{18}{1} = {c},
  cell{18}{6} = {c},
  cell{18}{7} = {c},
  cell{18}{8} = {c},
  cell{18}{9} = {c},
  cell{18}{10} = {c},
  cell{18}{11} = {c},
  cell{19}{1} = {c},
  cell{19}{6} = {c},
  cell{19}{7} = {c},
  cell{19}{8} = {c},
  cell{19}{9} = {c},
  cell{19}{10} = {c},
  cell{19}{11} = {c},
  cell{20}{1} = {c},
  cell{20}{6} = {c},
  cell{20}{7} = {c},
  cell{20}{8} = {c},
  cell{20}{9} = {c},
  cell{20}{10} = {c},
  cell{20}{11} = {c},
  cell{21}{1} = {c},
  cell{21}{6} = {c},
  cell{21}{7} = {c},
  cell{21}{8} = {c},
  cell{21}{9} = {c},
  cell{21}{10} = {c},
  cell{21}{11} = {c},
  cell{22}{1} = {c},
  cell{22}{6} = {c},
  cell{22}{7} = {c},
  cell{22}{8} = {c},
  cell{22}{9} = {c},
  cell{22}{10} = {c},
  cell{22}{11} = {c},
  cell{23}{1} = {c},
  cell{23}{6} = {c},
  cell{23}{7} = {c},
  cell{23}{8} = {c},
  cell{23}{9} = {c},
  cell{23}{10} = {c},
  cell{23}{11} = {c},
  cell{24}{1} = {c},
  cell{24}{6} = {c},
  cell{24}{7} = {c},
  cell{24}{8} = {c},
  cell{24}{9} = {c},
  cell{24}{10} = {c},
  cell{24}{11} = {c},
  cell{25}{1} = {c},
  cell{25}{6} = {c},
  cell{25}{7} = {c},
  cell{25}{8} = {c},
  cell{25}{9} = {c},
  cell{25}{10} = {c},
  cell{25}{11} = {c},
  cell{26}{1} = {c},
  cell{26}{6} = {c},
  cell{26}{7} = {c},
  cell{26}{8} = {c},
  cell{26}{9} = {c},
  cell{26}{10} = {c},
  cell{26}{11} = {c},
  cell{27}{1} = {c},
  cell{27}{6} = {c},
  cell{27}{7} = {c},
  cell{27}{8} = {c},
  cell{27}{9} = {c},
  cell{27}{10} = {c},
  cell{27}{11} = {c},
  cell{28}{1} = {c},
  cell{28}{6} = {c},
  cell{28}{7} = {c},
  cell{28}{8} = {c},
  cell{28}{9} = {c},
  cell{28}{10} = {c},
  cell{28}{11} = {c},
  cell{29}{1} = {c},
  cell{29}{6} = {c},
  cell{29}{7} = {c},
  cell{29}{8} = {c},
  cell{29}{9} = {c},
  cell{29}{10} = {c},
  cell{29}{11} = {c},
  cell{30}{1} = {c},
  cell{30}{6} = {c},
  cell{30}{7} = {c},
  cell{30}{8} = {c},
  cell{30}{9} = {c},
  cell{30}{10} = {c},
  cell{30}{11} = {c},
  cell{31}{1} = {c},
  cell{31}{6} = {c},
  cell{31}{7} = {c},
  cell{31}{8} = {c},
  cell{31}{9} = {c},
  cell{31}{10} = {c},
  cell{31}{11} = {c},
  cell{32}{1} = {c},
  cell{32}{6} = {c},
  cell{32}{7} = {c},
  cell{32}{8} = {c},
  cell{32}{9} = {c},
  cell{32}{10} = {c},
  cell{32}{11} = {c},
  cell{33}{1} = {c},
  cell{33}{6} = {c},
  cell{33}{7} = {c},
  cell{33}{8} = {c},
  cell{33}{9} = {c},
  cell{33}{10} = {c},
  cell{33}{11} = {c},
  cell{34}{1} = {c=5}{0.527\linewidth},
  hline{1,4,34-35} = {-}{},
  hline{2-3} = {6-11}{},
}
Scenario & Num of lines & Num of stops & Num of shared stops & Num of passengers & Method &  &  &  &  & \\
 &  &  &  &  & LLM+RL &  & Feedback &  & No-holding & \\
 &  &  &  &  & Avg. TT & Avg. WT & Avg. TT & Avg. WT & Avg. TT & Avg. WT\\
1 & 2 & [19, 16] & 3 & 21822 & \textbf{2527.46} & 177.06 & 2600.63 & 179.24 & 2635.87 & 214.78\\
2 & 4 & [16, 23, 23, 23] & 2 & 50400 & \textbf{2823.71} & 216.50 & 2894.20 & 215.96 & 2949.93 & 254.00\\
3 & 4 & [22, 23, 16, 19] & 7 & 48667 & 2629.68 & 206.59 & \textbf{2625.30} & 198.91 & 2700.98 & 238.01\\
4 & 5 & [16, 17, 18, 18, 16]~ & 10 & 50314 & \textbf{2464.43} & 168.73 & 2480.06 & 166.93 & 2564.90 & 209.89\\
5 & 2 & [16, 20] & 0 & 22122 & \textbf{2529.44} & 190.04 & 2560.98 & 193.06 & 2573.67 & 212.55\\
6 & 4 & [19, 19, 22, 16] & 7 & 46818 & \textbf{2480.16} & 191.72 & 2491.53 & 190.10 & 2628.29 & 237.75\\
7 & 8 & [21, 24, 24, 23, 19, 19, 20, 19]~ & 5 & 102148 & \textbf{2704.05} & 208.64 & 2745.59 & 204.79 & 2884.23 & 252.29\\
8 & 8 & [16, 23, 17, 23, 19, 16, 21, 18]~ & 9 & 89983 & \textbf{2573.80} & 195.55 & 2576.58 & 189.90 & 2685.28 & 232.67\\
9 & 7 & [21, 22, 16, 18, 22, 17, 19] & 21 & 82182 & 2563.42 & 193.75 & \textbf{2563.09} & 185.83 & 2703.33 & 233.12\\
10 & 2 & [16, 21] & 2 & 20576 & \textbf{2421.83} & 171.23 & 2439.55 & 168.58 & 2514.52 & 207.05\\
11 & 3 & [17, 18, 24] & 3 & 36133 & \textbf{2676.83} & 204.40 & 2693.56 & 201.88 & 2785.29 & 235.27\\
12 & 8 & [21, 17, 18, 22, 17, 19, 24, 17] & 22 & 96042 & \textbf{2640.68} & 197.72 & 2667.27 & 192.51 & 2809.18 & 237.71\\
13 & 2 & [19, 18] & 1 & 23159 & \textbf{2204.39} & 182.05 & 2230.00 & 180.26 & 2406.84 & 227.24\\
14 & 4 & [21, 22, 17, 20] & 14 & 48568 & \textbf{2643.03} & 207.16 & 2658.21 & 196.90 & 2764.17 & 239.62\\
15 & 4 & [19, 20, 16, 20] & 8 & 45894 & \textbf{2593.98} & 188.31 & 2625.05 & 183.06 & 2721.93 & 221.44\\
16 & 9 & [19, 20, 22, 24, 17, 19, 24, 18, 24] & 25 & 111020 & \textbf{2649.46} & 211.61 & 2659.50 & 205.48 & 2730.92 & 240.71\\
17 & 8 & [16, 16, 17, 22, 21, 22, 19, 16]~ & 23 & 87750 & \textbf{2602.54} & 195.85 & 2630.93 & 194.21 & 2716.48 & 231.83\\
18 & 7 & [21, 16, 20, 22, 16, 21, 17]~ & 28 & 79876 & \textbf{2579.83} & 192.92 & 2642.72 & 189.66 & 2694.07 & 231.59\\
19 & 8 & [16, 24, 22, 24, 21, 18, 23, 16] & 32 & 100262 & \textbf{2676.63} & 210.32 & 2698.42 & 205.95 & 2811.40 & 252.86\\
20 & 7 & [21, 24, 23, 19, 18, 19, 16] & 30 & 86460 & \textbf{2582.91} & 198.91 & 2681.58 & 201.58 & 2735.83 & 237.00\\
21 & 9 & [18, 17, 24, 22, 19, 21, 23, 20, 22]~ & 27 & 109794 & 2690.02 & 205.15 & \textbf{2685.84} & 196.24 & 2787.87 & 238.62\\
22 & 7 & [16, 21, 21, 21, 21, 23, 21]~ & 11 & 85168 & \textbf{2681.25} & 206.59 & 2713.65 & 205.46 & 2801.19 & 245.39\\
23 & 2 & [18, 17]~ & 3 & 20434 & \textbf{2372.52} & 167.78 & 2376.47 & 163.35 & 2502.24 & 209.46\\
24 & 9 & [22, 18, 23, 18, 21, 16, 17, 21, 20] & 33 & 107698 & \textbf{2638.42} & 199.29 & 2660.33 & 190.24 & 2777.09 & 237.35\\
25 & 7 & [18, 19, 17, 24, 22, 19, 19] & 27 & 82922 & \textbf{2661.10} & 196.40 & 2664.45 & 194.48 & 2791.55 & 238.83\\
26 & 7 & [22, 21, 16, 20, 22, 16, 21] & 3 & 83146 & \textbf{2704.50} & 203.16 & 2758.23 & 199.76 & 2810.78 & 243.45\\
27 & 8 & [19, 24, 18, 24, 23, 21, 18, 22]~ & 29 & 101528 & \textbf{2638.89} & 212.13 & 2643.67 & 203.83 & 2738.29 & 242.03\\
28 & 6 & [21, 20, 24, 22, 16, 16] & 9 & 69881 & \textbf{2592.54} & 199.97 & 2678.67 & 205.81 & 2715.22 & 236.72\\
29 & 2 & [19, 16]~ & 3 & 21822 & \textbf{2527.46} & 177.06 & 2600.63 & 179.24 & 2635.87 & 214.78\\
30 & 5 & [20, 17, 24, 21, 24] & 9 & 61377 & \textbf{2592.13} & 204.20 & 2615.04 & 199.50 & 2761.64 & 246.03\\
Average &  &  &  &  & \textbf{2588.90} & 196.03 & 2618.72 & 192.76 & 2711.30 & 233.33
\end{tblr}
\end{table}

\clearpage
\section{Conclusions}
\label{section:conclusions}

This study develops a novel LLM-enhanced RL paradigm that leverages generative AI to formulate and refine reward functions for RL-based control strategies. The paradigm incorporates several LLM-based modules to automate the initialization and improvement of reward functions. This automated workflow, combined with rigorous performance checks, ensures that the reward function is continuously optimized through feedback from the RL environment. 

The effectiveness and adaptiveness of the proposed paradigm are evaluated through extensive numerical tests in both synthetic and real-world bus holding control scenarios, varying in the number of bus lines, stops, and passenger demand. The results show that the proposed LLM-enhanced RL paradigm significantly outperforms vanilla RL controllers, the LLM-based controller, feedback controllers, \textcolor{black}{and optimization-based controllers} regarding average travel time in both single-line and multi-line systems. The paradigm demonstrates effective performance improvement and convergence within a manageable number of iterations (e.g., 10 iterations). Among the LLMs tested, those from the GPT family exhibit superior performance compared to Claude-Opus and Gemini-1.0-pro in the context of this study. Ablation studies confirm the necessity of all LLM-based modules within the paradigm to achieve promising control performance and to circumvent the need for expert input. Sensitive analysis reveals that the proposed paradigm provides robust performance across various passenger compositions and demand levels. Overall, the LLM-enhanced RL paradigm is validated as effective and adaptable for generic bus holding control strategies.

The proposed LLM-enhanced RL paradigm can be further applied to other smart mobility applications (e.g., traffic signal control and variable speed limit). This paper mainly focuses on the design and planning of LLMs, it is also interesting to fine-tune LLMs for better reward design capabilities. Additionally, the interpretability of the LLM-generated rewards might be better explained by the LLM itself with a better prompt design. In general, this study sheds light on the tool use of LLMs for smart mobility applications, which could potentially become a new research venue for transportation research.

\section*{Declaration of Competing Interest}

The authors declare that they have no known competing financial interests or personal relationships that could have appeared to influence the work reported in this paper.

\section*{CRediT authorship contribution statement}

Jiajie Yu: Methodology, Conceptualization, Visualization, Writing - original draft, and Writing - review \& editing. Yuhong Wang: Writing - review \& editing. Wei Ma: Conceptualization, Supervision, Writing - review \& editing, Funding acquisition.

\section*{Acknowledgments}
The work described in this paper is supported by the Innovation and Technology Fund - Mainland-Hong Kong Joint Funding Scheme (ITF-MHKJFS) (Project No. MHP/150/22), grants from the Research Grants Council of the Hong Kong Special Administrative Region, China (Project No. PolyU/15206322 and PolyU/15227424), and a grant from the Otto Poon Charitable Foundation Smart Cities Research Institute, The Hong Kong Polytechnic University (CD06). The contents of this article reflect the views of the authors, who are responsible for the facts and accuracy of the information presented herein.

\bibliography{sample}

\clearpage

\appendix

\section{Prompt examples}
\subsection{Prompts of reward initializer}
\label{appendix:reward_initializer}

\begin{tcolorbox}[breakable=true, boxrule={0.5pt}, sharp corners={all}]
\setlength{\parskip}{1ex}
\small

You are now a proficient reward designer for a reinforcement learning (RL) agent. You need to write proper reward functions for the agent. The agent will be trained for a bus holding control problem to improve the performance of the bus system. The detailed description of the task is as below.

\textbf{\textcolor{violet}{\#\# Background}}

In bus transit systems, due to the heterogeneous bus dwell time and traffic disturbance, the buses easily get bunch where two or more buses get too close even though the departure times of the buses are well-designed, particularly affecting bus services. Buses can be held at bus stops to adjust the headways to avoid bus bunching and balance the bus headways. However, unreasonable and excessive holding control for buses will increase the total travel time for passengers.

\textbf{\textcolor{violet}{\#\# Task description}}

- Objective: There are two bus lines in the bus system. The two bus lines share part of the stops. The objective is to balance the time headways between buses from the same bus line as well as from different bus lines and minimize the total/average travel time of bus passengers. Since there is a strong correlation and consistent performance between time headway and space headway, balancing the time headways can be regarded as balancing the space headways. Avoiding excessive holding control is necessary as it leads to delays in travel time.

\textbf{\textcolor{violet}{\#\# Basic definition:}}

- Time headway: the interval of arriving time between successive buses at the same bus stop. \\
- Space headway: the space between successive buses. \\
- Travel time of passenger: the time from the passenger arriving at the bus stop to his/her alighting time at the destination bus stop. It is the waiting time at the bus stop plus the time onboard the bus.

\textbf{\textcolor{violet}{\#\# Agent definition}}

- Overview of the agent:  The agent will be activated and decide an action every time a dwell bus is ready to continue the trip after the boarding process of passengers or at the end of the last holding action. The action will last for 5 seconds, or say, the action step is 5 seconds.\\
- Agent's action: It is a binary variable. The value of '0' represents holding the bus at the stops for the next action step; the value of '1' represents not holding the bus, and the bus can continue its trip.\\
- Agent's state: It is a list [s0, s1, s2, s3, s4, s5]. Each element in this list is introduced below:\\
\hbox{\ \ \ \ }- s0: It is the forward space headway (meters) between the current bus and the forward bus in the same bus line.\\
\hbox{\ \ \ \ }- s1: It is the backward space headway (meters) between the current bus and the backward bus in the same bus line.\\
\hbox{\ \ \ \ }- s2: It is the forward space headway (meters) between the current bus and the forward bus in the \textbf{**other**} bus line. \textbf{**The value is 0 when the dwell stop is not a shared stop.**}\\
\hbox{\ \ \ \ }- s3: It is the backward space headway (meters) between the current bus and the backward bus in the \textbf{**other**} bus line. \textbf{**The value is 0 when the dwell stop is not a shared stop.**}\\
\hbox{\ \ \ \ }- s4: It is the number of onboard passengers of the current bus. Passengers can not board or alight when the bus is held, so the number of onboarding passengers is unchanged during holding control.\\
\hbox{\ \ \ \ }- s5: It is the holding time that has been spent during this dwell.

\textbf{\textcolor{violet}{\#\# Input parameters of the reward function}}

- Agent's action, denoted as \textcolor{purple}{\textasciigrave action\textasciigrave} \ in the code.\\
- Agent's state at the beginning of the action, denoted as \textcolor{purple}{\textasciigrave current\_state\textasciigrave} \ in the code.\\
- Agent's state at the end of the action, denoted as \textcolor{purple}{\textasciigrave next\_state\textasciigrave} \ in the code. Please be aware that the s4 in \textcolor{purple}{\textasciigrave next\_state\textasciigrave} \ is the same as the s4 in \textcolor{purple}{\textasciigrave current\_state\textasciigrave} \ since passengers boarding and alighting are not allowed during holding control.

\textbf{\textcolor{violet}{\#\# General simulation environment information}}

- Simulation environment and settings: There are two bus lines in the bus system. Each of the bus lines has 27 bus stops, and 8 of these stops are shared stops that serve both lines. The maximum holding time T = 90 s. \\
- Arriving time of passengers: The passengers' arrival time at each bus stop follows the Poisson process. Please be aware the arrival time of passengers is exogenous inputs which can not be changed by the agent control.\\
- Passengers are not allowed to board and alight during holding control.

\textbf{\textcolor{violet}{\#\# Reward function requirements}}

- You should write a reward function to achieve the \textbf{**Task description**}. The information you can use to formulate the reward function has been listed in the \textbf{**Input parameters of the reward function**}. \\
- Stops serve a single bus line and two bus lines share the same reward function. Therefore, the formulation of the reward function should be generic.

\textbf{\textcolor{violet}{\#\# Output Requirements}}

- The reward function should be written in Python 3.8.16.\\
- Output the code block only. \textbf{**Do not output anything else outside the code block**}.\\
- You should include \textbf{**sufficient comments**} in your reward function to explain your thoughts, the objective, and \textbf{**implementation details**}. The implementation can be specified to a specific line of code.\\
- Ensure the reward value will not be extremely large or extremely small which makes the reward meaningless.\\
- If you need to import packages (e.g. math, numpy) or define helper functions, define them at the beginning of the function. Do not use unimported packages and undefined functions.\\
- Ensure there is no variable in the reward function that needs to be tuned. The reward function should be able to be used directly.

\textbf{\textcolor{violet}{\#\# Output format}}

Strictly follow the following format. \textbf{**Do not output anything else outside the code block**}.\\
\\
\hbox{\ \ \ \ }def reward\_function(current\_state, action, next\_state):
    \\
    \hbox{\ \ \ \ \ \ \ \ }\# Thoughts:\\
    \hbox{\ \ \ \ \ \ \ \ }\# ...\\
    \hbox{\ \ \ \ \ \ \ \ }\# (initial the reward)\\
    \hbox{\ \ \ \ \ \ \ \ }reward = 0\\
    \hbox{\ \ \ \ \ \ \ \ }\# (import packages and define helper functions)\\
    \hbox{\ \ \ \ \ \ \ \ }import numpy as np\\
    \hbox{\ \ \ \ \ \ \ \ }...\\
    \hbox{\ \ \ \ \ \ \ \ }return reward

Now write a reward function. Then in each iteration, I will use the reward function to train the RL agent and test it in the environment. I will give you possible reasons for the failure found during the testing, and you should modify the reward function accordingly. \textbf{**Do not output anything else outside the code block. Please double check the output code. Ensure there is no error. The variables or function used should be defined already.**}
  
\end{tcolorbox}

\subsection{Prompts of reward modifier}
\label{appendix:reward_modifier}

\begin{tcolorbox}[breakable=true, boxrule={0.5pt}, sharp corners={all}]
\setlength{\parskip}{1ex}
\small
You are now a proficient reward designer for a reinforcement learning (RL) agent. The agent is designed for a bus holding control problem to improve the performance of the bus system. The detailed description of the task is in the following section. I now have a reward function. The reward function has been used to train the RL agent several times and is tested in the environment. I will provide you with the current reward function, an analysis on the performance of the current reward function, and suggestions for reward improvement. You should help me modify and improve the current reward function.

\textbf{\textcolor{violet}{\#\# Background}}

In bus transit systems, due to the heterogeneous bus dwell time and traffic disturbance, the buses easily get bunch where two or more buses get too close even though the departure times of the buses are well-designed, particularly affecting bus services. Buses can be held at bus stops to adjust the headways to avoid bus bunching and balance the bus headways. However, unreasonable and excessive holding control for buses will increase the total travel time for passengers.

\textbf{\textcolor{violet}{\#\# Task description}}

- Objective: There are two bus lines in the bus system. The two bus lines share part of the stops. The objective is to balance the time headways between buses from the same bus line as well as from different bus lines and minimize the total/average travel time of bus passengers. Since there is a strong correlation and consistent performance between time headway and space headway, balancing the time headways can be regarded as balancing the space headways. Avoiding excessive holding control is necessary as it leads to delays in travel time.

\textbf{\textcolor{violet}{\#\# Basic definition}}

- Time headway: the interval of arriving time between successive buses at the same bus stop. \\
- Space headway: the space between successive buses. \\
- Travel time of passenger: the time from the passenger arriving at the bus stop to his/her alighting time at the destination bus stop. It is the waiting time of passengers at the bus stop plus the time onboard the bus.

\textbf{\textcolor{violet}{\#\# Agent definition}}

- Overview of the agent: The agent will be activated and decide an action every time a dwell bus is ready to continue the trip after the boarding process of passengers or at the end of last holding action. The action will last for 5 seconds, or say, the action step is 5 seconds.\\
- Agent's action: It is a binary variable. The value of '0' represents holding the bus at the stops for the next action step; the value of '1' represents not holding the bus, and the bus can continue its trip.\\
- Agent's state: It is a list [s0, s1, s2, s3, s4, s5]. Each element in this list is introduced below:\\
\hbox{\ \ \ \ }- s0: It is the forward space headway (meters) between the current bus and the forward bus in the same bus line.\\
\hbox{\ \ \ \ }- s1: It is the backward space headway (meters) between the current bus and the backward bus in the same bus line.\\
\hbox{\ \ \ \ }- s2: It is the forward space headway (meters) between the current bus and the forward bus in the \textbf{**other**} bus line. \textbf{**The value is 0 when the dwell stop is not a shared stop.**}\\
\hbox{\ \ \ \ }- s3: It is the backward space headway (meters) between the current bus and the backward bus in the \textbf{**other**} bus line. \textbf{**The value is 0 when the dwell stop is not a shared stop.**}\\
\hbox{\ \ \ \ }- s4: It is the number of onboard passengers of the current bus. Passengers can not board or alight when the bus is held, so the number of onboarding passengers is unchanged during holding control.\\
\hbox{\ \ \ \ }- s5: It is the holding time that has been spent during this dwell.
    
\textbf{\textcolor{violet}{\#\# Input parameters of the reward function}}

- Agent's action, denoted as \textcolor{purple}{\textasciigrave action\textasciigrave} \ in the code.\\
- Agent's state at the beginning of the action, denoted as \textcolor{purple}{\textasciigrave current\_state\textasciigrave} \ in the code.\\
- Agent's state at the end of the action, denoted as \textcolor{purple}{\textasciigrave next\_state\textasciigrave} \ in the code. Please be aware that the s4 in \textcolor{purple}{\textasciigrave next\_state\textasciigrave} \ are the same as the s4 in \textcolor{purple}{\textasciigrave current\_state\textasciigrave} \ since passengers boarding and alighting is not allowed during holding control.

\textbf{\textcolor{violet}{\#\# General simulation environment information}}

- Simulation environment and settings: There are two bus lines in the bus system. Each of the bus lines has 27 bus stops, and 8 of these stops are shared stops that serve both lines. The maximum holding time T = 90 s.\\
- Arriving time of passengers: The passengers' arrival time at each bus stop follows the Poisson process. Please be aware the arrival time of passengers is exogenous inputs which can not be changed by the agent control.\\
- Passengers are not allowed to board and alight during holding control.

\textbf{\textcolor{violet}{\#\# Reward function requirements}}

- You should write a reward function to achieve the \textbf{**Task description**}. The information you can use to formulate the reward function has been listed in the \textbf{**Input parameters of the reward function**}. \\
- Stops serve a single bus line and two bus lines share the same reward function. Therefore, the formulation of the reward function should be generic.

\textbf{\textcolor{violet}{\#\# Current reward function}}

\texttt{\{current\_reward\_function\}}
    
\textbf{\textcolor{violet}{\#\# Analysis and suggestions for current reward function}} 

The reward function is used to train the reinforcement learning agent several times. Here is some analysis of the agent's performance and suggestions for the current reward function improvement:

\textbf{\textcolor{violet}{\#\# Analysis of Training and Test Results}}

\texttt{\{analysis\}}

\textbf{\textcolor{violet}{\#\# Output Requirements}}

- Please consider the analysis and suggestions above. Modify and improve the current reward function. \\
\hbox{\ \ \ \ }1. You can both modify the current lines and add new lines. You can use any variable in the \textbf{**Input parameters of the reward function**} to define the reward function. \\
\hbox{\ \ \ \ }2. If necessary, you can write a \textbf{**totally different**} reward function than the current one.\\
\hbox{\ \ \ \ }3. Consider modifying the reward and penalty values in the current reward function to balance them.\\
\hbox{\ \ \ \ }4. In the first part of the reward function, you should provide your thoughts on modifying the reward function. \textbf{**The thoughts should be concise.**}\\
\hbox{\ \ \ \ }5. Ensure the reward value will not be extremely large or extremely small which makes the reward meaningless.\\
\hbox{\ \ \ \ }6. Ensure there is no variable in the reward function that needs to be tuned. The reward function should be able to be used directly.
- The reward function should be written in Python 3.8.16.\\
- Output the code block only. \textbf{**Do not output anything else outside the code block**}.\\
- You should include \textbf{**sufficient comments**} in your reward function to explain your thoughts, the objective and \textbf{**implementation details**}. The implementation can be specified to a specific line of code.\\
- If you need to import packages (e.g. math, numpy) or define helper functions, define them at the beginning of the function. Do not use unimported packages and undefined functions.\\
- \textbf{**Please double check the output code. Ensure there is no error. The variables or function used should be defined already**}

\textbf{\textcolor{violet}{\#\# Output format}}

Strictly follow the following format. \textbf{**Do not output anything else outside the code block**}. \\
\\
\hbox{\ \ \ \ }def reward\_function(current\_state, action, next\_state):\\
    \hbox{\ \ \ \ \ \ \ \ }\#Thoughts: \\
    \hbox{\ \ \ \ \ \ \ \ }\#...\\
     \\   
    \hbox{\ \ \ \ \ \ \ \ }\# (initial the reward)\\
    \hbox{\ \ \ \ \ \ \ \ }reward = 0\\
    \hbox{\ \ \ \ \ \ \ \ }\# (import packages and define helper functions)\\
    \hbox{\ \ \ \ \ \ \ \ }import numpy as np\\
    \hbox{\ \ \ \ \ \ \ \ }...\\
    \hbox{\ \ \ \ \ \ \ \ }return reward\\

Now write a new reward function to improve the current one based on \textbf{**Analysis and suggestions for current reward function**}. I will use the new reward function to train the RL agent and test it in the environment. \textbf{**Do not output anything else outside the code block**}. \textbf{**Please double check the output code. Ensure there is no error. The variables or functions used should be defined already.**}  
\end{tcolorbox} 

\subsection{Prompts of analyer}
\label{appendix:analyzer}

\begin{tcolorbox}[breakable=true, boxrule={0.5pt}, sharp corners={all}]
\setlength{\parskip}{1ex}
\small

You are now a proficient reward designer for a reinforcement learning (RL) agent. The agent is designed for a bus holding control problem to improve the performance of the bus system. The detailed description of the task is in the following section. I now have a reward function to train the agent to complete the described task. I have trained an RL agent and tested for several times and tested in the simulation environment. I will give you the information on the training and test results, i.e. trajectory of training rewards, trajectories of actions, rewards, and states during the test. You should help me write a proper analysis of possible reasons for inefficiency from both the training and test performances and your suggestions on the reward function improvement. 

\textbf{\textcolor{violet}{\#\# Background}}

In bus transit systems, due to the heterogeneous bus dwell time and traffic disturbance, the buses easily get bunch where two or more buses get too close even though the departure times of the buses are well-designed, particularly affecting bus services. Buses can be held at bus stops to adjust the headways to avoid bus bunching and balance the bus headways. However, unreasonable and excessive holding control for bus will increase the total travel time for passengers.

\textbf{\textcolor{violet}{\#\# Task description}}

- Objective: There are two bus lines in the bus system. The two bus lines share part of the stops. The objective is to balance the time headways between buses from the same bus line as well as from different bus lines and minimize the total/average travel time of bus passengers. Since there is a strong correlation and consistent performance between time headway and space headway, balancing the time headways can be regarded as balancing the space headways. Avoiding excessive holding control is necessary as it leads to delays in travel time.

\textbf{\textcolor{violet}{\#\# Basic definition}}

- Time headway: the interval of arriving time between successive buses at the same bus stop. \\
- Space headway: the space between successive buses. \\
- Travel time of passenger: the time from the passenger arriving at the bus stop to his/her alighting time at the destination bus stop. It is the waiting time of passengers at the bus stop plus the time onboard the bus.

\textbf{\textcolor{violet}{\#\# Agent definition}}

- Overview of the agent:  The agent will be activated and decide an action every time a dwell bus is ready to continue the trip after the boarding process of passengers or at the end of the last holding action. The action will last for 5 seconds, or say, the action step is 5 seconds.\\
- Agent's action: It is a binary variable. The value of '0' represents holding the bus at the stops for the next action step; the value of '1' represents not holding the bus, and the bus can continue its trip.\\
- Agent's state: It is a list [s0, s1, s2, s3, s4, s5]. Each element in this list is introduced below:\\
\hbox{\ \ \ \ }- s0: It is the forward space headway (meters) between the current bus and the forward bus in the same bus line.\\
\hbox{\ \ \ \ }- s1: It is the backward space headway (meters) between the current bus and the backward bus in the same bus line.\\
\hbox{\ \ \ \ }- s2: It is the forward space headway (meters) between the current bus and the forward bus in the \textbf{**other**} bus line. \textbf{**The value is 0 when the dwell stop is not a shared stop.**}\\
\hbox{\ \ \ \ }- s3: It is the backward space headway (meters) between the current bus and the backward bus in the \textbf{**other**} bus line. \textbf{**The value is 0 when the dwell stop is not a shared stop.**}\\
\hbox{\ \ \ \ }- s4: It is the number of onboard passengers of the current bus. Passengers can not board or alight when the bus is held, so the number of onboarding passengers is unchanged during holding control.\\
\hbox{\ \ \ \ }- s5: It is the holding time that has been spent during this dwell.

\textbf{\textcolor{violet}{\#\# Input parameters of the reward function}}

- Agent's action, denoted as \textcolor{purple}{\textasciigrave action\textasciigrave} \ in the code.\\
- Agent's state at the beginning of the action, denoted as \textcolor{purple}{\textasciigrave current\_state\textasciigrave} \ in the code.\\
- Agent's state at the end of the action, denoted as \textcolor{purple}{\textasciigrave next\_state\textasciigrave} \ in the code. Please be aware that the s4 in \textcolor{purple}{\textasciigrave next\_state\textasciigrave} \ is the same as the s4 in \textcolor{purple}{\textasciigrave current\_state\textasciigrave} \ since passengers boarding and alighting is not allowed during holding control.

\textbf{\textcolor{violet}{\#\# General simulation environment information}}

- Simulation environment and settings: There are two bus lines in the bus system. Each of the bus lines has 27 bus stops, and 8 of these stops are shared stops that serve both lines. The maximum holding time T = 90 s.\\
- Arriving time of passengers: The passengers' arrival time at each bus stop follows the Poisson process. Please be aware the arrival time of passengers is exogenous inputs which can not be changed by the agent control.\\
- Passengers are not allowed to board and alight during holding control.

\textbf{\textcolor{violet}{\#\# Input format}}

\textbf{\textcolor{violet}{\#\#\# Training and test trajectories}}

The trajectories are shown as a list, where each entry is a dictionary representing a trajectory. The first entry is the trajectory of total rewards at each training episode. The second entry contains the trajectory of states, actions, step rewards during the test, and the final results of the test.

Each item of "test\_history" is a list representing statistics of each single action step. When the "test\_history" is too long, it will be truncated to the last 50 steps.\\
"test\_results\_line\_1" and "test\_results\_line\_2" are the final results of bus line 1 and line 2 respectively. "test\_results\_shared\_part" is the relevant results of shared stops and passengers that can take either line 1 or line 2 to complete their trip. "test\_results\_overall" is the results for all passengers in the two lines.

The format is:\\

[\\
\hbox{\ \ \ \ }\{\\
\hbox{\ \ \ \ \ \ \ \ }"training\_history": \{\\
\hbox{\ \ \ \ \ \ \ \ \ \ \ \ }"total\_rewards": [total\_reward1, total\_reward2, ...],\\
\hbox{\ \ \ \ \ \ \ \ }\}\\
\hbox{\ \ \ \ }\},\\
\hbox{\ \ \ \ }\{\\
\hbox{\ \ \ \ \ \ \ \ }"test\_history": \{\\
\hbox{\ \ \ \ \ \ \ \ \ \ \ \ }"current\_states": [current\_state1, current\_state2, ...],\\
\hbox{\ \ \ \ \ \ \ \ \ \ \ \ }"actions": [action1, action2, ...],\\
\hbox{\ \ \ \ \ \ \ \ \ \ \ \ }"rewards": [reward1, reward2, ...],\\
\hbox{\ \ \ \ \ \ \ \ \ \ \ \ }"next\_states": [next\_states1, next\_states2, ...]\\
\hbox{\ \ \ \ \ \ \ \ }\},\\
\hbox{\ \ \ \ \ \ \ \ }"test\_results\_line\_1": \{\\
\hbox{\ \ \ \ \ \ \ \ \ \ \ \ }"SD\_time\_headways": SD\_time\_headways,\\
\hbox{\ \ \ \ \ \ \ \ \ \ \ \ }"avg\_passenger\_travel\_time": avg\_passenger\_travel\_time,\\
\hbox{\ \ \ \ \ \ \ \ \ \ \ \ }"avg\_passenger\_waiting\_time": avg\_passenger\_waiting\_time,\\
\hbox{\ \ \ \ \ \ \ \ \ \ \ \ }"avg\_holding\_time": avg\_holding\_time\\
\hbox{\ \ \ \ \ \ \ \ }\},\\
\hbox{\ \ \ \ \ \ \ \ }"test\_results\_line\_2": \{\\
\hbox{\ \ \ \ \ \ \ \ \ \ \ \ }"SD\_time\_headways": SD\_time\_headways,\\
\hbox{\ \ \ \ \ \ \ \ \ \ \ \ }"avg\_passenger\_travel\_time": avg\_passenger\_travel\_time,\\
\hbox{\ \ \ \ \ \ \ \ \ \ \ \ }"avg\_passenger\_waiting\_time": avg\_passenger\_waiting\_time,\\
\hbox{\ \ \ \ \ \ \ \ \ \ \ \ }"avg\_holding\_time": avg\_holding\_time\\
\hbox{\ \ \ \ \ \ \ \ }\},\\
\hbox{\ \ \ \ \ \ \ \ }"test\_results\_shared\_part": \{\\
\hbox{\ \ \ \ \ \ \ \ \ \ \ \ }"SD\_time\_headways": SD\_time\_headways,\\
\hbox{\ \ \ \ \ \ \ \ \ \ \ \ }"avg\_passenger\_travel\_time": avg\_passenger\_travel\_time,\\
\hbox{\ \ \ \ \ \ \ \ \ \ \ \ }"avg\_passenger\_waiting\_time": avg\_passenger\_waiting\_time\\
\hbox{\ \ \ \ \ \ \ \ }\},\\
\hbox{\ \ \ \ \ \ \ \ }"test\_results\_overall": \{\\
\hbox{\ \ \ \ \ \ \ \ \ \ \ \ }"avg\_passenger\_travel\_time": avg\_passenger\_travel\_time,\\
\hbox{\ \ \ \ \ \ \ \ \ \ \ \ }"avg\_passenger\_waiting\_time": avg\_passenger\_waiting\_time\\
\hbox{\ \ \ \ \ \ \ \ }\}\\
\hbox{\ \ \ \ }\}\\
]

where "SD\_time\_headways" is the standard deviation of bus time headways during the test. "avg\_passenger\_travel\_time" is the average travel time of passengers. "avg\_passenger\_waiting\_time" is the average waiting time of passengers at bus stops. "avg\_holding\_time" is the average holding time of buses.

\textbf{\textcolor{violet}{\#\# Output Requirements}}

Please write a proper analysis of the training performance (i.e., the convergence) and the test performance. Please try to give some suggestions on the reward improvement. You should not not be limited to the task description above, but also come up with other inefficient cases based on the training and test results.

\textbf{**The analysis and suggestions should be concise.**}

\textbf{\textcolor{violet}{\#\# Training and test results}}\\
\textbf{\textcolor{violet}{\#\#\# Training and test trajectories}}

\texttt{\{trajectories\}}

Now according to the \textbf{**Training and test results**}, please write your analysis and suggestions on the reward function improvement.

\end{tcolorbox} 

\subsection{Prompts of reward refiner}
\label{appendix:reward_refiner}

\begin{tcolorbox}[breakable=true, boxrule={0.5pt}, sharp corners={all}]
\setlength{\parskip}{1ex}
\small
You are now a proficient reward designer for a reinforcement learning (RL) agent. The agent is designed for a bus holding control problem to improve the performance of the bus system. The detailed description of the task is in the following section. 
Based on the previous reward function and the analysis as well as suggestions on the previous reward improvements, there is a modified reward function. The RL agent has been trained with the modified reward function, but the test results are not as good as those trained with the previous reward function. I will provide you with the modified reward function and corresponding test results as a failure modification. Please modify the previous reward function again based on the analysis and suggestions on the previous reward function, and considering the failure modification.

\textbf{\textcolor{violet}{\#\# Background}}

In bus transit systems, due to the heterogeneous bus dwell time and traffic disturbance, the buses easily get bunch where two or more buses get too close even though the departure times of the buses are well-designed, particularly affecting bus services. Buses can be held at bus stops to adjust the headways to avoid bus bunching and balance the bus headways. However, unreasonable and excessive holding control for buses will increase the total travel time for passengers.

\textbf{\textcolor{violet}{\#\# Task description}}

- Objective: There are two bus lines in the bus system. The two bus lines share part of the stops. The objective is to balance the time headways between buses from the same bus line as well as from different bus lines and minimize the total/average travel time of bus passengers. Since there is a strong correlation and consistent performance between time headway and space headway, balancing the time headways can be regarded as balancing the space headways. Avoiding excessive holding control is necessary as it leads to delays in travel time.

\textbf{\textcolor{violet}{\#\# Basic definition}}

- Time headway: the interval of arriving time between successive buses at the same bus stop. \\
- Space headway: the space between successive buses. \\
- Travel time of passenger: the time from the passenger arriving at the bus stop to his/her alighting time at the destination bus stop. It is the waiting time of the passenger at the bus stop plus the time onboard the bus.

\textbf{\textcolor{violet}{\#\# Agent definition}}

- Overview of the agent: The agent will be activated and decide an action every time a dwell bus is ready to continue the trip after the boarding process of passengers or at the end of the last holding action. The action will last for 5 seconds, or say, the action step is 5 seconds.\\
- Agent's action: It is a binary variable. The value of '0' represents holding the bus at the stops for the next action step (5 seconds); the value of '1' represents not holding the bus, and the bus can continue its trip.\\
- Agent's state: It is a list [s0, s1, s2, s3, s4, s5]. Each element in this list is introduced below:\\
\hbox{\ \ \ \ }- s0: It is the forward space headway (meters) between the current bus and the forward bus in the same bus line.\\
\hbox{\ \ \ \ }- s1: It is the backward space headway (meters) between the current bus and the backward bus in the same bus line.\\
\hbox{\ \ \ \ }- s2: It is the forward space headway (meters) between the current bus and the forward bus in the \textbf{**other**} bus line. \textbf{**The value is 0 when the dwell stop is not a shared stop.**}\\
\hbox{\ \ \ \ }- s3: It is the backward space headway (meters) between the current bus and the backward bus in the \textbf{**other**} bus line. \textbf{**The value is 0 when the dwell stop is not a shared stop.**}\\
\hbox{\ \ \ \ }- s4: It is the number of onboard passengers of the current bus. Passengers can not board or alight when the bus is held, so the number of onboarding passengers is unchanged during holding control.\\
\hbox{\ \ \ \ }- s5: It is the holding time that has been spent during this dwell.
    
\textbf{\textcolor{violet}{\#\# Input parameters of the reward function}}

- Agent's action, denoted as \textcolor{purple}{\textasciigrave action\textasciigrave} \ in the code.\\
- Agent's state at the beginning of the action, denoted as \textcolor{purple}{\textasciigrave current\_state\textasciigrave} \ in the code.\\
- Agent's state at the end of the action, denoted as \textcolor{purple}{\textasciigrave next\_state\textasciigrave} \ in the code. Please be aware that the s4 in \textcolor{purple}{\textasciigrave next\_state\textasciigrave} \ are the same as the s4 in \textcolor{purple}{\textasciigrave current\_state\textasciigrave} \ since passengers boarding and alighting is not allowed during holding control.

\textbf{\textcolor{violet}{\#\# General simulation environment information}}

- Simulation environment and settings: There are two bus lines in the bus system. Each of the bus lines has 27 bus stops, and 8 of these stops are shared stops that serve both lines. The maximum holding time T = 90 s.\\
- Arriving time of passengers: The passengers' arrival time at each bus stop follows the Poisson process. Please be aware the arrival time of passengers is exogenous inputs which can not be changed by the agent control.\\
- Passengers are not allowed to board and alight during holding control.

\textbf{\textcolor{violet}{\#\# Reward function requirements}}

- Please modify the previous reward function based on \textbf{**Analysis and suggestions on previous reward function**}, considering the \textbf{**Failure modification**}. The reward function needs to achieve the \textbf{**Task description**}. The information you can use to formulate the reward function has been listed in the \textbf{**Input parameters of the reward function**}. \\
- Stops serve a single bus line and two bus lines share the same reward function. Therefore, the formulation of the reward function should be generic.

\textbf{\textcolor{violet}{\#\# Previous reward function}}

\texttt{\{previous\_reward\_function\}}

\textbf{\textcolor{violet}{\#\# Analysis and suggestions on previous reward function}}

The previous reward function is used to train the reinforcement learning agent several times. Here is some analysis of failure and inefficiency and suggestions on the previous reward function:

\texttt{\{previous\_analysis\}}

\textbf{\textcolor{violet}{\#\# Failure modification}}

This is a failure modification of the previous reward function. The test results of the agent trained with the modified reward function are not as good as those trained with the previous reward function. This failure modification should be avoided when you modify the previous reward function again.

\textbf{\textcolor{violet}{\#\#\# Modified reward function}}

\texttt{\{current\_reward\_function\}}

\textbf{\textcolor{violet}{\#\#\# Test results of modified reward function}}

\texttt{\{current\_test\_results\}}

\textbf{\textcolor{violet}{\#\# Output Requirements}}

- Please consider the analysis and suggestions on the previous reward function and the failure modification above. Modify and improve the previous reward function. \\
\hbox{\ \ \ \ }1. You can both modify the current lines of the previous reward function and add new lines. You can use any variable in the \textbf{**Input parameters of the reward function**} to define the reward function. \\
\hbox{\ \ \ \ }2. If necessary, you can write a \textbf{**totally different**} reward function than the previous one.\\
\hbox{\ \ \ \ }3. Consider modifying the reward and penalty values in the previous reward function to balance them.\\
\hbox{\ \ \ \ }4. In the first part of the reward function, you should provide your thoughts on modifying the reward function. \textbf{**The thoughts should be concise.**}\\
\hbox{\ \ \ \ }5. Ensure the reward value will not be extremely large or extremely small which makes the reward meaningless.\\
\hbox{\ \ \ \ }6. Ensure there is no variable in the reward function that needs to be tuned. The reward function should be able to be used directly.
- The reward function should be written in Python 3.8.16.\\
- Output the code block only. \textbf{**Do not output anything else outside the code block**}.\\
- You should include \textbf{**sufficient comments**} in your reward function to explain your thoughts, the objective and \textbf{**implementation details**}. The implementation can be specified to a specific line of code.\\
- If you need to import packages (e.g. math, numpy) or define helper functions, define them at the beginning of the function. Do not use unimported packages and undefined functions.\\
- \textbf{**Please double-check the output code. Ensure there is no error. The variables or functions used should be defined already**}

\textbf{\textcolor{violet}{\#\# Output format}}

Strictly follow the following format. \textbf{**Do not output anything else outside the code block**}. \\
\\
\hbox{\ \ \ \ }def reward\_function(current\_state, action, next\_state):\\
    \hbox{\ \ \ \ \ \ \ \ }\#Thoughts: \\
    \hbox{\ \ \ \ \ \ \ \ }\#...\\
        \\
    \hbox{\ \ \ \ \ \ \ \ }\# (initial the reward)\\
    \hbox{\ \ \ \ \ \ \ \ }reward = 0\\
    \hbox{\ \ \ \ \ \ \ \ }\# (import packages and define helper functions)\\
    \hbox{\ \ \ \ \ \ \ \ }import numpy as np\\
    \hbox{\ \ \ \ \ \ \ \ }...\\
    \hbox{\ \ \ \ \ \ \ \ }return reward\\

Now write a new reward function to improve the \textbf{**Previous reward function**} based on \textbf{**Analysis and suggestions on previous reward function**}, and considering \textbf{**Failure modification**}. I will use the new reward function to train the RL agent and test it in the environment. \textbf{**Do not output anything else outside the code block**}. \textbf{**Please double-check the output code. Ensure there is no error. The variables or functions used should be defined already.**}  
\end{tcolorbox} 

\subsection{Prompts of LLM-based controller}
\label{appendix:LLM-based_controller}

The LLM-based controller is only conducted in case study 1. Due to the simplicity of the bus system, the RL agents' state in case study 1 is defined using only the forward and backward space headways and the number of onboard passengers. Therefore, there are only three parameters in the input state of the prompt.

\begin{tcolorbox}[breakable=true, boxrule={0.5pt}, sharp corners={all}]
\setlength{\parskip}{1ex}
\small
You are now a proficient bus-holding controller. Bus holding control aims to improve the performance of the bus system. The detailed description of the task is in the following section. Every time a bus arrives at a bus stop, I will provide you with the state. You should decide if the bus needs to be held or not. If the bus needs to be held, you need to decide the holding duration (a value from 0 to 90 seconds).

\textbf{\textcolor{violet}{\#\# Background}}

In bus transit systems, due to the heterogeneous bus dwell time and traffic disturbance, the buses easily get bunch where two or more buses get too close even though the departure times of the buses are well-designed, particularly affecting bus services. Buses can be held at bus stops to adjust the headways to avoid bus bunching and balance the bus headways. However, unreasonable and excessive holding control for buses will increase the total travel time for passengers.

\textbf{\textcolor{violet}{\#\# Task description}}

- Objective: Balance the time headways of buses and minimize the average travel time of bus passengers. Since there is a strong correlation and consistent performance between time headway and space headway, balancing the time headways can be regarded as balancing the space headways. 

\textbf{\textcolor{violet}{\#\# Basic definition}}

- Time headway: the interval of arriving time between successive buses at the same bus stop. \\
- Space headway: the space between successive buses. \\
- Travel time of passenger: the time from the passenger arriving at the bus stop to his/her alighting time at the destination bus stop. It is the waiting time of passengers at the bus stop plus the time onboard the bus.

\textbf{\textcolor{violet}{\#\# Format of input state}}

It is a list [s0, s1, s2]. Each element in this list is introduced below:\\
- s0: It is the space headway (meters) between the current bus and the forward bus.\\
- s1: It is the space headway (meters) between the current bus and the backward bus.\\
- s2: It is the number of onboard passengers of the current bus. 

\textbf{\textcolor{violet}{\#\# General simulation environment information}}

- Simulation environment and settings: It is a loop corridor consisting of 8 uniformly distributed bus stops and 6 buses. The length of the corridor is 10656 meters. All buses have a constant speed v = 5.55 m/s and the travel time excluding dwelling and holding between two consecutive bus stops is 4 minutes (i.e., the distance between two adjacent stops is 1332 meters). The buses travel cyclically in this corridor. The boarding times per passenger are set to 3.0 s/pax. The maximum holding time T = 90 s. \\
- Arriving time of passengers: The passengers' arrival time at each bus stop follows the Poisson process. Please be aware the arrival time of passengers is the exogenous input which can not be changed by the agent control.\\
- Positions of bus stops: The distances (meters) of the 8 bus stops from the origin are 666, 1998, 3330, 4662, 5994, 7326, 8658, and 9990, respectively.\\
- Passengers are not allowed to board and alight during holding control.

\textbf{\textcolor{violet}{\#\# Output requirements}}

- Please follow the two steps to make your decision:\\
\hbox{\ \ \ \ }- Step 1: Decide if the bus needs to be held. If the bus does not need to be held, please output 0. If the bus needs to be held, output 1 and then go to step 2.\\
\hbox{\ \ \ \ }- Step 2: Decide the holding duration which ranges from 0 to 90 seconds, and output the value only. If the bus does not need to be held, the output of step 2 is -1.\\
- The decision you made should aim to achieve the \textbf{**Task description**}. The information you can use to support your decision will be listed in the \textbf{**Input state**}. 

\textbf{\textcolor{violet}{\#\# Output format}}

Holding decision: insert output of step 1 here\\
Holding duration: insert output of step 2 here

\textbf{\textcolor{violet}{\#\# Input state}}

\texttt{\{input\_state\}}

Now please decide if the bus needs to be held based on \textbf{**Input state**}, and decide the holding duration if the bus needs to be held. Strictly follow the \textbf{**Output format**}. Do not output any analysis besides the output format.  
\end{tcolorbox}

\section{Final reward function}
\label{appendix:reward_function_case2}

In case study 2, the final reward function developed through the LLM-enhanced RL paradigm comprises five distinct components:

\begin{itemize}
    \item \textbf{Headway balance penalties for same line and different lines:} This component calculates penalties based on the absolute differences between the forward/backward space headways of the current bus and the ideal space headway. This penalty is designed to guide the space headway to be close to the ideal space headway.
    \item \textbf{Holding penalty:} The penalty for holding time is calculated as the square of the increase in holding time. This penalty discourages excessive holding delays.
    \item \textbf{Reward reductions in the standard deviation of headways:} This component rewards reductions in the variability of headways across the bus line. It calculates the standard deviation of headways for the current and next states, and assigns a reward based on the reduction in this standard deviation. This component encourages more consistent bus spacing. 
    \item \textbf{Passengers' holding delay penalty:} This penalty is calculated as the sum of the holding times of all onboard passengers, normalized by an average bus capacity. It aims to augment the penalty of holding delay.
    \item \textbf{Increase penalty for long holding duration:} This penalty component specifically targets holding actions that extend the total holding duration beyond 60 seconds. It aims to prevent prolonged holding duration.
\end{itemize}

These components are aggregated into a final multi-objective reward function, strategically balancing various operational goals to enhance the efficiency and reliability of bus systems. More details of the final reward formulation can be found below.

\begin{lstlisting}[language=Python]
def reward_function(current_state, action, next_state):
    # Thoughts:
    # This revised reward function aims to better align with the operational goals of the bus system, focusing on:
    # 1. Reducing headway variability to prevent bus bunching.
    # 2. Minimizing passenger waiting and travel times.
    # 3. Encouraging effective holding actions that contribute to improved system efficiency without excessive delays.
    # 4. Implementing a dynamic reward adjustment that considers the context of the system's current state, such as peak times.
    # 5. Using a multi-objective approach to balance the trade-offs between headway consistency, passenger time, and operational constraints.
    # The function has been structured to ensure the reward values are meaningful and not excessively large or small, enhancing training stability.

    # Import necessary package
    import numpy as np

    # Initialize the reward
    reward = 0

    # Extract state components
    # Current state headways and holding time
    s0_current, s1_current, s2_current, s3_current, passengers_current, holding_time_current = current_state
    # Next state headways and holding time
    s0_next, s1_next, s2_next, s3_next, passengers_next, holding_time_next = next_state

    # Define ideal headway (in meters)
    ideal_headway = 1650  # example value, should be set according to specific system characteristics

    # Calculate headway balance penalties for same line and different lines
    headway_penalty_same_line = np.abs(s0_next - ideal_headway) + np.abs(s1_next - ideal_headway)
    headway_penalty_diff_line = np.abs(s2_next - ideal_headway) + np.abs(s3_next - ideal_headway) if s2_next > 0 or s3_next > 0 else 0

    # Calculate holding penalty with a non-linear increase
    holding_penalty = (holding_time_next - holding_time_current)**2 if action == 0 else 0

    # Calculate the standard deviation of headways and reward reductions in it
    current_std_dev = np.std([s0_current, s1_current, s2_current, s3_current])
    next_std_dev = np.std([s0_next, s1_next, s2_next, s3_next])
    std_dev_reduction_reward = (current_std_dev - next_std_dev) * 10 if next_std_dev < current_std_dev else 0

    # Calculate passenger waiting time penalty
    waiting_time_penalty = holding_time_next * (passengers_current / 50)  # Normalize by an average bus capacity

    # Dynamic adjustment based on system state
    # Increase penalty for holding during peak times (assumed peak if holding time > 60 seconds)
    dynamic_penalty = 1.5 * holding_penalty if holding_time_next > 60 else holding_penalty

    # Multi-objective reward components
    # Balance between headway consistency and minimizing passenger time
    reward -= (headway_penalty_same_line + headway_penalty_diff_line + dynamic_penalty + waiting_time_penalty)
    reward += std_dev_reduction_reward

    return reward
\end{lstlisting}

\end{document}